%% file: ijcai23.tex
\documentclass{article}
\usepackage{ijcai23}

\input{math_commands.tex}

\usepackage{times}
\usepackage{soul}
\usepackage{url}
\usepackage[hidelinks]{hyperref}
\usepackage[utf8]{inputenc}
\usepackage[small]{caption}
\usepackage{graphicx}
\usepackage{amsmath}
\usepackage{amsthm}
\usepackage{booktabs}
\usepackage{algorithm}
\usepackage{algorithmic}
\usepackage[switch]{lineno}
\usepackage{multirow}
\usepackage{graphicx}
\usepackage{subfig}
\usepackage{xcolor}
\usepackage[scientific-notation=true]{siunitx}

\author{
Sagar Srinivas Sakhinana\thanks{Conceived, designed, implemented the research(programmed the software) and drafted the manuscript}, \{Krishna Sai Sudhir Aripirala, Shivam Gupta\}\thanks{Performed computational experiments, interpretation and visualization analysis of the results}, Venkataramana Runkana
\affiliations
TCS Research\\
\emails
\{sagar.sakhinana, k.aripirala, g.shivam4, venkat.runkana\}@tcs.com,
}

\begin{document}
\thispagestyle{myheadings}
\markboth{}{AI for Digital Twins and Cyber-Physical Applications Workshop, International Joint Conferences on Artificial Intelligence(IJCAI-23).} \nonumber

\title{Joint Hypergraph Rewiring and Memory-Augmented Forecasting Techniques in Digital Twin Technology}


\maketitle

\vspace{-1mm}
\section{ABSTRACT} 
Digital Twin technology creates virtual replicas of physical objects, processes, or systems by replicating their properties, data, and behaviors. This advanced technology offers a range of intelligent functionalities, such as modeling, simulation, and data-driven decision-making,  that facilitate design optimization, performance estimation, and monitoring operations. 
Forecasting plays a pivotal role in Digital Twin technology, as it enables the prediction of future outcomes, supports informed decision-making, minimizes risks, driving improvements in efficiency, productivity, and cost reduction. Recently, Digital Twin technology has leveraged Graph forecasting techniques in large-scale complex sensor networks to enable accurate forecasting and simulation of diverse scenarios, fostering proactive and data-driven decision-making. However, existing Graph forecasting techniques lack scalability for many real-world applications. They have limited ability to adapt to non-stationary environments, retain past knowledge, lack a mechanism to capture the higher-order spatio-temporal dynamics, and estimate uncertainty in model predictions. To surmount the challenges, we introduce a hybrid architecture that enhances the hypergraph representation learning backbone by incorporating fast adaptation to new patterns and memory-based retrieval of past knowledge. This balance aims to improve the slowly-learned backbone and achieve better performance in adapting to recent changes. In addition, it models the time-varying uncertainty of multi-horizon forecasts, providing estimates of prediction uncertainty. Our forecasting architecture has been validated through ablation studies and has demonstrated promising results across multiple benchmark datasets, surpassing state-of-the-art forecasting methods by a significant margin.
\vspace{-1mm}

\section{Introduction}
\vspace{-1mm}
Digital twins have several applications in various domains, including finance, retail and e-commerce, logistics and transport, healthcare, and many other domains. Digital Twins are useful in finance for risk management, trading, and investment decision-making. They enable financial institutions to simulate different scenarios and identify potential risks before they occur. They can help traders identify profitable opportunities and optimize their trades, while also allowing investors to model different economic scenarios and market conditions for better portfolio allocation strategies. Digital twins are useful in retail and ecommerce for creating virtual replicas of products, stores, and supply chains. This capability can contribute to transforming product design and development, streamlining operations, enhancing customer experiences, and driving sales growth. Digital Twins can be used in electricity pricing, auction, and design to optimize energy efficiency, reduce costs, and improve electricity markets. They can help energy analysts detect potential issues and optimize the layout and design of electricity grids to enhance energy efficiency and reduce costs. They can also assist electricity retailers in optimizing bidding strategies in electricity auctions to increase profits and reduce costs. Load forecasting is a crucial application of Digital Twins in electricity pricing, as it enables electricity distributors to accurately anticipate electricity demand and dynamically adjust pricing in real-time to prevent blackouts or brownouts. The digital twin technology involves creating a digital counterpart of a tangible entity, such as a machine, complex systems, or other physical objects. The creation of a digital twin involves utilizing diverse data sources, such as real-time sensor data, historical data, and other relevant information. By integrating this data into a processing system, the digital twin can effectively observe and record the key functionalities of the tangible entity. For instance, if the tangible entity under consideration is a gas turbine, a digital twin of the physical object would be created to mirror its exact specifications, such as size, shape, and technical features. Real-time sensor data from the turbine, including fuel injection rate, air-fuel ratio, inlet air temperature, and exhaust emissions, would be collected and fed into the digital twin. Subsequently, the digital twin would analyze this data and offer insights into the condition monitoring of the gas turbine. The digital twin can be employed to run simulations and analyze performance concerns for a wide range of applications, including fault diagnosis, safety monitoring, and performance optimization.  The digital twin technology offers the opportunity to test potential upgrades to a physical object in a virtual environment prior to real-world implementation. This approach provides valuable insights that can be implemented on the physical object, resulting in the ability to improve operational efficiency, minimize downtime, and reduce maintenance expenses. Of particular interest in this work is digital twin technology for forecasting of complex dynamical systems. Forecasting is a critical aspect of digital twin technology as it enables accurate predictions of the behavior of a physical object, enabling proactive maintenance, operational efficiency improvement and safety monitoring.   
Furthermore, the digital twin can forecast the expected behavior of the physical object in different scenarios, enabling operators to optimize its performance and reduce downtime, while minimizing risks associated with implementing untested changes on the actual physical object.
As a result, it is imperative to develop accurate models of physical systems in order to create Digital Twins that can faithfully replicate the behavior of the physical systems for forecasting purposes.

\vspace{-3mm}
\section{Related Work on Time Series Forecasting}
\vspace{-1mm}
Accurately forecasting the behavior of complex dynamical systems, which are characterized by high-dimensional multivariate time series(MTS) in interconnected sensor networks, is crucial for enabling well-informed decision-making in various applications. Forecasting MTS data is challenging due to the intricate relationships among multiple time series variables and the unique features of MTS data, including non-linearity, high-dimensionality and non-stationarity. 
The spatio-temporal graph neural networks(STGNNs) have become a popular approach to model the relational dependencies between time series variables in the MTS data for multivariate time series forecasting. Several researchers (e.g., \cite{zonghanwu2019,Bai2020nips,wu2020connecting,YuYZ18,tampsgcnets2022,LiYS018}) have contributed to this trend, and their work has significantly advanced the use of GNNs in time series forecasting task. Training STGNNs on the fly is challenging due to their inability to adjust to non-stationary environments and retain past knowledge. The ability of STGNNs to adapt quickly is critical, and successful approaches must handle changes to both new and recurring patterns effectively. However, STGNNs , despite their strong representation learning capabilities, face two major challenges when dealing with time series data streams. Firstly, training STGNNs on data streams in a straightforward manner requires a considerable number of samples to converge. This is because mini-batches or multiple epoch training, commonly used in offline training, are not feasible. Thus, when there is a distribution shift, such neural architectures can become cumbersome and require a large number of samples to learn new concepts effectively, which can ultimately result in suboptimal performance. In essence, the primary challenge lies in the absence of a mechanism within STGNNs to facilitate learning on continuously generated data streams effectively. As a result, the STGNNs must adapt to new trends and patterns in data streams over time. The second challenge arises from the fact that time series data frequently displays recurring patterns that may cease to exist temporarily and then reappear in the future. STGNNs are prone to the catastrophic forgetting phenomenon, whereby the model discards previously acquired knowledge when presented with new data, leading to suboptimal learning of recurring patterns. As a result, this limitation further hinders the overall performance of STGNNs for time series forecasting. Existing STGNNs can learn MTS data dynamics by simultaneously inferring discrete dependency graph structures or by leveraging domain expertise knowledge of predefined relationships among multiple time series variables. While complex dynamical systems consist of interconnected networks, these networks may have higher-order structural relations that extend beyond pairwise associations. Hypergraphs, which provide a more generalized representation of graphs, can effectively model such relations in high-dimensional MTS data. Furthermore, conventional STGNNs prioritize pointwise forecasting and do not offer uncertainty estimates associated with these multi-horizon forecasts. To tackle these challenges, we introduce the Joint Hypergraph Rewiring and Forecasting Neural Framework, which we will refer to as \textbf{JHgRF-Net} for brevity. The proposed framework achieves continual learning by balancing two objectives: (i) leveraging prior knowledge to facilitate rapid learning of current trends and patterns, and (ii) maintaining and updating previously acquired knowledge. The \textbf{JHgRF-Net} framework achieves dynamic balance between rapid adaptation to recent changes and retrieval of similar old knowledge by leveraging the interaction between two complementary components: the Spatio-Temporal Hypergraph Convolutional Network(STHgCN) and the Spatio-Temporal Transformer Network(STTN). The Mixture of Experts(MOE) approach is utilized to design algorithmic architecture for hypergraph time series forecasting. This approach involves using the aforementioned set of complementary modeling approaches, whose predictions are combined to create a robust mechanism capable of improving the overall accuracy of forecasting.
The STHgCN neural operator simultaneously infers discrete dependency hypergraph structure and learns MTS data dynamics. The STHgCN neural operator consists of two sequentially operating modules: hypergraph-structure learning(HgSL) and hypergraph representation learning(HgRL). The HgSL module infers the discrete dependency hypergraph structure and performs hypergraph rewiring to modify the hyperedges so that they better reflect the dependencies between hypernodes. This can involve adding or removing hyperedges to optimize the relational structure between hypernodes. The HgRL module models the spatio-temporal dynamics underlying the hypergraph-structured MTS data for multi-horizon forecasting. The STTN neural operator learns the underlying dynamics of MTS data beyond the original sparse relational hypergraph structure through a self-attention mechanism. The STTN neural operator learns the underlying dynamics of MTS data beyond the original sparse relational hypergraph structure through a self-attention mechanism. A gating mechanism is utilized to regulate the information flow from complementary components. This mechanism further distills knowledge and improves the accuracy and reliability of the model's predictions. Moreover, the framework captures time-varying uncertainty in forecasts. As a result, the framework provides accurate multi-horizon predictions and reliable uncertainty estimates of forecasts. Furthermore, the framework is designed to provide superior generalization and scalability for large-scale spatio-temporal MTS forecasting tasks that are commonly encountered in real-world applications.

\vspace{-3mm} 
\section{Problem Formulation}
\vspace{-1mm}
Let us consider a historical time series dataset with $n$ correlated variables observed over $\mathrm{T}$ time steps. The dataset is represented by the notation \thickmuskip=0.15\thickmuskip\resizebox{.135\textwidth}{!}{$\mathbf{X} = \big(\mathbf{x}_{1}, \ldots, \mathbf{x}_{\mathrm{T}}  \big)$}, where the subscript indicates the time step. The observations of all the variables at time step $t$ are denoted by \thickmuskip=0.15\thickmuskip\resizebox{.285\textwidth}{!}{$\mathbf{x}_{t} = \big(\mathbf{x}_t^{(1)}, \mathbf{x}_t^{(2)}, \ldots, \mathbf{x}_t^{(n)}\big) \hspace{1mm} \in \hspace{1mm} \mathbb{R}^{(n \times c)}$}, where the superscript refers to the variables. Each sensor can measure multiple physical quantities denoted by $c$. For example, in intelligent transportation systems, the traffic loop detectors or traffic sensors placed across travel lanes can simultaneously measure three parameters: traffic flow, speed, and volume. Therefore, in this particular case, $c$ = 3. In MTSF, we use a rolling-window technique to predict the future values of n-correlated variables for the forecast horizon. At each time step $t$, we define a look-back window which includes the prior $\tau$-steps of time series data to predict the next $\upsilon$-steps. We use a historical window of $n$-correlated variables, observed over the previous $\tau$-steps prior to time step $t$, represented by \resizebox{.18\textwidth}{!}{$\mathbf{X}_{(t - \tau : \hspace{1mm}t-1)} \hspace{0.5mm} \in \hspace{0.5mm} \mathbb{R}^{n \times \tau \times c}$}, to predict the future values of $n$-variables for the next $\upsilon$-steps, represented by \resizebox{.18\textwidth}{!}{$\mathbf{X}_{(t  : t + \upsilon - 1)} \hspace{0.5mm} \in \hspace{0.5mm}\mathbb{R}^{n \times \upsilon \times c}$}. To capture complex higher-order relationships among variables within the MTS data, we represent the historical data as continuous-time spatial-temporal hypergraphs denoted by $\mathbf{G}_{t}$. Hypergraphs consist of hypernodes($\mathbf{V}$), representing time series variables and hyperedges($\mathbf{E}$) that capture hierachial relationships among an arbitrary number of hypernodes.  The time-dependent hypernode feature matrix is denoted by $\mathbf{X}_{(t - \tau : \hspace{1mm}t-1)}$. We learn the implicit hypergraph structure through an embedding-based similarity metric learning approach. The incidence matrix $\mathbf{I} \in \mathbb{R}^{n \times m}$ describes the hypergraph structure, where $\mathbf{I}_{p, \hspace{0.5mm}q}=1$ if hyperedge $q$ is incident with hypernode $p$, and 0 otherwise. Hypergraph sparsity is determined by the number of hyperedges in the hypergraph.  In a sparse hypergraph, the number of hyperedges($\text{m}(|\mathbf{E}|)$) is relatively small compared to the number of hypernodes($\text{n}(|\mathbf{V}|)$), while in a dense hypergraph, the number of hyperedges is relatively large. Sparser hypergraphs generally result in more efficient algorithms, due to the impact of hypergraph sparsity on computational efficiency and algorithmic complexity. A hypergraph with more hyperedges has a denser and more complex structure, resulting in a higher level of connectivity among the hypernodes. Conversely, a hypergraph with fewer hyperedges has a sparser structure with fewer connections between the hypernodes. The proposed framework aims to learn a differentiable function $F(\theta)$ that can predict the future estimates $\mathbf{X}_{(t : t + \upsilon - 1)}$, of historical window inputs  $\mathbf{X}_{(t - \tau : \hspace{1mm}t-1)}$, given a hypergraph $\mathbf{G}_{t}$. To put it briefly, the function $F(\theta)$ takes in the past observations and hypergraph structure, represented by $[\mathbf{x}_{(t - \tau)}, \cdots, \mathbf{x}_{(t-1)} ; \mathbf{G}_{t}]$, and predict future observations, denoted as $[\mathbf{x}_{(t + 1)}, \cdots, \mathbf{x}_{(t + \upsilon-1)}]$. This is mathematically represented as:

\vspace{-4mm}
\resizebox{0.945\linewidth}{!}{
\begin{minipage}{\linewidth}
\begin{align}
\left[\mathbf{x}_{(t - \tau)}, \cdots, \mathbf{x}_{(t-1)} ; \mathbf{G}_{t}\right] \stackrel{F(\theta)}{\longrightarrow}\left[\mathbf{x}_{(t + 1)}, \cdots, \mathbf{x}_{(t + \upsilon-1)}\right] \nonumber
\end{align} 
\end{minipage} 
}

\vspace{2mm}
The MTSF task formulated on the implicit hypergraph(\resizebox{.02\textwidth}{!}{$\mathbf{G}_{t}$}), can be expressed as shown below:

\vspace{-3mm}
\resizebox{0.945\linewidth}{!}{
\begin{minipage}{\linewidth}
\begin{align}
\min _{\theta} \mathcal{L}_{\text{MAE}}\big(\mathbf{X}_{(t : t + \upsilon-1)}, \widehat{\mathbf{X}}_{(t : t + \upsilon-1)} ; \mathbf{X}_{(t - \tau : \hspace{1mm}t-1)}, \mathbf{G}_{t}\big) \nonumber
\end{align} 
\end{minipage}
}

\vspace{1mm}
The function $F(\theta)$ involves a set of parameters $\theta$ which can be trained to optimize its performance. The predicted future observations is denoted by $\widehat{\mathbf{X}}_{(t : t + \upsilon-1)}$. To train the learning algorithm, we minimize the loss function denoted by $\mathcal{L}_{\text{MAE}}$, i.e., the mean absolute error(MAE), which is defined as:

\vspace{-2mm}
\resizebox{0.945\linewidth}{!}{
\begin{minipage}{\linewidth}
\begin{align}
\mathcal{L}_{\text{MAE}} \hspace{0.5mm} = \hspace{0.5mm}\frac{1}{\upsilon}\left|\mathbf{X}_{(t : t + \upsilon-1)}-\widehat{\mathbf{X}}_{(t : t + \upsilon-1)}\right| \nonumber
\end{align}
\end{minipage}
}

Here, $\mathbf{X}_{(t : t + \upsilon-1)}$ is the actual future MTS data, and $\frac{1}{\upsilon}$ is a scaling factor.

\vspace{-1mm}
\begin{figure}[!ht]
\resizebox{1.225\linewidth}{!}{ 
\hspace*{-35mm}\includegraphics[keepaspectratio,height=18.0cm,trim=2.5cm 5.5cm 0cm 6.0cm,clip]{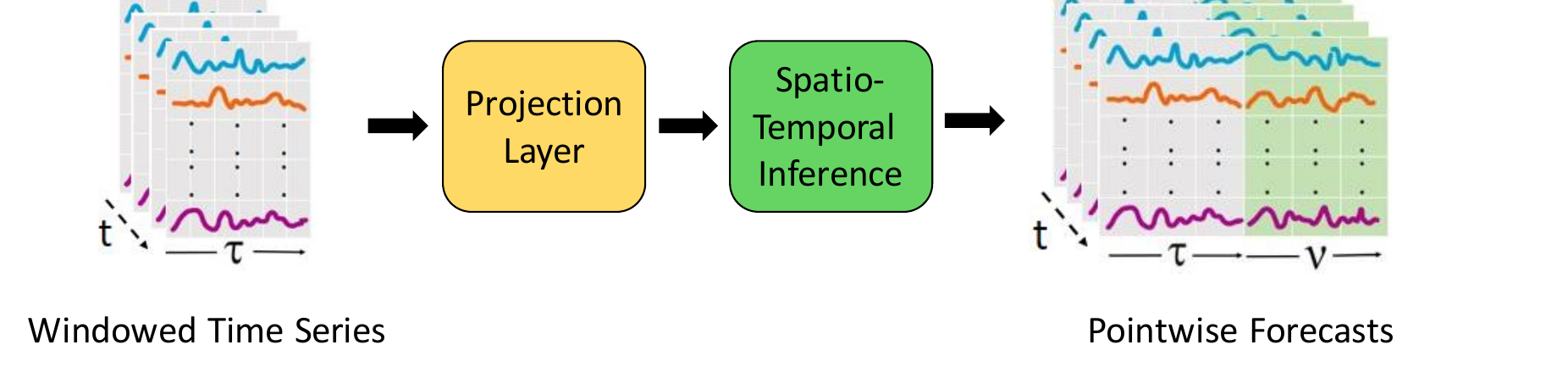} 
}
\vspace{-8mm}
\caption{Overview of \textbf{JHgRF-Net} framework.}
\label{fig:figure1}
\end{figure}

\vspace{-7mm}
\section{OUR APPROACH}
\vspace{-1mm}
Our proposed neural forecasting framework consists of two key components: the projection layer and the spatio-temporal feature extractor, as shown in Figure \ref{fig:figure1}. The spatio-temporal inference component includes two distinct methods for hypergraph representation learning: the Spatio-Temporal Hypergraph Convolutional Network(STHgCN) and the Spatio-Temporal Transformer Network(STTN). The STHgCN method employs hypergraph as a mathematical model for learning the underlying higher-order relations of the time series variables. This is achieved by optimizing the discrete hypergraph structure underlying the observed data. It then peforms the gated hypergraph convolution operations on the hypergraph-structured MTS data to model the intricate spatio-temporal dynamics within the latent hypernode-level representations.
The final representations can then be used to predict multi-horizon forecasts.  The STTN method is a powerful technique for modeling the hypergraph-structured MTS data. The STTN method extends transformer networks to handle arbitrary sparse hypergraph structures with full attention as a useful inductive bias. This enables the model to learn intra- and inter-correlations among the variables without being limited by the hierarchical structural information underlying the MTS data. It leverages task-specific relations between variables beyond the original sparse structure to generate expressive hypernode-level representations that improve forecast accuracy. We use a gating mechanism to regulate the flow of information from the two methods. This enables us to learn optimal representations of the hypernode-level representations that capture the accurate dynamics of complex interconnected sensor networks. To summarize, our framework performs the joint optimization of the different learning components to generate accurate forecasts across multiple forecast horizons, while also ensuring reliable estimates of uncertainty for time-series forecasting tasks.

\vspace{0mm}
\subsection{PROJECTION LAYER}
\vspace{-1mm}
The proposed framework uses a projection layer with gated linear networks(GLN, \cite{dauphin2017language}) to obtain non-linear representations of input data. Specifically, the input data \resizebox{.18\textwidth}{!}{$\mathbf{X}_{(t - \tau : \hspace{1mm}t-1)} \hspace{0.5mm} \in \hspace{0.5mm} \mathbb{R}^{n \times \tau \times c}$} is transformed through a gating mechanism, resulting in \resizebox{.18\textwidth}{!}{$\overline{\mathbf{X}}_{(t  : t + \upsilon - 1)} \hspace{0.5mm}  \in \hspace{0.5mm} \mathbb{R}^{n \times \upsilon \times d}$}, which represents the non-linear transformed input data. It is described as follows:

\vspace{-2mm}
\resizebox{0.945\linewidth}{!}{
\begin{minipage}{\linewidth}
\begin{align}
\centering
\overline{\mathbf{X}}_{(t  : t + \upsilon - 1)}   = \big( \sigma(\text{W}_{0}\mathbf{X}_{(t - \tau : \hspace{1mm}t-1)}) \otimes \text{W}_{1}\mathbf{X}_{(t - \tau : \hspace{1mm}t-1)}\big)\text{W}_{2} \nonumber
\end{align}
\end{minipage}
}

\vspace{1mm}
Here, the trainable weight matrices are $\text{W}_{0}, \text{W}_{1} \in \mathbb{R}^{c \times d}, \text{W}_{2} \in \mathbb{R}^{\tau \times \upsilon}$, and the element-wise multiplication is denoted by $\otimes$. The utilization of a non-linear activation function $\sigma$ improves representation learning and enables the framework to effectively learn and model complex patterns present in the MTS data.

\vspace{-4mm}
\begin{figure}[h]
\resizebox{1.105\linewidth}{!}{ 
\hspace*{-18.5mm}\includegraphics[keepaspectratio,height=4.5cm,trim=1.0cm 4.5cm 0cm 2.0cm,clip]{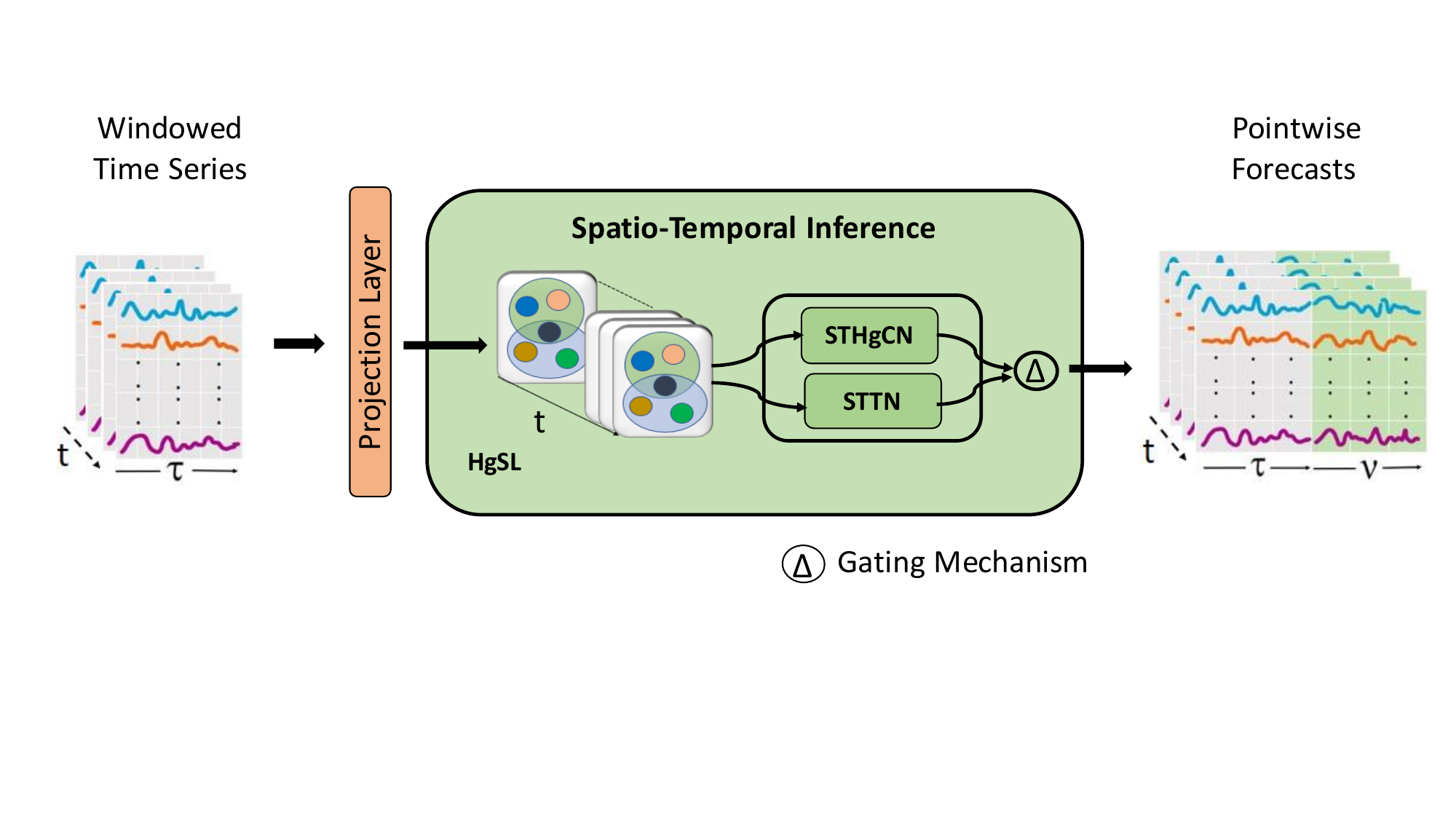} 
}
\vspace{-8mm}
\caption{The above figure illustrates the \textbf{JHgRF-Net} framework, which jointly learns to incorporate a discrete dependency hypergraph structure between multiple time series variables in order to capture complex interactions. The hypergraph representation learning algorithm is used to model intrinsic spatio-temporal dependencies for multi-horizon forecasting.}
\label{fig:figure2}
\end{figure}

\vspace{-7mm}
\subsection{SPATIAL-INFERENCE}
\vspace{-1mm}
Figure \ref{fig:figure2} illustrates the spatio-temporal feature extractor of the framework, which consists of two distinct methods(STHgCN and STTN). Further information regarding each method will be elaborated in the subsequent sections.

\vspace{-1mm}
\subsubsection{Spatio-Temporal Hypergraph Convolutional Network(STHgCN)}

The STHgCN method comprises sequentially operating modules, including hypergraph structure learning(HgSL) and hypergraph representation learning(HgRL) modules. The following sections will elaborate on each module and provide more details.

\vspace{-2mm}
\paragraph{Implicit hypergraph Inferenece}
\vspace{1mm}
The HgSL module uses an embedding-based similarity metric learning technique to capture higher-order dependency relationships between different variables in the MTS data and computes an optimal discrete hypergraph structure for a hypergraph-structured representation of the MTS data.
In short, the implicit hypergraph provides a spatio-temporal inductive bias that enables a structured representation of the MTS data, capturing the underlying relationships and dependencies among the variables. The hypernodes and hyperedges of the hypergraph are represented by the differentiable embeddings in the $d$-dimensional vector space, $\mathbf{z_{i}}, \mathbf{z_{j}} \in \mathbb{R}^{(d)}$, where $1 \leq i \leq n$ and $1 \leq j \leq m$. By leveraging the learned embeddings to transform the MTS data into a hypergraph-structured time series data, the HgSL module computes the optimal hypergraph topology that captures the task-relevant relationships and dependencies among the variables, making it a powerful tool for learning relational hypergraph structures from complex MTS data. The pairwise similarity(\resizebox{.0275\textwidth}{!}{$\text{P}_{i,j}$}) between any pair of $\mathbf{z_{i}}$ and $\mathbf{z_{j}}$ is computed as follows:

\vspace{-5mm}
\resizebox{0.945\linewidth}{!}{
\begin{minipage}{\linewidth}
\begin{align}
\text{P}_{i,j} \hspace{0.5mm} = \hspace{0.5mm} \sigma \big([\text{S}_{i,j} || 1- \text{S}_{i,j}]\big); \text{S}_{i,j} \hspace{0.5mm} = \hspace{0.5mm} \frac{\mathbf{z^{T}_{i}} \mathbf{z_{j}} + 1}{2\left\|\mathbf{z_{i}}\right\| \cdot\left\|\mathbf{z_{j}}\right\|}  \nonumber
\end{align}
\end{minipage}
}

\vspace{1mm}
where \resizebox{.00985\textwidth}{!}{$\Vert$} denotes vector concatenation. The differentiable, sigmoid activation function is applied to map the pairwise scores to the interval [0,1]. The hyperedge probability over hypernodes of the hypergraph is represented as \resizebox{.115\textwidth}{!}{$\text{P}^{(k)}_{i,j} \in \mathbb{R}^{nm \times 2}$}, where $k \in \{0,1\}$. The scalar value of \resizebox{.085\textwidth}{!}{$\text{P}^{(k)}_{i,j} \in [0,1]$} indicates the relationship between a pair of hypernodes and hyperedges, indexed by $(i,j)$. To be precise, \resizebox{.0325\textwidth}{!}{$\text{P}^{(0)}_{i,j}$} represents the probability of hypernode $i$ being connected to hyperedge $j$, while \resizebox{.0325\textwidth}{!}{$\text{P}^{(1)}_{i,j}$} denotes the probability that hypernode $i$ is not connected to hyperedge $j$. To accurately and efficiently sample discrete hypergraph structures from the hyperedge probability distribution $\text{P}_{i,j}$, we leverage the Gumbel-softmax trick introduced in \cite{jang2016categorical}. This technique is powerful in capturing complex relationships among variables in MTS data, making the HgSL module more effective. The connectivity pattern of the hypergraph structure is then represented using an incidence matrix $\mathbf{I} \in \mathbb{R}^{n \times m}$, which captures the relationships between hypernodes and hyperedges in the hypergraph.
 By using the Gumbel-softmax trick, we can learn the hypergraph structure in an end-to-end differentiable manner. Thus, it becomes possible to apply the gradient-based optimization methods during model training, enabling an inductive-learning approach to learn complex underlying structures within the MTS data. The Gumbel-Softmax trick involves using random noise from the Gumbel distribution to perturb the hyperedge probability distribution and then sampling the optimal discrete structure from the distribution using the Softmax function. The incidence matrix is obtained as,

\vspace{-2mm}
\resizebox{0.975\linewidth}{!}{
\begin{minipage}{\linewidth}
\begin{align}
\mathbf{I}_{i,j} \hspace{0.5mm} = \hspace{0.5mm}\exp \big(\big(g^{(k)}_{i,j} + \text{P}^{(k)}_{i,j} + \epsilon\big) / \gamma\big)\big/{\sum \exp \big(\big(g^{(k)}_{i,j} + \text{P}^{(k)}_{i,j} + \epsilon\big) / \gamma\big)} \nonumber
\end{align}
\end{minipage}
}

\vspace{2mm}
where the temperature parameter($\gamma$) of Gumbel-Softmax trick is set to 0.05, and $\epsilon$ is a small constant added to avoid numerical instability. Random noise, denoted by \resizebox{.325\textwidth}{!}{$g^{(k)}_{i j} \sim \operatorname{Gumbel}(0,1) = \log (-\log (\text{U}(0,1))$} is sampled from the Gumbel distribution, where $\text{U}$ represents the uniform distribution with a range of 0 to 1. We optimize the hypergraph distribution parameters to ensure that the learned hypergraph is sparse, eliminating redundant hyperedges over hypernodes. The forecasting task provides indirect supervisory information that helps to reveal the hypergraph relation structure in the observed MTS data. In summary, the HgSL module learns the latent hypergraph structure of multiple interacting time series variables to create a structured representation of the time series data, which facilitates downstream multi-horizon forecasting with predictive uncertainty estimation.

\vspace{-2mm}
\paragraph{Hypergraph Attention Network(HgAT)}
\vspace{1mm}
The HgAT neural operator extends attention-based convolution operations to non-Euclidean domains, such as hypergraphs. It accurately models the complex hypergraph-structured MTS data, thereby improving multi-horizon forecast accuracy. The HgAT operator captures spatial correlations among time-series variables by encoding relational inductive bias within the hypergraph's connectivity. It performs message-passing schemes to propagate information through the hypergraph-structured MTS data, which is characterized by an incidence matrix (represented by \resizebox{.08\textwidth}{!}{$\mathbf{I} \in \mathbb{R}^{n \times m}$}) and a feature matrix (represented by \resizebox{.19\textwidth}{!}{$\overline{\mathbf{X}}_{(t  : t + \upsilon - 1)} \hspace{0.5mm}  \in \hspace{0.5mm} \mathbb{R}^{n \times \upsilon \times d}$}) to compute the hypernode representation matrix (represented by \resizebox{.175\textwidth}{!}{$\mathbf{H}_{(t  : t + \upsilon - 1)} \in \hspace{0.5mm} \mathbb{R}^{n \times \upsilon \times d}$}). Each row in the matrix $\mathbf{H}_{(t  : t + \upsilon - 1)}$ represents the hypernode representations, \resizebox{.0825\textwidth}{!}{$\mathbf{h}^{t}_{i} \in \mathbb{R}^{\upsilon \times d}$}. The HgAT operator captures relationships among time-series variables by encoding structural and feature characteristics of spatio-temporal hypergraphs in hypernode representations. It adapts to changes in time-series variable dependencies over time in the hypernode representations $\mathbf{h}^{t}_{i}$. The HgAT operator models spatio-temporal correlations among time-series variables in hypergraph-structured MTS data using intra-edge and inter-edge neighborhood aggregation schemes. The intra-edge aggregation considers hypernodes associated with a specific hyperedge, while inter-edge aggregation considers hyperedges connected to a specific hypernode. In hypergraph-structured MTS data, hyperedges capture relationships between multiple time-series variables, which can have varying degrees of correlation and complexity. Let the notation $\mathbf{N}_{j, i}$ represent a subset of hypernodes $i$ associated with a specific hyperedge $j$. The intra-edge neighborhood of a hypernode $i$, denoted as $\mathbf{N}_{j,i} \backslash i$, captures a localized cluster of semantically-corelated time-series variables and their higher-order relationships. The inter-edge neighborhood of a hypernode $i$, represented by $\mathbf{N}_{i, j}$, includes the set of hyperedges $j$ connected to that hypernode, providing a more comprehensive understanding that each variable may have multiple and potentially complex relationships with other time-series variables in the data. We use attention-based intra-edge neighborhood aggregation to obtain latent hyperedge representations, which leads to a more comprehensive understanding of the MTS data. This approach can be described as follows:

\vspace{0mm}
\resizebox{0.935\linewidth}{!}{
\begin{minipage}{\linewidth}
\begin{equation}
\mathbf{h}^{(t, \ell)}_{j} =  \sum_{z=1}^{\mathbf{Z}} \sigma \big( \hspace{-0.25mm}  \sum_{i \hspace{0.5mm}\in \hspace{0.5mm}{\mathbf{N}_{j, i}}} \hspace{-1mm}  \alpha^{(t, \ell, z)}_{j, i} \mathbf{W}^{(z)}_{0}\mathbf{h}^{(t, \ell-1, z)}_{i} \big) \nonumber
\end{equation}
\end{minipage}
}

where the hyperedge representations at layer $\ell$ are denoted by $\mathbf{h}^{(l)}_{j} \in \hspace{0.5mm} \mathbb{R}^{\upsilon \times d}$. Each hypernode's initial representation is its corresponding feature vector,

\vspace{0mm}
\resizebox{0.945\linewidth}{!}{
\begin{minipage}{\linewidth}
\begin{equation}
\mathbf{h}^{(t, 0, z)}_{i} = \hspace{1mm}\overline{\mathbf{x}}^{(t)}_{i} \nonumber
\end{equation}
\end{minipage}
}

where $\overline{\mathbf{x}}^{(t)}_{i}  \in \hspace{0.5mm} \mathbb{R}^{\upsilon \times d}$ represents the $i^{th}$ row of the feature matrix \resizebox{.17\textwidth}{!}{$\overline{\mathbf{X}}_{(t  : t + \upsilon - 1)} \in \mathbb{R}^{n \times \upsilon \times d}$}. At each layer, the HgAT operator produces multiple representations denoted by $\mathbf{h}^{(l,z)}_{j}$ of the input data, each with its own set of parameters, and combines them by summation. This enables the HgAT operator to capture various aspects of the relations underlying the intra-edge neighborhood in the hypergraph-structured MTS data. To determine the attention coefficient $\alpha_{j, i}$ for the hypernode $i$ incident with hyperedge $j$, we compute its relative importance as follows:

\vspace{-3mm}
\resizebox{0.945\linewidth}{!}{
\hspace{0.0cm}\begin{minipage}{\linewidth}
\begin{align}
e^{(t, \ell, z)}_{j, i} &= \operatorname{ReLU}\big(\text{W}^{(z)}_{0} \mathbf{h}^{(t, \ell-1, z)}_{i}\big) \nonumber \\
\alpha^{(t, \ell, z)}_{j, i} &= \frac{\exp \big(e^{(t, \ell, z)}_{j, i}\big)}{{\textstyle \sum_{k \hspace{0.5mm}\in \hspace{0.25mm}{\mathbf{N}_{j, i}}\hspace{0.25mm} \cup \hspace{1mm}i} \exp \big(e^{(t, \ell, z)}_{j, k}\big)}}  \nonumber
\end{align}
\end{minipage}
}

\vspace{1mm}
where $e_{j, i}$ denotes the unnormalized attention score. The HgAT method utilizes an attention-based inter-edge neighborhood aggregation scheme, which captures complex dependencies and relationships between hyperedges and hypernodes. It generates expressive hypernode representations by summing over ReLU activations of linear transformations of previous layer hypernode representations and weighted hyperedge representations. This is described below,

\vspace{0mm}
\resizebox{0.945\linewidth}{!}{
\hspace{0cm}\begin{minipage}{\linewidth}
\begin{equation}
\mathbf{h}^{(t, \ell)}_{i}=\sum_{z=1}^{\mathbf{Z}} \operatorname{ReLU}\big(\text{W}^{(z)}_{0}\mathbf{h}^{(t, \ell-1, z)}_{i} + \sum_{j \in \mathbf{N}_{i, j}} \beta^{(t, \ell, z)}_{i, j} \text{W}^{(z)}_{1} \mathbf{h}^{(t, \ell, z)}_{j}\big) \nonumber
\end{equation}
\end{minipage}
} 

The weight matrices that are trained are represented as \resizebox{.15\textwidth}{!}{$\text{W}^{(z)}_{0}, \text{W}^{(z)}_{1} \in \mathbb{R}^{d \times d}$}. The $\operatorname{ReLU}$ activation function is used to introduce non-linearity while updating the hypernode-level representations. The attention scores $\beta_{i, j}$ are normalized and determine the relevance of each hyperedge $j$ that is incident with hypernode $i$. This allows the HgAT operator to focus on the most significant hyperedges, and the attention scores are computed as follows:

\vspace{-2mm}
\resizebox{0.955\linewidth}{!}{
\hspace{0.5cm}\begin{minipage}{\linewidth}
\begin{align}
\phi^{(t, \ell, z)}_{i, j} &= \operatorname{ReLU}\big(\text{W}^{(z)}_{3} \cdot \big(\text{W}^{(z)}_{2}\mathbf{h}^{(t, \ell-1, z)}_{i} \oplus \text{W}^{(z)}_{2} \mathbf{h}^{(t  , \ell, z)}_{j}\big)\big) \nonumber \\
\beta^{(t, \ell, z)}_{i, j} &=  \frac{\exp (\phi^{(t, \ell, z)}_{i, j})}{{\textstyle \sum_{k \hspace{0.5mm}\in \hspace{0.5mm}{\mathbf{N}_{i, j}} \hspace{0.25mm} \cup \hspace{1mm}j} \exp (\phi^{(t,\ell, z)}_{i, k})}}\nonumber
\end{align}
\end{minipage}
} 

\vspace{1mm}
where \resizebox{.10\textwidth}{!}{$\text{W}^{(z)}_{2} \in \mathbb{R}^{d \times d}$} and \resizebox{.10\textwidth}{!}{$\text{W}^{(z)}_{3} \in \mathbb{R}^{2d}$} are trainable weight matrix and vector, respectively. $\oplus$ denotes the concatenation operator. The unnormalized attention score is denoted by $\phi_{i, j}$. Batch normalization and dropout techniques are used to enhance generalization and mitigate overfitting. A gating mechanism is employed to selectively combine features from $\overline{\mathbf{x}}^{(t)}_{i}$ and $\mathbf{h}^{(t, \ell)}_{i}$ in a differentiable way. These methods improve the HgAT operator reliability and accuracy for the downstream MTSF task.

\vspace{-1mm}
\resizebox{0.955\linewidth}{!}{
\hspace{0cm}\begin{minipage}{\linewidth}
\begin{align}
g^{(t)}  &= \sigma \big( f_s(\mathbf{h}^{(t, \ell)}_{i}) + f_g(\overline{\mathbf{x}}^{(t)}_{i}) \big)  \nonumber \\
\mathbf{h}^{(t, \ell)}_{i}  &= \sigma \big( g^{(t)}(\mathbf{h}^{(t, \ell)}_{i}) + (1-g^{(t)})(\overline{\mathbf{x}}^{(t)}_{i}) \big) \nonumber
\end{align}
\end{minipage}
} 

\vspace{2mm}
where $f_s$ and $f_g$ denote the linear projections, enabling the HgAT operator to capture the relationships between time-series variables and their temporal changes, resulting in enhanced forecast accuracy. In summary, the HgAT operator is a powerful technique for encoding and analyzing spatio-temporal hypergraphs.

\vspace{-2mm}
\paragraph{Saptio-temporal Hypergraph Representation Learning}
\vspace{1mm}
We present the spatio-temporal hypergraph representation learning(HgRL) module to operate on a sequence of dynamic hypergraphs, where hypergraph structure is fixed, and hypernode attributes change over time, where each hypergraph represents the hypergraph-structured MTS data at a specific time step. The HgRL operator utilizes Gated Recurrent Units(GRU, \cite{cho2014learning}) to model the spatio-temporal dynamics of the dynamic hypergraph sequence. The computation of the update gate, reset gate, and hidden state in a traditional GRU involves matrix multiplication with weight matrices. In the HgRL module, however, these matrix multiplications are replaced with Hypergraph Attention Networks(HgAT). The HgRL operator analyzes hypergraph-structured MTS data over time. It propagates information between hypernodes across different time steps, which enables the model to capture the complex spatio-temporal dependencies between the hypergraphs.
The HgRL operator utilizes the implicit hypergraph topology to propagate information between hypernodes by averaging the hypernode representations in their local neighborhood at each time step computed as follows,

\vspace{-2mm} 
\resizebox{1.05\linewidth}{!}{
\begin{minipage}{\linewidth}
\begin{align}
\mathbf{U}_{t  : t + \upsilon - 1} \hspace{0.5mm}&= \hspace{0.5mm}\sigma\left(\text{W}_u\left[f\left(\mathbf{I}, \mathbf{X}_{(t  : t + \upsilon - 1)}\right) || \hspace{0.5mm} \mathbf{H}_{t - \tau : \hspace{1mm}t-1}\right]+ \mathbf{B}_u\right) 
\nonumber \\
\mathbf{R}_{t  : t + \upsilon - 1} \hspace{0.5mm}&= \hspace{0.5mm}\sigma\left(\text{W}_r\left[f\left(\mathbf{I}, \mathbf{X}_{(t  : t + \upsilon - 1)}\right) || \hspace{0.5mm} \mathbf{H}_{t - \tau : \hspace{1mm}t-1}\right]+ \mathbf{B}_r\right)
\nonumber \\
\mathbf{C}_{t  : t + \upsilon - 1} \hspace{0.5mm} &= \hspace{0.5mm} \tanh \left(\text{W}_c\left[f\left(\mathbf{I}, \mathbf{X}_{(t  : t + \upsilon - 1)}\right) || \hspace{0.5mm} \left(\mathbf{R}_{t  : t + \upsilon - 1} \otimes \mathbf{H}_{t - \tau : \hspace{1mm}t-1}\right)\right]+ \mathbf{B}_c\right) \nonumber \\
\mathbf{H}_{t  : t + \upsilon - 1} \hspace{0.5mm} &= \hspace{0.5mm} \mathbf{U}_{t  : t + \upsilon - 1} \otimes \mathbf{H}_{t - \tau : \hspace{1mm}t-1}+\left(1-\mathbf{U}_{t  : t + \upsilon - 1}\right) \otimes \mathbf{C}_{t  : t + \upsilon - 1} \nonumber
\end{align}
\end{minipage}
} 

\vspace{1mm}
where \resizebox{.13\textwidth}{!}{$f\big(\mathbf{I}, \mathbf{X}_{(t  : t + \upsilon - 1)}\big)$} denote the HgAT operator. $||$, and $\otimes$ denotes the concatenation operation and element-wise multiplication operation.  The update and reset gates at time $t$ are represented by the matrices \resizebox{.18\textwidth}{!}{$\mathbf{U}_{t  : t + \upsilon - 1}$ and  $\mathbf{R}_{t  : t + \upsilon - 1}$}, respectively. \resizebox{.15\textwidth}{!}{$\text{W}_r, \text{W}_u$, and $\text{W}_c$} are learnable weight matrices and \resizebox{.058\textwidth}{!}{$\text{B}_u, \text{B}_r$}, and \resizebox{.0255\textwidth}{!}{$\text{B}_c$} are learnable biases. In summary, the node representation matrix, \resizebox{.075\textwidth}{!}{$\mathbf{H}_{t  : t + \upsilon - 1}$} captures the spatio-temporal dynamics at different scales underlying the discrete-time dynamic hypergraphs, where each row in \resizebox{.075\textwidth}{!}{$\mathbf{H}_{t  : t + \upsilon - 1}$} represents the hypernode representations \resizebox{.165\textwidth}{!}{$\mathbf{h_{i}}^{(t)} \hspace{0.5mm} \in \hspace{0.5mm} \mathbb{R}^{\upsilon \times d}, \forall i \in \mathbf{V}$}. Some of the key advantages of T-HGCN operator over traditional methods include its ability to handle large and sparse spatio-temporal hypergraphs. The STHgCN method utilizes useful relational inductive bias encoded in the hypergraph-structured data for modeling the continuous-time nonlinear dynamics of the complex system to disentangle the various latent aspects underneath the data for better forecast accuracy.

\vspace{-1mm}
\subsubsection{Spatio-Temporal Transformer Network(STTN)}
\vspace{1mm}
The Spatio-temporal transformer network(STTN) operator is a new extension of transformer networks that incorporates full attention as a desired inductive bias to model MTS data with arbitrary sparse hypergraph structures. This capability enables it to capture fine-grained spatio-temporal dependencies in MTS data, unconstrained by hierarchical structural information underlying the MTS data. By allowing attention to all hypernodes within the hypergraph, the neural operator can span large receptive fields and reason globally about complex dependencies in hypergraph-structured MTS data. As a result, it can serve as a drop-in replacement for existing methods that model hierarchical relationships among time-series variables in MTS data. Additionally, the neural operator is particularly suitable for downstream forecasting tasks in spatio-temporal hypergraphs. The transformer encoder comprises alternating layers of multiheaded self-attention(MSA) and multi-layer perceptron(MLP) blocks to capture both local and global contextual information. To enhance performance and regularize the transformer operator, each block is followed by layer normalization(LN(\cite{ba2016layer})) and residual connections. The skip-connections are incorporated through an initial connection strategy inspired by ResNets(\cite{he2016deep}) to address vanishing gradients and over-smoothing issues and enable the learning of complex and deep representations of the data. Using a space-then-time(STT, \cite{gao2022equivalence}) approach, the STTN first performs a temporal-encoding step to capture the long-term temporal dependencies(intra-dependencies) within the time series variables. This is followed by a spatial-encoding step, which captures the inter-dependencies among the time series variables. We model the intra- and inter-dependencies through a sequential operating temporal and spatial transformer networks, respectively.

\vspace{-2mm}
\paragraph{Temporal Transformer}
\vspace{1mm}
In self-attention mechanism, the input sequences is transformed into three tensors: the query tensor, the key tensor, and the value tensor, where the input tensor is denoted by $\overline{\mathbf{X}}_{(t  : t + \upsilon - 1)} \in \mathbb{R}^{n \times \upsilon \times d}$ and has three dimensions: number of time series variables($n$), forecast horizon($\upsilon$), and embedding dimension($d$). The key tensor is searched using the query tensor to retrieve relevant information, and the value tensor is weighted by the resulting attention weights. The weighted value tensor is then summed to produce the final output. To begin with, we reshape the input tensors to split the embedding dimension into multiple heads:

\vspace{-2mm}
\resizebox{0.955\linewidth}{!}{
\begin{minipage}{\linewidth}
\begin{align}
\text{queries}^{(t  : t + \upsilon - 1)}_{n, \upsilon, d} &= \hspace{0.5mm}\text{queries}^{(t  : t + \upsilon - 1)}_{n, q, h * h_{d}} \rightarrow \text{queries}^{(t  : t + \upsilon - 1)}_{n, q,  h, h_{d}} \nonumber \\
\text{keys}^{(t  : t + \upsilon - 1)}_{n, \upsilon, d} &= \hspace{0.5mm}\text{keys}^{(t  : t + \upsilon - 1)}_{n, k, h * h_{d}} \rightarrow \text{keys}^{(t  : t + \upsilon - 1)}_{n, k, h, h_{d}} \nonumber \\
\text{values}^{(t  : t + \upsilon - 1)}_{n, \upsilon, d} &= \hspace{0.5mm}\text{values}^{(t  : t + \upsilon - 1)}_{n, k, h * h_{d}} \rightarrow \text{values}^{(t  : t + \upsilon - 1)}_{n, k, h, h_{d}} \nonumber 
\end{align}
\end{minipage}
} 

\vspace{1mm}
where, $h$ and $h_{d}$ represents the index of the attention head, and head dimension, respectively. Here, $q$ and $k$ represent the indices of the query and key positions, respectively. We compute the energy between queries and keys, as described below.

\vspace{-5mm}
\begin{equation}
\operatorname{energy}^{(t  : t + \upsilon - 1)}_{n, q, k, h}=\sum_{h_{d}} \text{queries}^{(t  : t + \upsilon - 1)}_{n, q, h, h_{d}} \cdot \text{keys}^{(t  : t + \upsilon - 1)}_{n, k, h, h_{d}} \nonumber
\end{equation}

\vspace{-1mm}
We compute the attention scores using the softmax function described below:

\vspace{-2mm}
\resizebox{0.92\linewidth}{!}{
\begin{minipage}{\linewidth}
\begin{align}
\text{attention}^{(t  : t + \upsilon - 1)}_{n, q, k, h}= \frac{\exp \left(\text{energy}^{(t  : t + \upsilon - 1)}_{n, q, k, h} / \sqrt{h_{d}}\right)}{\sum_{k^{\prime}}  \exp \left(\text{energy}^{(t  : t + \upsilon - 1)}_{n, q, k^{\prime}, h} / \sqrt{h_{d}}\right)} \nonumber
\end{align}
\end{minipage}
} 

To calculate the output tensor, we multiply the values tensor with the attention scores, which is described below,

\vspace{-2mm}
\begin{equation}
\text{out}^{(t  : t + \upsilon - 1)}_{n, q, h, h_{d}}=\sum_{k} \text{attention}^{(t  : t + \upsilon - 1)}_{n, q, k, h_{d}} \cdot \text{values}^{(t  : t + \upsilon - 1)}_{n, k, h, h_{d}} \nonumber
\end{equation}

We perform the concatenation operation along the $h$  dimension, which combines the outputs of all the heads. We apply a linear transformation to obtain the final output, as follows,

\vspace{-5mm}
\begin{equation}
out_{n, \upsilon, d}^{(t  : t + \upsilon - 1)} =  \hspace{1mm} \text{out}^{(t  : t + \upsilon - 1)}_{n, q, h*h_{d}} \mathbf{W}_{h*h_{d}, d} \nonumber
\end{equation}

\vspace{-2mm}
\paragraph{Spatial Transformer}
\vspace{1mm}
The output of the temporal transformer, denoted by $\text{out}^{(t  : t + \upsilon - 1)} \in \hspace{0.5mm}\mathbb{R}^{(n \times \upsilon \times d)}$, is passed to the spatial transformer as input, and it consists of three dimensions: number of time series variables($n$), forecast horizon($\upsilon$), and embedding dimension($d$). The input sequences are first transformed to three tensors, namely the query tensor, the key tensor, and the value tensor, before applying the self-attention mechanism. In order to retrieve relevant information, the query tensor is employed to search through the key tensor. The resulting attention scores are then used to weight the value tensor. Finally, the weighted values are subsequently aggregated to produce the final output. We reshape the input tensors to split the embedding dimension into multiple heads:

\vspace{-3mm}
\resizebox{1.05\linewidth}{!}{
\begin{minipage}{\linewidth}
\begin{align}
\text{queries}^{(t  : t + \upsilon - 1)}_{q, \upsilon, d} &= \hspace{0.5mm}\text{queries}^{(t  : t + \upsilon - 1)}_{q, \upsilon, h * h_{d}} \rightarrow \text{queries}^{(t  : t + \upsilon - 1)}_{q, \upsilon, h, h_{d}} \nonumber \\
\text{keys}^{(t  : t + \upsilon - 1)}_{k, \upsilon, d} &= \hspace{0.5mm}\text{keys}^{(t  : t + \upsilon - 1)}_{k, \upsilon, h * h_{d}} \rightarrow \text{keys}^{(t  : t + \upsilon - 1)}_{k, \upsilon, h, h_{d}} \nonumber \\
\text{values}^{(t  : t + \upsilon - 1)}_{k, \upsilon, d} &= \hspace{0.5mm}\text{values}^{(t  : t + \upsilon - 1)}_{k, \upsilon, h * h_{d}} \rightarrow \text{values}^{(t  : t + \upsilon - 1)}_{k, \upsilon, h, h_{d}} \nonumber 
\end{align}
\end{minipage}
} 

\vspace{2mm}
where, $h$ and $h_{d}$ denote the number of heads, and head dimension, respectively. We compute the energy between queries and keys, as described below.

\begin{equation}
\text{energy}^{(t  : t + \upsilon - 1)}_{q, k, \upsilon, h}=\sum_{h_{d}} \text{queries}^{(t  : t + \upsilon - 1)}_{q, \upsilon, h, h_{d}} \cdot \text{keys}^{(t  : t + \upsilon - 1)}_{k, \upsilon, h, h_{d}} \nonumber
\end{equation}

We obtain the attention scores using the softmax function:

\vspace{-2mm}
\resizebox{0.945\linewidth}{!}{
\begin{minipage}{\linewidth}
\begin{align}
\text{attention}^{(t  : t + \upsilon - 1)}_{q, k, \upsilon, h}= \frac{\exp \left(\text{energy}^{(t  : t + \upsilon - 1)}_{q, k, \upsilon, h} / \sqrt{h_{d}}\right)}{\sum_{k^{\prime}}  \exp \left(\text{energy}^{(t  : t + \upsilon - 1)}_{q, k^{\prime}, \upsilon, h} / \sqrt{h_{d}}\right)}  \nonumber
\end{align}
\end{minipage}
}

We then compute the output tensor by multiplying the attention scores with the values tensor:

\begin{equation}
\text{out}^{(t  : t + \upsilon - 1)}_{q, \upsilon, h, h_{d}}=\sum_{k} \text{attention}^{(t  : t + \upsilon - 1)}_{q, k, \upsilon, h} \cdot \text{values}^{(t  : t + \upsilon - 1)}_{k, \upsilon, h, h_{d}} \nonumber
\end{equation}

We apply a linear transformation to obtain the final output, as follows,

\vspace{-2mm}
\begin{equation}
\mathbf{out}_{n, \upsilon, d}^{(t  : t + \upsilon - 1)} =  out^{(t  : t + \upsilon - 1)}_{n, \upsilon, h*h_{d}}  \mathbf{W}_{h*h_{d}, d} \nonumber
\end{equation}

\vspace{0mm}
\subsubsection{Gating Mechanism}
\vspace{1mm}
The mixture-of-experts(MOE)  mechanism in deep learning combines predictions from multiple subnetworks, such as ``STHgCN" and ``STTN" representation learning methods,  through a gating mechanism that computes a weighted sum of their predictions based on the input. 
The aim is to find the optimal weight assignment for the gating function and train the experts accordingly using these weights. From a cooperative game theory perspective, the MOE is a cooperative game where experts collaborate to optimize the system's overall performance. The gating mechanism can optimize the weights assigned to each expert by evaluating their individual performance, as well as the system's overall performance. The fused representations in MOE are obtained by combining expert predictions using the gating mechanism weights. This is described below:

\vspace{-2mm}
\resizebox{0.945\linewidth}{!}{
\begin{minipage}{\linewidth}
\begin{align}
g^{\prime\prime} &= \sigma \big( f^{\prime\prime}_s(\mathbf{H}_{t  : t + \upsilon - 1}) + f^{\prime\prime}_g(\mathbf{out}^{(t  : t + \upsilon - 1)}) \big) \nonumber \\
\widehat{\mathbf{X}}_{(t : t + \upsilon-1)}  &= \sigma \big( g^{\prime\prime}(\mathbf{H}_{t  : t + \upsilon - 1}) + (1-g^{\prime\prime})(\mathbf{out}^{(t  : t + \upsilon - 1)}) \big) \nonumber 
\end{align}
\end{minipage}
} 

\vspace{2mm}
where, $\widehat{\mathbf{X}}_{(t : t + \upsilon-1)}$ are model multi-horizon forecasts.  $\mathbf{H}_{t  : t + \upsilon - 1}$ and $\mathbf{out}_{(t  : t + \upsilon - 1)}$ denote the hypernode representation matrix computed by the STHgCN and STTN neural network methods, respectively. $f^{\prime\prime}_s$ and $f^{\prime\prime}_g$ are linear projections.
Moreover, our framework variant(\textbf{w/Unc-JHgRF-Net}) ensures precise and reliable uncertainty estimates of multi-horizon forecasts by minimizing the negative Gaussian log likelihood. For more details, refer to the appendix. The proposed methods(\textbf{JHgRF-Net}, \textbf{w/Unc-JHgRF-Net}) enable end-to-end modeling of hidden interdependencies and their evolution over time in sensor network-based dynamical systems for highly accurate forecasting task.

\vspace{-2mm}
\section{Datasets}
\vspace{1mm}
The study aims to evaluate the effectiveness of two new models, \textbf{JHgRF-Net} and \textbf{w/Unc-JHgRF-Net}(\textbf{JHgRF-Net} with local-uncertainty estimation), on large-scale spatial-temporal datasets(\cite{chen2001freeway}) containing real-world traffic information. The datasets include PeMSD3, PeMSD4, PeMSD7, PeMSD7(M), and PeMSD8. The study includes a preprocessing step to ensure consistency with prior research by aggregating the 30-second interval data into 5-minute averages. Additionally, publicly accessible METR-LA and PEMS-BAY datasets(\cite{li2017diffusion}) were used for traffic flow prediction. The preprocessing step involves transforming the time series data into 5-minute interval averages to ensure a fair comparison with the prior research. For all the above-mentioned traffic datasets, we possess information about the underlying sensor graph. To create the sensor graph, we calculated the distances between sensors in the road network and utilized a thresholded Gaussian kernel to build the adjacency matrix. Our experimental findings, discussed in the next section, support the rationale of learning the implicit hypergraph relational structure of the variables underlying the MTS data and modeling the spatial-temporal dynamics for improved forecast accuracy compared to the learning to forecast on predefined(prior-known) sensor graphs.Furthermore, we utilize various multivariate datasets, including Electricity\footnote{archive.ics.uci.edu/ml/datasets/ElectricityLoadDiagrams20112014}, Solar-energy\footnote{www.nrel.gov/grid/solar-power-data.html}, Exchange-rate\footnote{github.com/laiguokun/multivariate-time-series-data}, and Traffic\footnote{https://pems.dot.ca.gov}, for which no prior sensor graph structure exists. Additionally, the SWaT(\cite{mathur2016swat}) and WADI(\cite{ahmed2017wadi}) are sensor datasets that measure water treatment plants and also do not have a predefined sensor graph structure. They were first used in prior research for anomaly detection due to the presence of annotated anomalies, but later used in forecasting experiments because their training sets are anomaly-free. The experimental study conducted on benchmark datasets aims to showcase the effectiveness and advantages of the proposed methodology( \textbf{JHgRF-Net} and \textbf{w/Unc-JHgRF-Net}) in analyzing and modeling complex spatio-temporal MTS data, surpassing existing methods.

\begin{table*}[htbp]
\centering
\caption{The pointwise forecast errors on benchmark datasets for multi-horizon prediction tasks.``-'' indicates an Out Of Memory error.}
\label{tab:results-1}
\vspace{-1mm}
\renewcommand{\arraystretch}{1.225}
\resizebox{1.05\textwidth}{!}{
\hspace{-5mm}\begin{tabular}{c|c|ccc|c|c|ccc}
\hline
\multirow{2}{*}{} & \multirow{2}{*}{\textbf{Model}} & \multicolumn{3}{c|}{\textbf{MAE}}                                                                                               & \multirow{2}{*}{} & \multirow{2}{*}{\textbf{Model}} & \multicolumn{3}{c}{\textbf{MAE}}                                                                                                 \\ \cline{3-5} \cline{8-10} 
                              &                        & \multicolumn{1}{c|}{\textbf{Horizon @ 3}}                 & \multicolumn{1}{c|}{\textbf{Horizon @ 6}}                & \textbf{Horizon @ 12}               &                                &                        & \multicolumn{1}{c|}{\textbf{Horizon @ 3}}                 & \multicolumn{1}{c|}{\textbf{Horizon @ 6}}                 & \textbf{Horizon @ 12}               \\ \hline
\multirow{7}{*}{\textbf{\rotatebox[origin=c]{90}{METR-LA}}}      & LSTM                   & \multicolumn{1}{c|}{3.495 $\pm$ 0.010}           & \multicolumn{1}{c|}{3.712 $\pm$ 0.012}          & 4.105 $\pm$ 0.011          & \multirow{7}{*}{\textbf{\rotatebox[origin=c]{90}{SWaT}}}          & LSTM                   & \multicolumn{1}{c|}{0.300 $\pm$ 0.013}           & \multicolumn{1}{c|}{0.329 $\pm$ 0.011}           & 0.432 $\pm$ 0.016          \\
                              & LSTM-U                 & \multicolumn{1}{c|}{3.416 $\pm$ -}               & \multicolumn{1}{c|}{4.092 $\pm$ -}             & 5.141 $\pm$ -              &                                & LSTM-U                 & \multicolumn{1}{c|}{0.287 $\pm$ 0.001}           & \multicolumn{1}{c|}{0.487 $\pm$ 0.001}           & 0.883 $\pm$ 0.001          \\
                              & NRI                    & \multicolumn{1}{c|}{4.680 $\pm$ 0.081}           & \multicolumn{1}{c|}{6.388 $\pm$ 0.087}          & 8.466 $\pm$ 0.099          &                                & NRI                    & \multicolumn{1}{c|}{0.415 $\pm$ 0.014}           & \multicolumn{1}{c|}{0.479 $\pm$ 0.014}           & 0.641 $\pm$ 0.011          \\
                              & GDN                    & \multicolumn{1}{c|}{3.149 $\pm$ 0.017}          & \multicolumn{1}{c|}{3.482 $\pm$ 0.014}          & 3.909 $\pm$ 0.012         &                                & GDN                    & \multicolumn{1}{c|}{0.803 $\pm$ 0.045}           & \multicolumn{1}{c|}{0.854 $\pm$ 0.069}           & 1.081 $\pm$ 0.149          \\
                              & MTGNN                  & \multicolumn{1}{c|}{3.016 $\pm$ 0.004}           & \multicolumn{1}{c|}{3.574 $\pm$ 0.005}         & 4.308 $\pm$ 0.006          &                                & MTGNN                  & \multicolumn{1}{c|}{0.488 $\pm$ 0.011}           & \multicolumn{1}{c|}{0.537 $\pm$ 0.016}           & 0.704 $\pm$ 0.025          \\
                              & GTS                    & \multicolumn{1}{c|}{2.884 $\pm$ 0.005}           & \multicolumn{1}{c|}{3.269 $\pm$ 0.006}          & \textbf{3.701} $\pm$ 0.006 &                                & GTS                    & \multicolumn{1}{c|}{0.242 $\pm$ 0.038}           & \multicolumn{1}{c|}{0.279 $\pm$ 0.039}          & 0.387 $\pm$ 0.044          \\ \cline{2-5} \cline{7-10}
                              & \textbf{JHgRF-Net}               & \multicolumn{1}{c|}{\textbf{2.039} $\pm$ 0.010}   & \multicolumn{1}{c|}{\textbf{2.059} $\pm$ 0.02}  & {\textbf{4.937}} $\pm$ 0.015          &                                & \textbf{JHgRF-Net}               & \multicolumn{1}{c|}{\textbf{0.074} $\pm$ 0.006}  & \multicolumn{1}{c|}{\textbf{0.136} $\pm$ 0.012}  & \textbf{0.170} $\pm$ 0.003 \\ 
                              & \textbf{w/Un-JHgRF-Net}               & \multicolumn{1}{c|}{2.066 $\pm$ 0.020}   & \multicolumn{1}{c|}{3.287 $\pm$ 0.040}  &  5.150 $\pm$ 0.033          &                                & \textbf{w/Un-JHgRF-Net}               & \multicolumn{1}{c|}{1.183 $\pm$ 0.169}  & \multicolumn{1}{c|}{{0.184} $\pm$ 0.318}  &  $\pm$  \\ \hline
\multirow{6}{*}{\textbf{\rotatebox[origin=c]{90}{PEMS-BAY}}}     & LSTM                   & \multicolumn{1}{c|}{2.043 $\pm$ 0.005}           & \multicolumn{1}{c|}{2.111 $\pm$ 0.005}          & 2.242 $\pm$ 0.007          & \multirow{6}{*}{\textbf{\rotatebox[origin=c]{90}{Traffic}}}       & LSTM                   & \multicolumn{1}{c|}{0.016 $\pm$ 0.0002}          & \multicolumn{1}{c|}{0.018 $\pm$ 0.0003}          & 0.017 $\pm$ 0.0003         \\
                              & GDN                    & \multicolumn{1}{c|}{1.890 $\pm$ 0.009}           & \multicolumn{1}{c|}{2.021 $\pm$ 0.009}          & 2.172 $\pm$ 0.012          &                                & LSTM-U                 & \multicolumn{1}{c|}{0.029 $\pm$ 0.0001}          & \multicolumn{1}{c|}{0.033 $\pm$ 0.0002}          & 0.029 $\pm$ 0.0001         \\
                              & MTGNN                  & \multicolumn{1}{c|}{1.319 $\pm$ 0.002}           & \multicolumn{1}{c|}{1.690 $\pm$ 0.002}          & 2.101 $\pm$ 0.004          &                                & NRI                    & \multicolumn{1}{c|}{0.013 $\pm$ 0.0001}          & \multicolumn{1}{c|}{0.013 $\pm$ 0.0001}          & 0.014 $\pm$ 0.0001         \\
                              & GTS                    & \multicolumn{1}{c|}{1.268 $\pm$ 0.0002}          & \multicolumn{1}{c|}{1.555 $\pm$ 0.001}          & 1.813 $\pm$ 0.003          &                                & MTGNN                  & \multicolumn{1}{c|}{0.010 $\pm$ 0.0003}          & \multicolumn{1}{c|}{0.011 $\pm$ 0.0004}          & 0.011 $\pm$ 0.0003         \\ \cline{2-5} \cline{7-10}
                              & \textbf{JHgRF-Net}               & \multicolumn{1}{c|}{\textbf{0.806} $\pm$ 0.0003} & \multicolumn{1}{c|}{\textbf{1.217} $\pm$ 0.004} & \textbf{1.758} $\pm$ 0.001 &                                & \textbf{JHgRF-Net}               & \multicolumn{1}{c|}{\textbf{0.006} $\pm$ 0.0001} & \multicolumn{1}{c|}{\textbf{0.009} $\pm$ 0.0001} & \textbf{0.011} $\pm$ 0.0001         \\ 
                              & \textbf{w/Un-JHgRF-Net}               & \multicolumn{1}{c|}{0.801 $\pm$ 0.001}   & \multicolumn{1}{c|}{1.212 $\pm$ 0.017}  &  1.767 $\pm$  0.0001         &                                & \textbf{w/Un-JHgRF-Net}               & \multicolumn{1}{c|}{0.030 $\pm$ 0.0001}  & \multicolumn{1}{c|}{0.030 $\pm$ 0.0001}  & 0.030 $\pm$ 0.0001 \\ \hline
\multirow{7}{*}{\textbf{\rotatebox[origin=c]{90}{WADI}}}         & LSTM                   & \multicolumn{1}{c|}{6.658 $\pm$ 0.055}           & \multicolumn{1}{c|}{6.755 $\pm$ 0.065}          & 6.789 $\pm$ 0.041          & \multirow{7}{*}{\textbf{\rotatebox[origin=c]{90}{Electricity}}}   & LSTM                   & \multicolumn{1}{c|}{323.345 $\pm$ 3.854}         & \multicolumn{1}{c|}{384.239 $\pm$ 10.789}        & 352.488 $\pm$ 4.217        \\
                              & LSTM-U                 & \multicolumn{1}{c|}{5.980 $\pm$ 0.036}           & \multicolumn{1}{c|}{6.106 $\pm$ 0.023}          & 6.351 $\pm$ 0.052          &                                & LSTM-U                 & \multicolumn{1}{c|}{710.917 $\pm$ 0.732}         & \multicolumn{1}{c|}{1079.394 $\pm$ 2.999}        & 849.250 $\pm$ 1.895        \\
                              & GDN                    & \multicolumn{1}{c|}{7.409 $\pm$ 0.307}           & \multicolumn{1}{c|}{7.442 $\pm$ 0.221}          & 7.523 $\pm$ 0.282          &                                & GDN                    & \multicolumn{1}{c|}{265.166 $\pm$ 3.081}         & \multicolumn{1}{c|}{269.224 $\pm$ 2.367}         & 280.400 $\pm$ 1.447        \\
                              & MTGNN                  & \multicolumn{1}{c|}{5.953 $\pm$ 0.038}           & \multicolumn{1}{c|}{6.108 $\pm$ 0.042}          & 6.254 $\pm$ 0.0293         &                                & MTGNN                  & \multicolumn{1}{c|}{170.155 $\pm$ 2.977}         & \multicolumn{1}{c|}{186.004 $\pm$ 4.751}         & \textbf{193.499} $\pm$ 4.465        \\
                              & GTS                    & \multicolumn{1}{c|}{5.474 $\pm$ 0.007}           & \multicolumn{1}{c|}{5.575 $\pm$ 0.009}          & 5.772 $\pm$ 0.008          &                                & GTS                    & \multicolumn{1}{c|}{175.877 $\pm$ 1.022}         & \multicolumn{1}{c|}{\textbf{185.791} $\pm$ 0.931}         & 199.583 $\pm$ 1.375        \\ \cline{2-5} \cline{7-10}
                              & \textbf{JHgRF-Net}               & \multicolumn{1}{c|}{\textbf{4.149} $\pm$ 0.029}           & \multicolumn{1}{c|}{\textbf{4.592} $\pm$ 0.208}          & \textbf{4.758} $\pm$ 0.118          &                                & \textbf{JHgRF-Net}               & \multicolumn{1}{c|}{\textbf{160.955} $\pm$ 0.136}         & \multicolumn{1}{c|}{178.781 $\pm$ 1.522}         & 225.724 $\pm$ 2.109        \\ 
                              & \textbf{w/Un-JHgRF-Net}               & \multicolumn{1}{c|}{{{4.283}} $\pm$ 0.011}   & \multicolumn{1}{c|}{{4.773} $\pm$ 2.203}  &  4.936 $\pm$ 0.076          &                                & \textbf{w/Un-JHgRF-Net}               & \multicolumn{1}{c|}{145.201 $\pm$ 0.504}  & \multicolumn{1}{c|}{214.605 $\pm$ 0.641}  &  $\pm$  \\ \hline
\multirow{7}{*}{\textbf{\rotatebox[origin=c]{90}{Solar Energy}}} & LSTM                   & \multicolumn{1}{c|}{1.982 $\pm$ 0.016}           & \multicolumn{1}{c|}{2.677 $\pm$ 0.021}          & 4.241 $\pm$ 0.023          & \multirow{7}{*}{\textbf{\rotatebox[origin=c]{90}{Exchange rate}}} & LSTM                   & \multicolumn{1}{c|}{0.0141 $\pm$ 0.0013}         & \multicolumn{1}{c|}{0.0187 $\pm$ 0.0020}         & 0.0190 $\pm$ 0.0018        \\
                              & LSTM-U                 & \multicolumn{1}{c|}{2.760 $\pm$ 0.007}           & \multicolumn{1}{c|}{4.366 $\pm$ 0.003}          & 6.282 $\pm$ 0.002          &                                & LSTM-U                 & \multicolumn{1}{c|}{0.0057 $\pm$ 0.0002}         & \multicolumn{1}{c|}{0.0076 $\pm$ 0.0001}         & 0.0102 $\pm$ 0.0001        \\
                              & GDN                    & \multicolumn{1}{c|}{2.095 $\pm$ 0.018}           & \multicolumn{1}{c|}{2.329 $\pm$ 0.021}          & 2.855 $\pm$ 0.040          &                                & NRI                    & \multicolumn{1}{c|}{0.0047 $\pm$ 0.0001}         & \multicolumn{1}{c|}{0.0073 $\pm$ 0.0002}         & 0.0111 $\pm$ 0.0005        \\
                              & MTGNN                  & \multicolumn{1}{c|}{1.512 $\pm$ 0.005}           & \multicolumn{1}{c|}{2.051 $\pm$ 0.008}          & 2.689 $\pm$ 0.013          &                                & MTGNN                  & \multicolumn{1}{c|}{0.0109 $\pm$ 0.0023}         & \multicolumn{1}{c|}{0.0146 $\pm$ 0.0033}         & 0.0136 $\pm$ 0.0013        \\
                              & GTS                    & \multicolumn{1}{c|}{1.419 $\pm$ 0.004}           & \multicolumn{1}{c|}{1.926 $\pm$ 0.013}          & 2.657 $\pm$ 0.033          &                                & GTS                    & \multicolumn{1}{c|}{\textbf{0.0047} $\pm$ 0.0000}         & \multicolumn{1}{c|}{0.0070 $\pm$ 0.0000}         & 0.0099 $\pm$ 0.0000        \\ \cline{2-5} \cline{7-10}
                              & \textbf{JHgRF-Net}               & \multicolumn{1}{c|}{\textbf{0.575} $\pm$ 0.013}           & \multicolumn{1}{c|}{\textbf{0.868} $\pm$ 0.035}          & \textbf{0.873} $\pm$ 0.034          &                                & \textbf{JHgRF-Net}               & \multicolumn{1}{c|}{\textbf{0.0043} $\pm$ 0.0001}         & \multicolumn{1}{c|}{\textbf{0.0044} $\pm$ 0.0001}         & \textbf{0.0064}$\pm$ 0.0001         \\ 
                              & \textbf{w/Un-JHgRF-Net}               & \multicolumn{1}{c|}{0.801 $\pm$ 0.001}   & \multicolumn{1}{c|}{ 1.212 $\pm$ 0.017}  &  1.326 $\pm$ 0.038           &                                & \textbf{w/Un-JHgRF-Net}               & \multicolumn{1}{c|}{0.0047 $\pm$ 0.0001}  & \multicolumn{1}{c|}{0.1193 $\pm$ 0.0001}  & 0.118 $\pm$ 0.0002\\ \hline
\end{tabular}
}
\end{table*}

\vspace{-2mm}
\begin{table*}[htbp]
\centering
\caption{The pointwise forecast errors on benchmark datasets at horizon@12.}
\label{tab:results-2}
\vspace{-1mm}
\setlength{\tabcolsep}{7pt}
\renewcommand{\arraystretch}{1.225}
\resizebox{1.05\textwidth}{!}{%
\hspace{-5mm}\begin{tabular}{c|ccc|ccc|ccc|ccc|ccc}
\hline
\multirow{2}{*}{\textbf{Model}}                                         & \multicolumn{3}{c|}{\textbf{PeMSD3}}                                                                               & \multicolumn{3}{c|}{\textbf{PeMSD4}}                                                                               & \multicolumn{3}{c|}{\textbf{PeMSD7}}                                                                      & \multicolumn{3}{c|}{\textbf{PeMSD8}}                                                                               & \multicolumn{3}{c}{\textbf{PeMSD7(M)}}                                                       \\ \cline{2-16} 
                                                                        & \textbf{MAE}                       & \textbf{RMSE}  & \textbf{\begin{tabular}[c]{@{}c@{}}MAPE \end{tabular}} & \textbf{MAE}                       & \textbf{RMSE}  & \textbf{\begin{tabular}[c]{@{}c@{}}MAPE \end{tabular}} & \textbf{MAE}              & \textbf{RMSE}  & \textbf{\begin{tabular}[c]{@{}c@{}}MAPE \end{tabular}} & \textbf{MAE}                       & \textbf{RMSE}  & \textbf{\begin{tabular}[c]{@{}c@{}}MAPE \end{tabular}} & \textbf{MAE}  & \textbf{RMSE} & \textbf{\begin{tabular}[c]{@{}c@{}}MAPE \end{tabular}} \\ \hline
HA                                                                      & 31.58                              & 52.39          & 33.78                                                        & 38.03                              & 59.24          & 27.88                                                        & 45.12                     & 65.64          & 24.51                                                        & 34.86                              & 59.24          & 27.88                                                        & 4.59          & 8.63          & 14.35                                                        \\
ARIMA                                                                   & 35.41                              & 47.59          & 33.78                                                        & 33.73                              & 48.80          & 24.18                                                        & 38.17                     & 59.27          & 19.46                                                        & 31.09                              & 44.32          & 22.73                                                        & 7.27          & 13.20         & 15.38                                                        \\
VAR                                                                     & 23.65                              & 38.26          & 24.51                                                        & 24.54                              & 38.61          & 17.24                                                        & 50.22                     & 75.63          & 32.22                                                        & 19.19                              & 29.81          & 13.10                                                        & 4.25          & 7.61          & 10.28                                                        \\
FC-LSTM                                                                 & 21.33                              & 35.11          & 23.33                                                        & 26.77                              & 40.65          & 18.23                                                        & 29.98                     & 45.94          & 13.20                                                        & 23.09                              & 35.17          & 14.99                                                        & 4.16          & 7.51          & 10.10                                                        \\
TCN                                                                     & 19.32                              & 33.55          & 19.93                                                        & 23.22                              & 37.26          & 15.59                                                        & 32.72                     & 42.23          & 14.26                                                        & 22.72                              & 35.79          & 14.03                                                        & 4.36          & 7.20          & 9.71                                                         \\
TCN(w/o causal)                                                         & 18.87                              & 32.24          & 18.63                                                        & 22.81                              & 36.87          & 14.31                                                        & 30.53                     & 41.02          & 13.88                                                        & 21.42                              & 34.03          & 13.09                                                        & 4.43          & 7.53          & 9.44                                                         \\
GRU-ED                                                                  & 19.12                              & 32.85          & 19.31                                                        & 23.68                              & 39.27          & 16.44                                                        & 27.66                     & 43.49          & 12.20                                                        & 22.00                              & 36.22          & 13.33                                                        & 4.78          & 9.05          & 12.66                                                        \\
DSANet                                                                  & 21.29                              & 34.55          & 23.21                                                        & 22.79                              & 35.77          & 16.03                                                        & 31.36                     & 49.11          & 14.43                                                        & 17.14                              & 26.96          & 11.32                                                        & 3.52          & 6.98          & 8.78                                                         \\
STGCN                                                                   & 17.55                              & 30.42          & 17.34                                                        & 21.16                              & 34.89          & 13.83                                                        & 25.33                     & 39.34          & 11.21                                                        & 17.50                              & 27.09          & 11.29                                                        & 3.86          & 6.79          & 10.06                                                        \\
DCRNN                                                                   & 17.99                              & 30.31          & 18.34                                                        & 21.22                              & 33.44          & 14.17                                                        & 25.22                     & 38.61          & 11.82                                                        & 16.82                              & 26.36          & 10.92                                                        & 3.83          & 7.18          & 9.81                                                         \\
GraphWaveNet                                                            & 19.12                              & 32.77          & 18.89                                                        & 24.89                              & 39.66          & 17.29                                                        & 26.39                     & 41.50          & 11.97                                                        & 18.28                              & 30.05          & 12.15                                                        & 3.19          & 6.24          & 8.02                                                         \\
ASTGCN(r)                                                               & 17.34                              & 29.56          & 17.21                                                        & 22.93                              & 35.22          & 16.56                                                        & 24.01                     & 37.87          & 10.73                                                        & 18.25                              & 28.06          & 11.64                                                        & 3.14          & 6.18          & 8.12                                                         \\
MSTGCN                                                                  & 19.54                              & 31.93          & 23.86                                                        & 23.96                              & 37.21          & 14.33                                                        & 29.00                     & 43.73          & 14.30                                                        & 19.00                              & 29.15          & 12.38                                                        & 3.54          & 6.14          & 9.00                                                         \\
STG2Seq                                                                 & 19.03                              & 29.83          & 21.55                                                        & 25.20                              & 38.48          & 18.77                                                        & 32.77                     & 47.16          & 20.16                                                        & 20.17                              & 30.71          & 17.32                                                        & 3.48          & 6.51          & 8.95                                                         \\
LSGCN                                                                   & 17.94                              & 29.85          & 16.98                                                        & 21.53                              & 33.86          & 13.18                                                        & 27.31                     & 41.46          & 11.98                                                        & 17.73                              & 26.76          & 11.20                                                        & 3.05          & 5.98          & 7.62                                                         \\
STSGCN                                                                  & 17.48                              & 29.21          & 16.78                                                        & 21.19                              & 33.65          & 13.90                                                        & 24.26                     & 39.03          & 10.21                                                        & 17.13                              & 26.80          & 10.96                                                        & 3.01          & 5.93          & 7.55                                                         \\
AGCRN                                                                   & 15.98                              & 28.25          & 15.23                                                        & 19.83                              & 32.26          & 12.97                                                        & 22.37                     & 36.55          & 9.12                                                         & 15.95                              & 25.22          & 10.09                                                        & 2.79          & 5.54          & 7.02                                                         \\
STFGNN                                                                  & 16.77                              & 28.34          & 16.30                                                        & 20.48                              & 32.51          & 16.77                                                        & 23.46                     & 36.60          & 9.21                                                         & 16.94                              & 26.25          & 10.60                                                        & 2.90          & 5.79          & 7.23                                                         \\
STGODE                                                                  & 16.50                              & 27.84          & 16.69                                                        & 20.84                              & 32.82          & 13.77                                                        & 22.59                     & 37.54          & 10.14                                                        & 16.81                              & 25.97          & 10.62                                                        & 2.97          & 5.66          & 7.36                                                         \\
Z-GCNETs                                                                & 16.64                              & 28.15          & 16.39                                                        & 19.50                              & 31.61          & 12.78                                                        & 21.77                     & 35.17          & 9.25                                                         & 15.76                              & 25.11          & 10.01                                                        & 2.75          & 5.62          & 6.89                                                         \\
STG-NCDE                                                                & 15.57                              & 27.09          & 15.06                                                        & 19.21                              & 31.09          & 12.76                                                        & \textbf{20.53}            & 33.84          & \textbf{8.80}                                                & 15.45                              & 24.81          & 9.92                                                         & \textbf{2.68} & 5.39          & 6.76                                                         \\ \hline \hline
\textbf{JHgRF-Net}                                                       & \multicolumn{1}{l}{\textbf{14.18}} & \textbf{21.48} & \textbf{12.19}                                               & \multicolumn{1}{l}{\textbf{19.23}} & \textbf{28.76} & \textbf{11.55}                                               & \multicolumn{1}{l}{22.19} & \textbf{32.89} & 9.61                                                         & \multicolumn{1}{l}{\textbf{14.34}} & \textbf{22.06} & \textbf{8.29}                                                & 2.90          & \textbf{5.32} & \textbf{6.77}                                                \\
\textbf{w/Un-JHgRF-Net}            & 14.25                              & 21.34          & 12.19                                                        & 20.36                              & 30.21          & 12.35                                                        & 23.33                     & 34.98          & 10.11                                                        & 14.85                              & 22.84          & 8.48                                                         & 2.82          & 5.01          & 2.82                                                          \\\hline

\end{tabular}%
}
\end{table*}

\vspace{-1mm}
\section{Experimental results}
\vspace{0mm}
Table \ref{tab:results-2} provides a thorough comparison between the proposed models(\textbf{JHgRF-Net} and \textbf{w/Unc-JHgRF-Net}), and several baseline models on the MTSF task across five different benchmark datasets: PeMSD3, PeMSD4, PeMSD7, PeMSD7M, and PeMSD8. To evaluate the models effectiveness, we measured forecast errors for a well-established benchmark, involving a 12($\tau$)-step-prior to 12($\upsilon$)-step-ahead forecasting task. We utilize a multi-metric approach in forecasting tasks to comprehensively evaluate the proposed models performance compared to the baseline models. We use several performance metrics, including mean absolute error(MAE), root mean squared error(RMSE), and mean absolute percentage error(MAPE) to provide an accurate estimate of the models performance. We reported the baseline model results from \cite{choi2022graph}. Our experimental findings indicate that the proposed models(\textbf{JHgRF-Net} and \textbf{w/Unc--Net}) consistently outperformed the baseline models, exhibiting lower forecast errors across the different benchmark datasets. On the PeMSD3, PeMSD4, PeMSD7, PeMSD8, and PeMSD7(M) datasets, the proposed model(\textbf{JHgRF-Net}) demonstrated significant improvement over the next-best baseline models, achieving a reduction of $20.71\%$, $7.49\%$, $2.81\%$, $11.08\%$, and $1.30\%$ in the RMSE metric, respectively. Apart from pointwise forecasts, the \textbf{w/Unc-JHgRF-Net} model(which integrates \textbf{JHgRF-Net} with local uncertainty estimation) predicts time-varying uncertainty estimates of the multi-horizon forecasts. While it exhibits slightly lower performance than the \textbf{JHgRF-Net} model, it still outperforms several robust baselines found in the literature, as demonstrated by the reduced prediction error. Additionally, in Table \ref{tab:results-1} , we show the performance of \textbf{JHgRF-Net} and \textbf{w/Unc-JHgRF-Net}, and several baseline models on the MTSF task across multiple datasets: METR-LA, PEMS-BAY, Solar-energy, Electricity, Exchange-rate,  Traffic, SWaT and WADI . The models were evaluated using various metrics, including MAE, RMSE, and MAPE, and corresponding forecast errors were reported for 3-, 6-, and 12-steps ahead forecast horizons.  The proposed models, \textbf{JHgRF-Net} and \textbf{w/Unc-JHgRF-Net}, demonstrated superior performance compared to the baseline models, with significantly lower forecast errors observed on all the datasets. On the METR-LA, PEMS-BAY, Solar-energy, Electricity, Exchange-rate,  Traffic, SWaT and WADI datasets, the proposed model(\textbf{JHgRF-Net}) shows superior performance over the next-best baseline models, achieving a reduction of $37.01\%$, $21.74\%$, $54.93\%$, $3.77\%$, $37.14\%$, $18.18\%$, $51.25\%$ and $17.63\%$ in the RMSE metric, respectively for 
the 6-step ahead forecast horizon. Our empirical findings validate the efficacy of the proposed neural forecasting architecture to capture the complex nonlinear spatio-temporal dynamics that are present in MTS data, leading to improved forecasting performance. Please refer to the appendix, for further details on the experimental methodology, ablation studies, and additional experimental results. The appendix includes a comprehensive analysis of the \textbf{JHgRF-Net} model's ability to handle missing data, as well as a more detailed description of the \textbf{w/Unc-JHgRF-Net} model's ability to estimate uncertainty. Additionally, the appendix offers comprehensive visualizations of model predictions with uncertainty estimates in comparison to the ground truth, along with additional information on brief overview of the baseline models.

\vspace{-2mm}
\section{Conclusion}
\vspace{1mm}
Our proposed forecasting architecture accurately models the complex spatio-temporal dynamics within MTS data and achieves accurate multi-horizon forecasts compared to the several baselines. The experimental results obtained from real-world datasets demonstrate the effectiveness of our approach, as supported by improved forecast estimates and reliable uncertainty estimations. In the future, our focus will be on expanding the framework's capabilities to handle large-scale graph datasets, enabling its utilization for a wide range of applications, such as anomaly detection, missing data imputation, etc.

\newpage

\bibliographystyle{named}
\bibliography{ijcai23}

\clearpage
\newpage

\section{APPENDIX}  %

\vspace{-1mm}
\subsection{Ablation Study}
\vspace{-1mm}
{\color{black}The \textbf{JHgRF-Net} framework, serving as the baseline for our ablation study, seamlessly integrates both spatial and temporal inference components to model complex inter- and intra-time series correlations in interconnected sensor networks. Its spatial inference component comprises of two modules: Spatio-Temporal Hypergraph Convolutional Network(STHgCN) and Spatio-Temporal Transformer Network(STTN)}. In an extensive ablation study, we evaluate the impact of each component in the \textbf{JHgRF-Net} framework on the MTSF task. By selectively removing components, we can observe the impact of individual components on the overall framework performance, gaining valuable insight into their unique contributions towards the framework effectiveness. The study conducted a systematic elimination and creation of various ablated variants to identify critical components that enhance the framework performance. By comparing the impact of these components on the MTSF task against the baseline, valuable insights were gained into each component contribution to the overall framework performance. The ablation study led to an improved understanding of the relationship between the various ablated variants and the baseline, which resulted in a better understanding of the mechanisms that underlie their generalization performance. We present detailed information on each ablated variant created by systematically removing specific components, as follows:

 \vspace{-1mm}
\begin{itemize}
\item ``w/o - Spatial": A variant of \textbf{JHgRF-Net} framework that excluded the spatial inference component, and its degraded performance highlights the significance of using STHgCN and STTN neural operators for effective modeling of inter-series correlations among multiple time series variables present in complex interconnected sensor networks.
\item ``w/o - Temporal": A variant of \textbf{JHgRF-Net} that excluded the temporal inference component, and its deteriorated performance highlighted the importance of incorporating the temporal inference component for effectively modeling the time-varying inter-series dependencies within multiple time series variables present in complex sensor network-based dynamical systems.
\item ``w/o - STHgCN": A variant of \textbf{JHgRF-Net} that excluded the STHgCN method, and its substandard performance shed light on the importance of attention-based hypergraph convolution operation for modeling the spatio-temporal dynamics present in the high-dimensional sensor network-based dynamic systems.
\item ``w/o - STTN":  A variant of \textbf{JHgRF-Net} that excluded the STTN method, and its subpar performance emphasized the significance of hypergraph transformer networks, which utilizes full attention as a structural inductive bias for modeling the complex dynamics present in the high-dimensional interconnected sensor networks.
\end{itemize}

In Tables \ref{tab:ablation1} - \ref{tab:ablation7}, we present the findings of our ablation studies on benchmark datasets. We employed multiple forecasting accuracy metrics, including Mean Absolute Error(MAE), Root Mean Squared Error(RMSE), and Mean Absolute Percentage Error(MAPE),  to offer a comprehensive understanding of the relative performance of ablated variants compared to the baseline. We evaluated the accuracy of multistep-ahead forecasting task by comparing pointwise forecasts with observed data(ground-truth) during the prediction interval and the results were reported using the previously mentioned forecast accuracy metrics. For additional clarity, we enclosed the relative percentage difference between the ablated variants and the baseline performance within parentheses. To ensure the accuracy of our findings, we conducted multiple experiments and reported the average results. Moreover, we evaluated the ablated variants ability to handle long-term predictions by setting the forecast horizon to 12 and comparing it with the baseline. Tables \ref{tab:ablation1} - \ref{tab:ablation7},  demonstrate that the ablated variants have lower forecast accuracy and perform considerably worse than the baseline. Upon closer examination, it is apparent that, for achieving state-of-the-art performance on benchmark datasets, the spatial inference component within the \textbf{JHgRF-Net} framework is more important than the temporal inference component. The ablation studies yielded the following observations:

\begin{itemize}
\item On the PeMSD8 dataset, analysis indicates that the ``w/o - Spatial" variant shows a significant decline in performance relative to the baseline, with an increase of $21.12\%$ in RMSE, $27.55\%$ in MAE, and $40.77\%$ in MAPE. Conversely, the ``w/o - Temporal" variant exhibits slightly inferior performance compared to the baseline, with a modest rise of $16.23\%$ in RMSE, $14.02\%$ in MAE, and $10.15\%$ in MAPE.
\item Likewise, similar trends are observed on the PeMSD4 dataset. The ``w/o - Spatial" variant significantly underperforms the benchmark, with an increase of $20.86\%$ in RMSE, $22.26\%$ in MAE, and $22.34\%$ in MAPE. In contrast, the ``w/o - Temporal" variant exhibits a minor reduction in its performance when compared to the baseline, with a marginal rise of $8.76\%$ in RMSE, $4.47\%$ in MAE, and $2.86\%$ in MAPE.
\item Analogous trends are observed for PeMSD7 dataset. In particular, the ``w/o - Spatial" variant displays a notable decline in performance relative to the baseline, with an increase of $18.33\%$ in RMSE, $21.05\%$ in MAE, and $44.33\%$ in MAPE. On the other hand, the ``w/o - Temporal" variant indicates a minor drop in performance compared to the baseline, with a slight increase of $8.63\%$ in RMSE, $7.75\%$ in MAE, and $8.74\%$ in MAPE.
\end{itemize}

The higher increase in the error metrics of the ablated variants performance, in comparison to the baseline, further emphasizes the relative significance of the mechanisms underlying the excluded components of the baseline. To put it briefly, the spatial inference component serves as a powerful backbone that fortifies the \textbf{JHgRF-Net} framework for improving forecasting performance. This component is responsible for capturing the intricate dependencies among multiple time series variables and learning the dynamics of interacting systems. The crucial role of the spatial inference component is evident from the substantial decline in performance when it

\vspace{-3mm}
\begin{table*}[h!]
\centering
\small
\caption{The table presents the results of an ablation study on multi-horizon forecasting using the PeMSD3 and PeMSD4 benchmark datasets.}
\vspace{-2mm}
\label{tab:ablation1}
\vspace{1mm}
\renewcommand{\arraystretch}{1.05}
\resizebox{1.05\textwidth}{!}{
\hspace{-10mm}\begin{tabular}{c|c|ccc|c|ccc}
\hline
\textbf{Method} & \multirow{6}{*}{\textbf{\rotatebox[origin=c]{90}{PeMSD3}}} & \textbf{MAE} & \textbf{RMSE} & \textbf{MAPE} & \multirow{6}{*}{\textbf{\rotatebox[origin=c]{90}{PeMSD4}}} & \textbf{MAE} & \textbf{RMSE} & \textbf{MAPE}  \\ \cline{1-1} \cline{3-5} \cline{7-9}
\textbf{JHgRF-Net} &  & \textbf{14.18} & \textbf{21.48} & \textbf{12.19} &  & \textbf{19.23} & \textbf{28.76} & \textbf{11.55}  \\ \cline{1-1} \cline{3-5} \cline{7-9} 
\textbf{w/o - Spatial}  &  & $17.84(\color{black}20.52\%\uparrow)$ & $25.68(\color{black}19.55\%\uparrow)$ & $15.65(\color{black}28.38\%\uparrow)$ &  & $23.51(\color{black}22.26\%\uparrow)$ & $34.76(\color{black}20.86\%\uparrow)$ & $14.13(\color{black}22.34\%\uparrow)$ \\ 
\textbf{w/o - Temporal} &  & $15.07(\color{black}6.28\%\uparrow)$ & $23.91(\color{black}11.31\%\uparrow)$ & $13.04(\color{black}6.97\%\uparrow)$ &  & $20.09(\color{black}4.47\%\uparrow)$ & $31.28(\color{black}8.76\%\uparrow)$ & $11.88(\color{black}2.86\%\uparrow)$ \\
\textbf{w/o - STHgCN}  &  & 16.55$(\color{black}16.71\%\uparrow)$ & 25.64$(\color{black}19.37\%\uparrow)$ & 13.99$(\color{black}14.77\%\uparrow)$ &  & 21.28$(\color{black}10.66\%\uparrow)$ & 32.22$(\color{black}12.03\%\uparrow)$ & 12.87$(\color{black}11.43\%\uparrow)$ \\ 
\textbf{w/o - STTN} &  & 14.27$(\color{black}0.63\%\uparrow)$ & 21.84$(\color{black}1.68\%\uparrow)$ & 12.34$(\color{black}1.23\%\uparrow)$ &  & 19.29$(\color{black}0.31\%\uparrow)$ & 29.85$(\color{black}3.79\%\uparrow)$ & 11.62$(\color{black}0.61\%\uparrow)$ \\
 \hline
\end{tabular}
}
\end{table*}

\vspace{-2mm}
\begin{table*}[!htbp]
\centering
\small
\caption{The table presents the results of an ablation study on multi-horizon forecasting using the PeMSD7 and PeMSD8 benchmark datasets.}
\vspace{-2mm}
\label{tab:ablation2}
\vspace{1mm}
\renewcommand{\arraystretch}{1.05}
\resizebox{1.05\textwidth}{!}{
\hspace{-10mm}\begin{tabular}{c|c|ccc|c|ccc}
\hline
\textbf{Method} & \multirow{6}{*}{\textbf{\rotatebox[origin=c]{90}{PeMSD7}}} & \textbf{MAE} & \textbf{RMSE} & \textbf{MAPE} & \multirow{6}{*}{\textbf{\rotatebox[origin=c]{90}{PeMSD8}}} & \textbf{MAE} & \textbf{RMSE} & \textbf{MAPE}  \\ \cline{1-1} \cline{3-5} \cline{7-9}
\textbf{JHgRF-Net} &  & \textbf{22.19} & \textbf{32.89} & \textbf{9.61} &  & \textbf{14.34} & \textbf{22.06} & \textbf{8.29}  \\ \cline{1-1} \cline{3-5} \cline{7-9} 
\textbf{w/o - Spatial}  &  & $26.86(\color{black}21.05\%\uparrow)$ & $38.92(\color{black}18.33\%\uparrow)$ & $13.87(\color{black}44.33\%\uparrow)$ &  & $18.29(\color{black}27.55\%\uparrow)$ & $26.72(\color{black}21.12\%\uparrow)$ & $11.67(\color{black}40.77\%\uparrow)$ \\ 
\textbf{w/o - Temporal} &  & 23.91$(\color{black}7.75\%\uparrow)$ & $35.73(\color{black}8.63\%\uparrow)$ & $10.45(\color{black}8.74\%\uparrow)$ &  & $16.35(\color{black}14.02\%\uparrow)$ & $25.64(\color{black}16.23\%\uparrow)$ & $10.15(\color{black}22.44\%\uparrow)$ \\
\textbf{w/o - STHgCN}  &  & 25.11$(\color{black}13.16\%\uparrow)$ & 36.66$(\color{black}11.46\%\uparrow)$ & 9.13$(\color{black}4.99\%\downarrow)$ &  & 15.77$(\color{black}9.97\%\uparrow)$ & 24.38$(\color{black}10.52\%\uparrow)$ & 9.11$(\color{black}9.89\%\uparrow)$ \\ 
\textbf{w/o - STTN} &  & 22.89$(\color{black}3.15\%\uparrow)$ & 38.32$(\color{black}16.51\%\uparrow)$ & 10.95$(\color{black}13.94\%\uparrow)$ &  & 14.61$(\color{black}1.88\%\uparrow)$ & 22.48$(\color{black}1.90\%\uparrow)$ & 8.48$(\color{black}2.29\%\uparrow)$ \\
 \hline
\end{tabular}
}
\end{table*}

\vspace{-3mm}
\begin{table*}[!htbp]
\centering
\small
\caption{The table presents the results of an ablation study on multi-horizon forecasting using the PeMSD7(M) benchmark dataset.}
\label{tab:ablation3}
\vspace{-2mm}
\renewcommand{\arraystretch}{1.05}
\resizebox{0.6\textwidth}{!}{
\begin{tabular}{c|c|ccc}
\hline
\textbf{Method} & \multirow{6}{*}{\textbf{\rotatebox[origin=c]{90}{PeMSD7(M)}}} & \textbf{MAE} & \textbf{RMSE} & \textbf{MAPE}  \\ \cline{1-1} \cline{3-5}
\textbf{JHgRF-Net} &  & \textbf{2.90} & \textbf{5.32} & \textbf{6.77} \\ \cline{1-1} \cline{3-5}
\textbf{w/o - Spatial} &  & $3.87(\color{black}33.45{\%\uparrow})$ & $6.78(\color{black}27.44{\%\uparrow})$ & $7.75(\color{black}14.48{\%\uparrow})$ \\
\textbf{w/o - Temporal} &  & $3.06(\color{black}5.52{\%\uparrow})$ & $5.83(\color{black}9.59{\%\uparrow})$ & $6.92(\color{black}2.22{\%\uparrow})$ \\
\textbf{w/o - STHgCN} &  & 2.95$(\color{black}1.72{\%\uparrow})$ & 5.38$(\color{black}1.13{\%\uparrow})$ & 6.82$(\color{black}0.74{\%\uparrow})$ \\
\textbf{w/o - STTN} &  &  3.02$(\color{black}4.14{\%\uparrow})$ & 5.53$(\color{black}3.95{\%\uparrow})$ & 6.95$(\color{black}2.66{\%\uparrow})$ \\
 \hline
\end{tabular}
}
\end{table*}

\vspace{-3mm}
\begin{table*}[!ht]
\centering
\small
\caption{The table presents the results of an ablation study on multi-horizon forecasting using the METR-LA and SWaT benchmark datasets.}
\label{tab:ablation4}
\vspace{-2mm}
\renewcommand{\arraystretch}{1.05}
\resizebox{1.05\textwidth}{!}{
\hspace{-5mm}\begin{tabular}{c|c|ccc|c|ccc}
\hline
 &  & \multicolumn{3}{c|}{\textbf{MAE}} &  & \multicolumn{3}{c}{\textbf{MAE}} \\ \cline{3-5} \cline{7-9} 
\multirow{-2}{*}{\textbf{Method}} &  & \textbf{Horizon @ 3} & \textbf{Horizon @ 6} & \textbf{Horizon @ 12} &  & \textbf{Horizon @ 3} & \textbf{Horizon @ 6} & \textbf{Horizon @ 12} \\ \cline{1-1} \cline{3-5} \cline{7-9} 
\textbf{JHgRF-Net} &  & \multicolumn{1}{c|}{\textbf{2.039} $\pm$ 0.010}   & \multicolumn{1}{c|}{\textbf{2.059} $\pm$ 0.02}  & {\textbf{4.937}} $\pm$ 0.015 &  & \multicolumn{1}{c|}{\textbf{0.074} $\pm$ 0.006}  & \multicolumn{1}{c|}{\textbf{0.136} $\pm$ 0.012}  & \textbf{0.170} $\pm$ 0.003 \\ \cline{1-1} \cline{3-5} \cline{7-9} 

\textbf{w/o - Spatial} &  & \multicolumn{1}{c|}{$3.17(\color{black}55.47{\%\uparrow})$}   & \multicolumn{1}{c|}{$3.56(\color{black}72.90{\%\uparrow})$}  & $6.03(\color{black}22.14{\%\uparrow})$ &  & \multicolumn{1}{c|}{$0.623(\color{black}741.90{\%\uparrow})$}  & \multicolumn{1}{c|}{$0.585(\color{black}330.15{\%\uparrow})$}  &  $0.698(\color{black}310.59{\%\uparrow})$\\
\textbf{w/o - Temporal} &  & \multicolumn{1}{c|}{$2.33(\color{black}14.27{\%\uparrow})$}   & \multicolumn{1}{c|}{$2.94(\color{black}42.79{\%\uparrow})$}  & $4.65(\color{black}5.81{\%\downarrow})$ &  & \multicolumn{1}{c|}{$0.325(\color{black}339.19{\%\uparrow})$}  & \multicolumn{1}{c|}{$0.334(\color{black}145.59{\%\uparrow})$}  &  $0.405(\color{black}138.24{\%\uparrow})$\\
\textbf{w/o - STHgCN} &  & \multicolumn{1}{c|}{2.185$(\color{black}7.16{\%\uparrow})$}   & \multicolumn{1}{c|}{3.412 $(\color{black}65.71{\%\uparrow})$}  & 5.101$(\color{black}3.32{\%\uparrow})$ &  & \multicolumn{1}{c|}{0.481$(\color{black}550{\%\uparrow})$}  & \multicolumn{1}{c|}{0.475$(\color{black}249.26{\%\uparrow})$}  & 0.519$(\color{black}205.29{\%\uparrow})$ \\
\textbf{w/o - STTN} & \multirow{-6}{*}{\textbf{\rotatebox[origin=c]{90}{METR-LA}}} &\multicolumn{1}{c|}{2.051$(\color{black}0.59{\%\uparrow})$}   & \multicolumn{1}{c|}{3.267$(\color{black}58.67{\%\uparrow})$}  &  4.957$(\color{black}0.40{\%\uparrow})$           & \multirow{-6}{*}{\textbf{\rotatebox[origin=c]{90}{SWaT}}} &\multicolumn{1}{c|}{0.102$(\color{black}37.84{\%\uparrow})$}  & \multicolumn{1}{c|}{0.143$(\color{black}5.15{\%\uparrow})$}  & 0.206$(\color{black}21.18{\%\uparrow})$ \\ \hline
\end{tabular}
}
\vspace{-2mm}
\end{table*}

\vspace{-3mm}
\begin{table*}[htbp]
\centering
\small
\caption{The table presents the results of an ablation study on multi-horizon forecasting using the PeMS-BAY and Traffic benchmark datasets.}
\label{tab:ablation5}
\vspace{-2mm}
\renewcommand{\arraystretch}{1.075}
\resizebox{1.075\textwidth}{!}{
\hspace{-10mm}\begin{tabular}{c|c|ccc|c|ccc}
\hline
 &  & \multicolumn{3}{c|}{\textbf{MAE}} &  & \multicolumn{3}{c}{\textbf{MAE}} \\ \cline{3-5} \cline{7-9} 
\multirow{-2}{*}{\textbf{Method}} &  & \textbf{Horizon @ 3} & \textbf{Horizon @ 6} & \textbf{Horizon @ 12} &  & \textbf{Horizon @ 3} & \textbf{Horizon @ 6} & \textbf{Horizon @ 12} \\ \cline{1-1} \cline{3-5} \cline{7-9} 
\textbf{JHgRF-Net} &  & \multicolumn{1}{c|}{\textbf{0.806}} & \multicolumn{1}{c|}{\textbf{1.217}} & \textbf{1.758} & & \multicolumn{1}{c|}{\textbf{0.006}} & \multicolumn{1}{c|}{\textbf{0.009}} & \textbf{0.011} \\ \cline{1-1} \cline{3-5} \cline{7-9} 

\textbf{w/o - Spatial} &  & \multicolumn{1}{c|}{$1.17(\color{black}45.16{\%\uparrow})$}   & \multicolumn{1}{c|}{$1.465(\color{black}20.38{\%\uparrow})$}  & $2.095(\color{black}19.17{\%\uparrow})$ &  & \multicolumn{1}{c|}{$0.0198(\color{black}230{\%\uparrow})$}  & \multicolumn{1}{c|}{$0.0163(\color{black}81.11{\%\uparrow})$}  &  $0.0173(\color{black}57.27{\%\uparrow})$\\
\textbf{w/o - Temporal} &  & \multicolumn{1}{c|}{$0.976(\color{black}21.09{\%\uparrow})$}   & \multicolumn{1}{c|}{$1.341(\color{black}10.19{\%\uparrow})$}  & $1.933(\color{black}9.95{\%\uparrow})$ &  & \multicolumn{1}{c|}{$0.0073(\color{black}21.67{\%\uparrow})$}  & \multicolumn{1}{c|}{$0.0142(\color{black}57.78{\%\uparrow})$}  &  $0.0162(\color{black}47.27{\%\uparrow})$\\

\textbf{w/o - STHgCN} &  & \multicolumn{1}{c|}{0.863$(\color{black}7.07{\%\uparrow})$}   & \multicolumn{1}{c|}{1.272$(\color{black}4.52{\%\uparrow})$}  & 1.831$(\color{black}4.15{\%\uparrow})$  & & \multicolumn{1}{c|}{0.0144$(\color{black}140{\%\uparrow})$}  & \multicolumn{1}{c|}{0.0139$(\color{black}54.44{\%\uparrow})$}  & 0.0151$(\color{black}37.27{\%\uparrow})$ \\
\textbf{w/o - STTN} &\multirow{-6}{*}{\textbf{\rotatebox[origin=c]{90}{PeMS-BAY}}} & \multicolumn{1}{c|}{0.814$(\color{black}0.99{\%\uparrow})$}   & \multicolumn{1}{c|}{1.251$(\color{black}2.79{\%\uparrow})$}  & 1.769$(\color{black}0.63{\%\uparrow})$   &\multirow{-6}{*}{\textbf{\rotatebox[origin=c]{90}{Traffic}}} & \multicolumn{1}{c|}{0.0067$(\color{black}11.67{\%\uparrow})$}  & \multicolumn{1}{c|}{0.0101$(\color{black}12.22{\%\uparrow})$}  & 0.0116$(\color{black}5.45{\%\uparrow})$ \\ \hline
\end{tabular}
}
\vspace{-2mm}
\end{table*}

\vspace{-3mm}
\begin{table*}[htbp]
\centering
\small
\caption{The table presents the results of an ablation study on multi-horizon forecasting using the WADI and Electricity benchmark datasets.}
\label{tab:ablation6}
\vspace{-2mm}
\renewcommand{\arraystretch}{1.075}
\resizebox{1.075\textwidth}{!}{
\hspace{-10mm}\begin{tabular}{c|c|ccc|c|ccc}
\hline
 &  & \multicolumn{3}{c|}{\textbf{MAE}} &  & \multicolumn{3}{c}{\textbf{MAE}} \\ \cline{3-5} \cline{7-9} 
\multirow{-2}{*}{\textbf{Method}} &  & \textbf{Horizon @ 3} & \textbf{Horizon @ 6} & \textbf{Horizon @ 12} &  & \textbf{Horizon @ 3} & \textbf{Horizon @ 6} & \textbf{Horizon @ 12} \\ \cline{1-1} \cline{3-5} \cline{7-9} 
\textbf{JHgRF-Net} &  & \multicolumn{1}{c|}{\textbf{4.149}}           & \multicolumn{1}{c|}{\textbf{4.592}}          & \textbf{4.758}          &  & \multicolumn{1}{c|}{\textbf{160.955}}         & \multicolumn{1}{c|}{238.781}         & 225.724         \\ \cline{1-1} \cline{3-5} \cline{7-9} 

\textbf{w/o - Spatial} &  & \multicolumn{1}{c|}{$6.031(\color{black}45.36{\%\uparrow})$}   & \multicolumn{1}{c|}{$5.982(\color{black}30.27{\%\uparrow})$}  & $5.733(\color{black}20.49{\%\uparrow})$ &  & \multicolumn{1}{c|}{$209.325(\color{black}30.05{\%\uparrow})$}  & \multicolumn{1}{c|}{$312.475(\color{black}30.86{\%\uparrow})$}  &  $298.685(\color{black}32.32{\%\uparrow})$\\
\textbf{w/o - Temporal} &  & \multicolumn{1}{c|}{$4.865(\color{black}17.26{\%\uparrow})$}   & \multicolumn{1}{c|}{$4.803(\color{black}4.59{\%\uparrow})$}  & $4.843(\color{black}1.79{\%\uparrow})$ &  & \multicolumn{1}{c|}{$176.089(\color{black}9.40{\%\uparrow})$}  & \multicolumn{1}{c|}{$263.168(\color{black}10.21{\%\uparrow})$}  &  $247.905(\color{black}9.83{\%\uparrow})$\\

\textbf{w/o - STHgCN} &  & \multicolumn{1}{c|}{5.171$(\color{black}24.63{\%\uparrow})$}   & \multicolumn{1}{c|}{5.003$(\color{black}8.95{\%\uparrow})$}  & 4.917$(\color{black}3.34{\%\uparrow})$  & & \multicolumn{1}{c|}{183.040$(\color{black}13.72{\%\uparrow})$}  & \multicolumn{1}{c|}{269.294$(\color{black}12.78{\%\uparrow})$}  & 265.346$(\color{black}17.55{\%\uparrow})$ \\
\textbf{w/o - STTN} &\multirow{-5}{*}{\textbf{\rotatebox[origin=c]{90}{WADI}}} & \multicolumn{1}{c|}{4.197$(\color{black}1.16{\%\uparrow})$}   & \multicolumn{1}{c|}{4.744$(\color{black}3.31{\%\uparrow})$}  & 4.835$(\color{black}1.61{\%\uparrow})$  &\multirow{-5}{*}{\textbf{\rotatebox[origin=c]{90}{Electricity}}} & \multicolumn{1}{c|}{168.228$(\color{black}4.52{\%\uparrow})$}  & \multicolumn{1}{c|}{257.872$(\color{black}7.99{\%\uparrow})$}  & 238.459$(\color{black}5.64{\%\uparrow})$\\ \hline
\end{tabular}
}
\vspace{-2mm}
\end{table*}

 is excluded as compared to the baseline, emphasizing its indispensable nature. Our proposed neural forecast architecture is built upon two fundamental methods known as the STHgCN and STTN neural operators, which collectively make up the spatial inference component. The following observations were made from the ablation studies: 

\vspace{-1mm}
\begin{itemize}
\item The ``w/o - STHgCN" variant yielded inferior results compared to the benchmark, with a difference of $12.03\%$, $10.66\%$, and $11.43\%$ in terms of RMSE, MAE, and MAPE metrics, respectively, for PeMSD4 dataset. Similarly on PeMSD8, the variants exhibited a $10.52\%$, $9.97\%$, and $9.89\%$ decrease in performance with respect to the same metrics as compared to the benchmark. These results provide evidence in support of the notion that incorporating STHgCN method in the learning process can result in better performance in multi-horizon forecasting tasks.
\item The ``w/o - STTN" variant exhibited a slight increase in the RMSE, MAE, and MAPE metrics compared to the baseline, with differences of $3.79\%$, $0.31\%$, and $0.61\%$ on PeMSD4, and $1.90\%$, $1.88\%$, and $2.29\%$ on PeMSD8, respectively. Nonetheless, the integration of the STTN method was found to be crucial, as it resulted in a notable improvement in forecast accuracy.
\end{itemize}

Based on the ablation studies, we can conclude that STHgCN method is more effective than the STTN method in accurately modeling spatio-temporal dependencies in MTS data, leading to better multi-horizon forecasts. Additional results from the ablation study on benchmark datasets are presented in Tables \ref{tab:ablation1} - \ref{tab:ablation7}. The results indicate that the proposed \textbf{JHgRF-Net} framework exhibits strong generalization capabilities, even when dealing with intricate patterns across an extensive variety of datasets, and it can efficiently scale to handle large-scale graph datasets. In summary, the ablation studies provide evidence in favor of the hypothesis that joint optimization of spatial-temporal inference components can lead to enhanced performance in multi-horizon forecasting tasks. In addition, the experimental findings support the rationale of inclusion of STHgCN and STTN neural operators to model the interdependencies among the multiple variables and learn the dynamics of the complex interconnected systems.

\vspace{-1mm}
\begin{table*}[!ht]
\centering
\small
\caption{The table presents the results of an ablation study on multi-horizon forecasting using the Solar Energy and Exchange Rate benchmark datasets.}
\label{tab:ablation7}
\vspace{-3mm}
\renewcommand{\arraystretch}{1.075}
\resizebox{1.075\textwidth}{!}{
\hspace{-10mm}\begin{tabular}{c|c|ccc|c|ccc}
\hline
 &  & \multicolumn{3}{c|}{\textbf{MAE}} &  & \multicolumn{3}{c}{\textbf{MAE}} \\ \cline{3-5} \cline{7-9} 
\multirow{-2}{*}{\textbf{Method}} &  & \textbf{Horizon @ 3} & \textbf{Horizon @ 6} & \textbf{Horizon @ 12} &  & \textbf{Horizon @ 3} & \textbf{Horizon @ 6} & \textbf{Horizon @ 12} \\ \cline{1-1} \cline{3-5} \cline{7-9} 
\textbf{JHgRF-Net} &  & \multicolumn{1}{c|}{\textbf{0.575} $\pm$ 0.013}           & \multicolumn{1}{c|}{\textbf{0.868}}          & \textbf{0.873}      &  & \multicolumn{1}{c|}{\textbf{0.0043}}         & \multicolumn{1}{c|}{\textbf{0.0044}}         & \textbf{0.0064}          \\ \cline{1-1} \cline{3-5} \cline{7-9} 

\textbf{w/o - Spatial} &  & \multicolumn{1}{c|}{$1.132(\color{black}96.87{\%\uparrow})$}   & \multicolumn{1}{c|}{$1.059(\color{black}22{\%\uparrow})$}  & $3.087(\color{black}253.61{\%\uparrow})$ &  & \multicolumn{1}{c|}{$0.0284(\color{black}560.47{\%\uparrow})$}  & \multicolumn{1}{c|}{$0.0182(\color{black}313.64{\%\uparrow})$}  &  $0.0178(\color{black}178.12{\%\uparrow})$\\
\textbf{w/o - Temporal} &  & \multicolumn{1}{c|}{$0.726(\color{black}26.26{\%\uparrow})$}   & \multicolumn{1}{c|}{$0.883(\color{black}1.73{\%\uparrow})$}  & $1.898(\color{black}117.41{\%\uparrow})$ &  & \multicolumn{1}{c|}{$0.0097(\color{black}125.58{\%\uparrow})$}  & \multicolumn{1}{c|}{$0.0081(\color{black}84.09{\%\uparrow})$}  &  $0.0091(\color{black}42.18{\%\uparrow})$\\

\textbf{w/o - STHgCN} &  & \multicolumn{1}{c|}{0.826$(\color{black}43.65{\%\uparrow})$}   & \multicolumn{1}{c|}{0.896$(\color{black}3.23{\%\uparrow})$}  & 2.132$(\color{black}144.22{\%\uparrow})$ & & \multicolumn{1}{c|}{0.0122$(\color{black}183.72{\%\uparrow})$}  & \multicolumn{1}{c|}{0.0085$(\color{black}93.18{\%\uparrow})$}  & 0.0093$(\color{black}45.31{\%\uparrow})$\\
\textbf{w/o - STTN} &\multirow{-6}{*}{\textbf{\rotatebox[origin=c]{90}{Solar Energy}}} & \multicolumn{1}{c|}{0.596$(\color{black}3.65{\%\uparrow})$}   & \multicolumn{1}{c|}{0.874 $(\color{black}0.69{\%\uparrow})$}  &  1.412$(\color{black}61.74{\%\uparrow})$  &\multirow{-6}{*}{\textbf{\rotatebox[origin=c]{90}{Exchange Rate}}} & \multicolumn{1}{c|}{0.0066$(\color{black}53.49{\%\uparrow})$}  & \multicolumn{1}{c|}{0.0076$(\color{black}72.73{\%\uparrow})$}  & 0.0086$(\color{black}34.37{\%\uparrow})$\\ \hline
\end{tabular}
}
\vspace{-2mm}
\end{table*}

\vspace{-1mm}
\subsection{Prediction error for multi-horizon forecasting}
\vspace{-1mm}
We conducted comprehensive experiments to evaluate the capability of the neural forecasting architecture, \textbf{JHgRF-Net}, to generate accurate multi-horizon forecasts on several benchmark datasets. The forecast errors of the \textbf{JHgRF-Net} framework performance on benchmark datasets are shown in Figure \ref{fig:horizon_appendix}. The framework performance was evaluated using various metrics, such as MAPE and MAE. Lower values of forecast errors indicate better model performance.  The results demonstrate that the framework outperformed the baselines on all the prediction horizons. These findings suggest that the proposed framework has the potential to accurately model the nonlinear spatio-temporal dependencies and improve multi-horizon forecast accuracy through effectively exploiting the relational inductive biases within the hypergraph-structured MTS data.

\vspace{1mm}
\subsection{Irregular time series forecasting}
\vspace{-1mm}
The \textbf{JHgRF-Net} framework ability to handle missing data in large, complex sensor networks was evaluated by simulating two commonly observed missingness patterns(\cite{marisca2022learning}, \cite{cini2022sparse}): point-missing and block-missing patterns. These patterns were created to mimic the missingness patterns observed in real-world data of such complex interconnected sensor networks. In point-missing pattern, observations of each variable were randomly dropped within a historical window, with missing ratios of 10$\%$, 30$\%$, and 50$\%$. Similarly, in block-missing pattern, available data for each variable was randomly masked within a historical window, also with missing ratios ranging from 10$\%$, 30$\%$, and 50$\%$. Moreover, sensor failures were simulated with a probability of 0.15$\%$, leading to blocks of missing data for the multivariate time series data. To evaluate the \textbf{JHgRF-Net} framework performance on MTS data with missing values and to analyze the impact of increasing missing data percentage on framework performance, we split several benchmark datasets into three mutually exclusive sets - training, validation, and test - based on their chronological order. The METR-LA and PEMS-BAY datasets were split in a ratio of 7:1:2, while the other datasets(PeMSD3, PeMSD4, PeMSD7, PeMSD8, and PeMSD7(M), Traffic, Solar-Energy, Electricity, Exchange-Rate) were split in a ratio of 6:2:2. We utilized multiple forecasting metrics to evaluate the \textbf{JHgRF-Net} framework performance in handling missing data. We trained the \textbf{JHgRF-Net} framework on fully observed data to establish a benchmark for the MTSF task with missing values. The tables \ref{tab:misssingtable_1} - \ref{tab:missingtable_5} present the forecasting results of the framework performance on the irregular-time-series datasets. The experimental studies demonstrate that the \textbf{JHgRF-Net} framework is reliable and robust in handling missing data, which is widely prevalent in real-world applications. The framework performance deteriorates slightly compared to the benchmark when there is a lower percentage of missing data. With a further increase in the percentage of missing data, the framework performance continues to decline, resulting in lower forecast accuracy across all benchmark datasets, regardless of the missing data pattern. Instead of relying on imputed values for model predictions, the proposed framework utilizes observed data for multi-horizon forecasting, hence demonstrates its robustness to handle missing data. Moreover, by capturing complex dependencies and patterns within multivariate time series data present in interconnected networks, the framework generates more dependable out-of-sample forecasts, resulting in enhanced multi-horizon forecast accuracy.

\vspace{-4mm}
\newcommand{\addpic}{\includegraphics[width=3em]{example-image}}
\newcolumntype{C}{>{\centering\arraybackslash}m{11.0em}}
\begin{table}[!ht]
\setlength{\tabcolsep}{1pt}
\caption{The figure shows the pointwise prediction error for multi-horizon forecasting tasks on benchmark datasets.}
\label{fig:horizon_appendix}
\vspace{1mm}
\begin{tabular}{l|*2{C}@{}}
\toprule
 & \textbf{MAE}  & \textbf{MAPE}  \\ 
\midrule
\rotatebox[origin=c]{90}{\textbf{PeMSD3}} & \includegraphics[width=0.425\columnwidth]{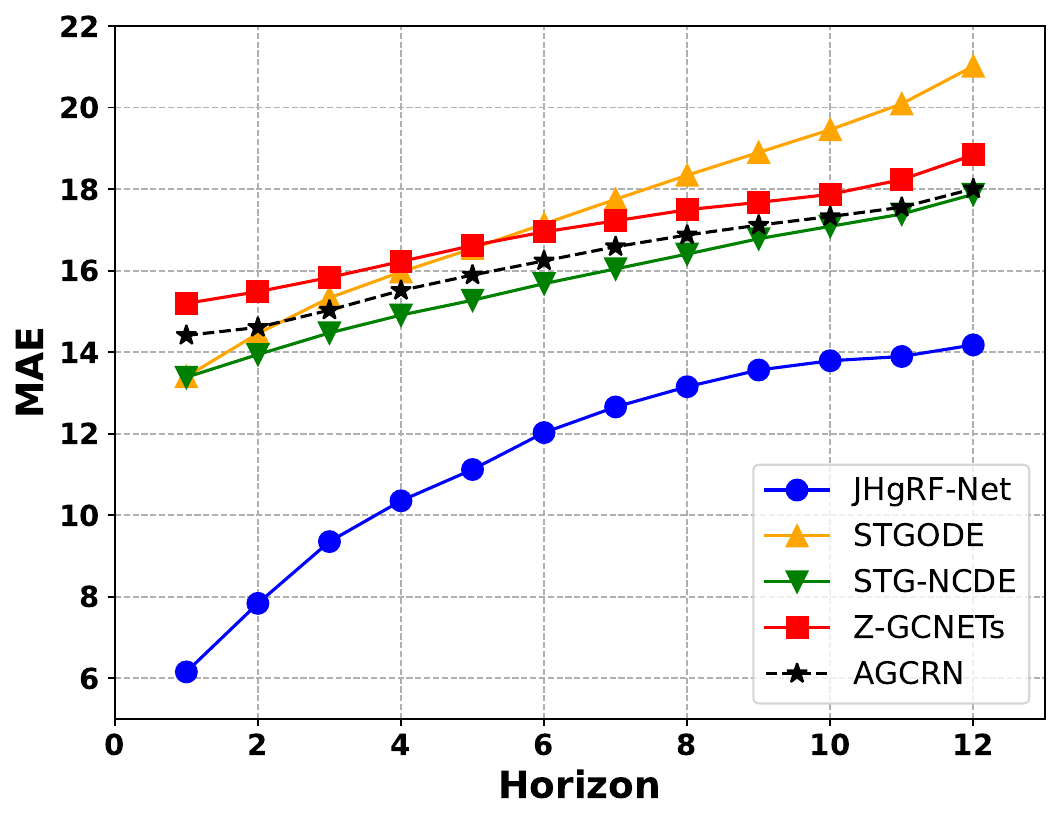} & \includegraphics[width=0.425\columnwidth]{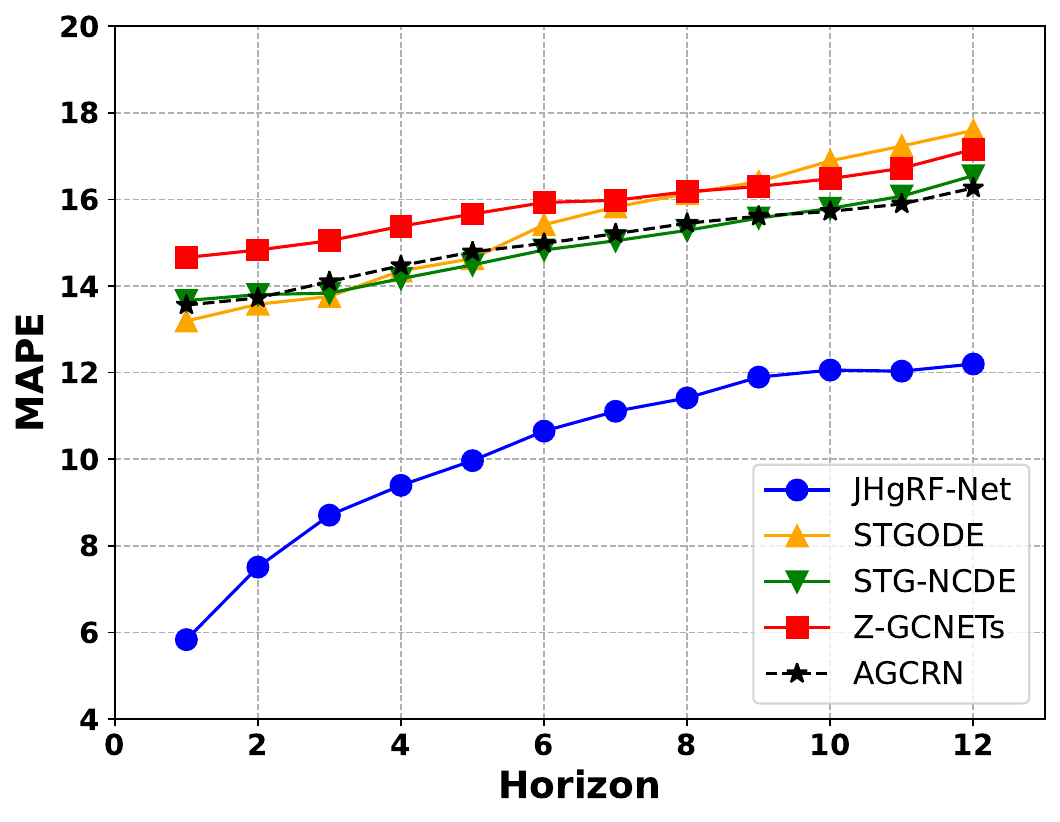}  \\ \hline 
\rotatebox[origin=c]{90}{\textbf{PeMSD4}} & \includegraphics[width=0.425\columnwidth]{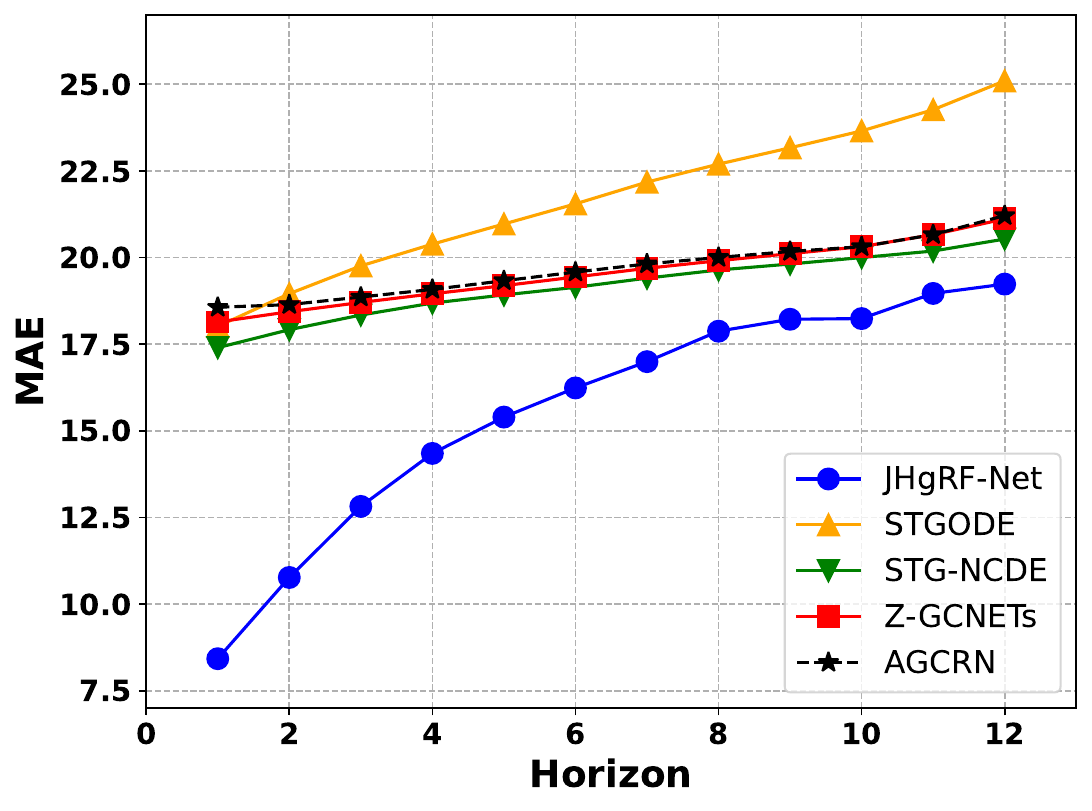}  & \includegraphics[width=0.425\columnwidth]{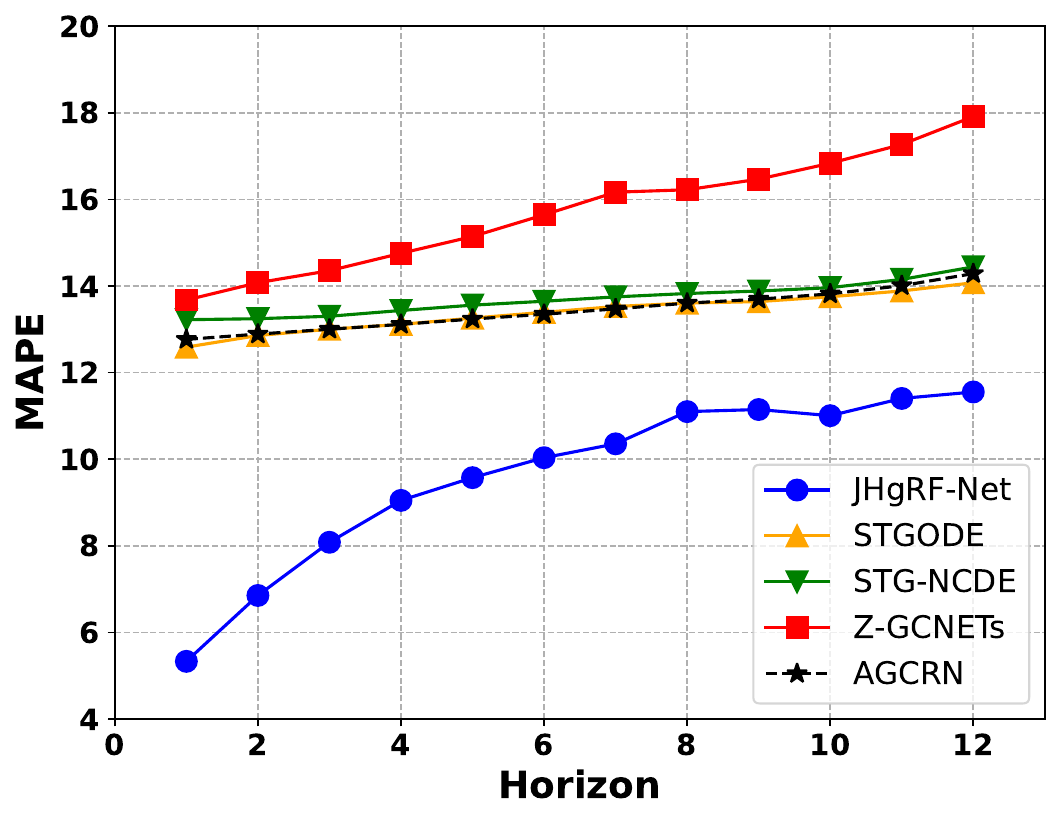}  \\ \hline
\rotatebox[origin=c]{90}{\textbf{PeMSD7}} & \includegraphics[width=0.425\columnwidth]{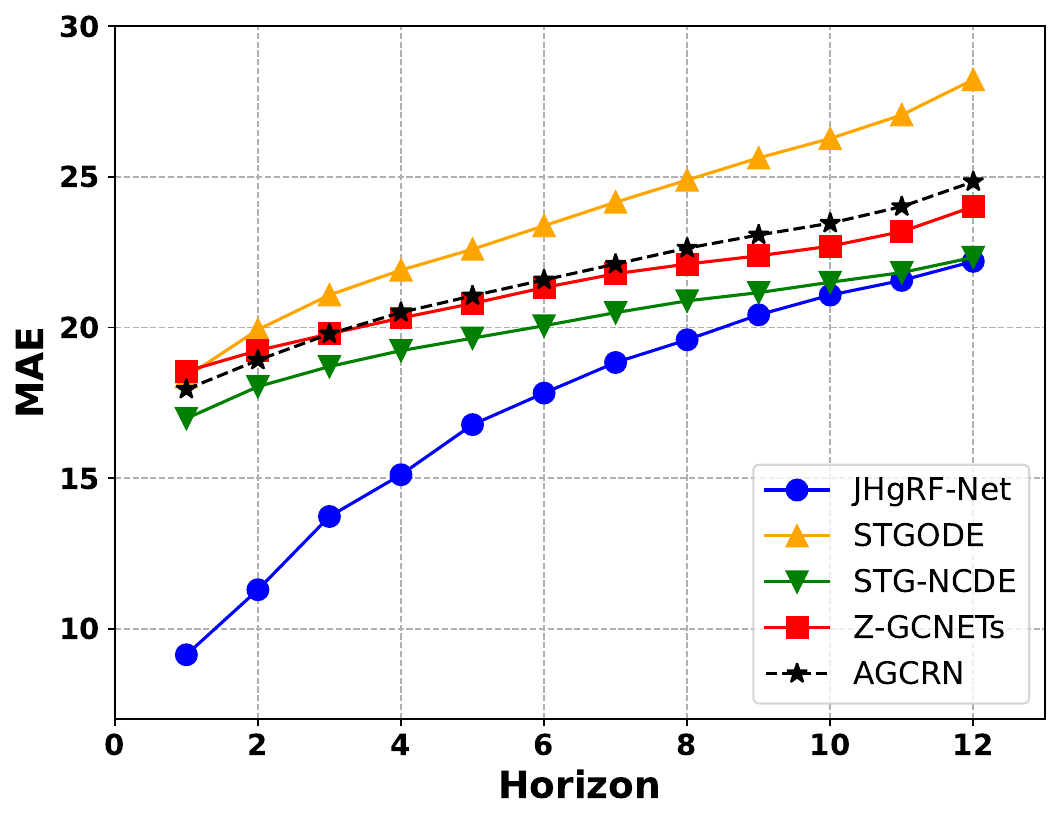}  & \includegraphics[width=0.425\columnwidth]{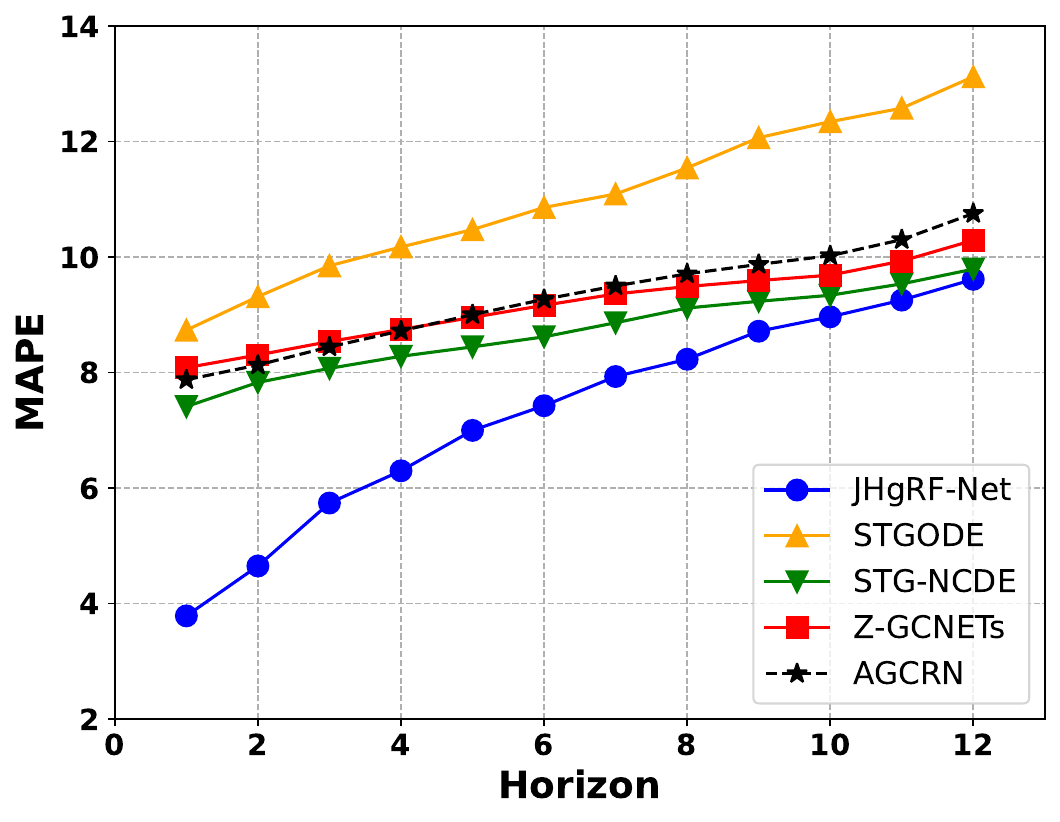}  \\ \hline
\rotatebox[origin=c]{90}{\textbf{PeMSD8}} & \includegraphics[width=0.425\columnwidth]{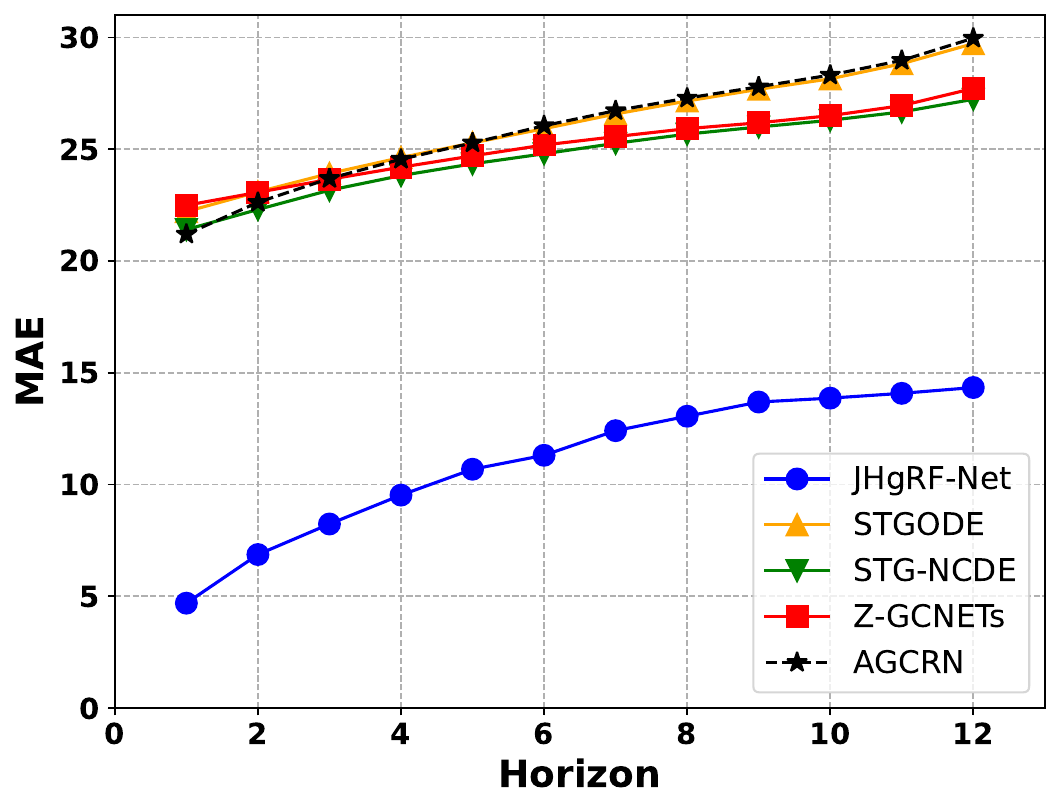}  & \includegraphics[width=0.425\columnwidth]{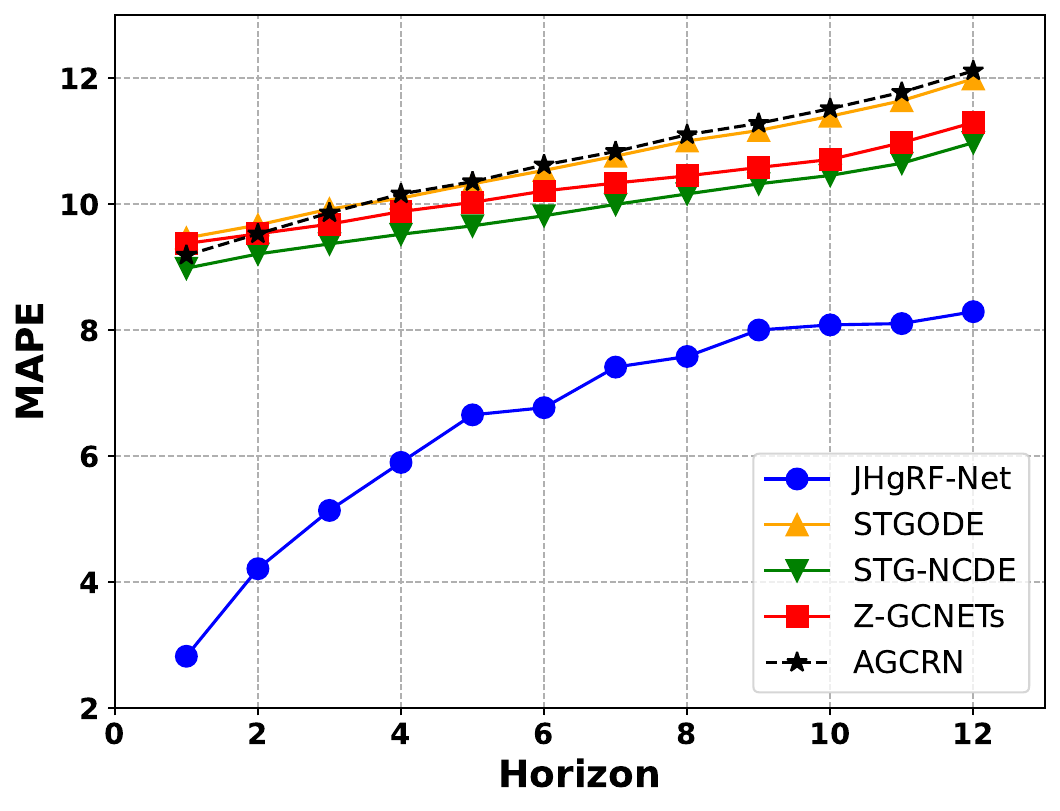}  \\ \hline
\rotatebox[origin=c]{90}{\textbf{PeMSD7M}} & \includegraphics[width=0.425\columnwidth]{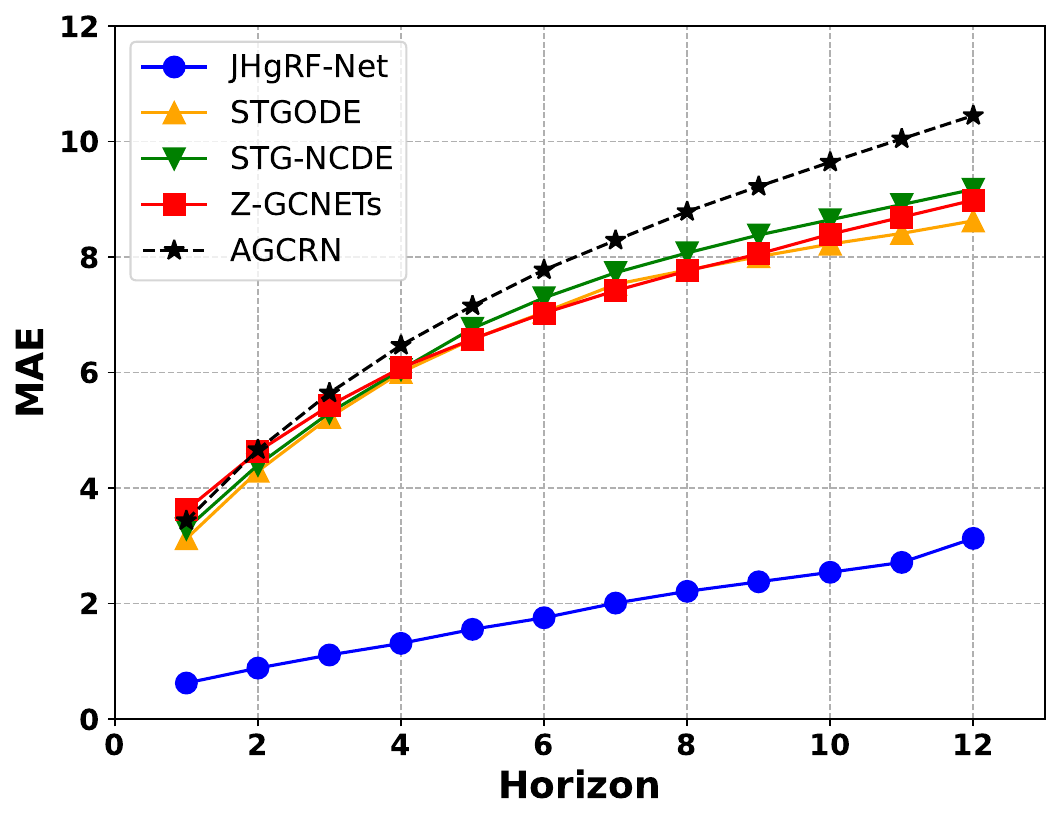} & \includegraphics[width=0.425\columnwidth]{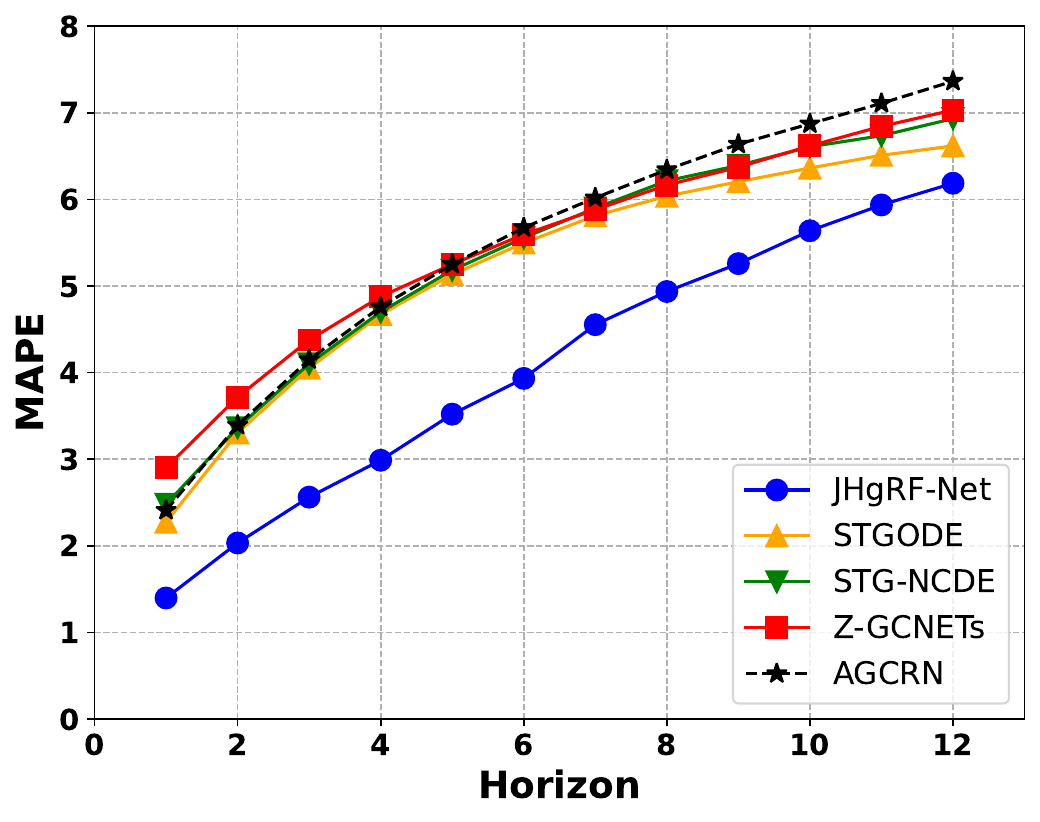}  \\ 
\bottomrule 
\end{tabular}
\end{table} 

\begin{table*}[htbp]
\centering
\renewcommand{\arraystretch}{1.3}
\caption{Pointwise forecasting error on irregular PeMSD3, PeMSD4 and PeMSD7 datasets at horizon@12}
\label{tab:misssingtable_1}
\vspace{1mm}
\setlength{\tabcolsep}{5pt}
\resizebox{0.85\textwidth}{!}{%
\begin{tabular}{c|c|c|c|c|c|c|c|c|c|c|c|c|c}
\hline
\textbf{\begin{tabular}[c]{@{}c@{}}Missing\\      Rate\end{tabular}} & \textbf{Model}     & \multirow{8}{*}{\rotatebox[origin=c]{90}{\textbf{PeMSD3}}} & \textbf{MAE} & \textbf{RMSE} & \textbf{MAPE} & \multirow{8}{*}{\rotatebox[origin=c]{90}{\textbf{PeMSD4}}} & \textbf{MAE} & \textbf{RMSE} & \textbf{MAPE} & \multirow{8}{*}{\rotatebox[origin=c]{90}{\textbf{PeMSD7}}} & \textbf{MAE} & \textbf{RMSE} & \textbf{MAPE} \\ \cline{1-2} \cline{4-6} \cline{8-10} \cline{12-14} 
\textbf{0\%}                                                         & \textbf{JHgRF-Net} &                                  & 14.18        & 21.48         & 12.19         &                                  & 19.23        & 28.76         & 11.55         &                                  & 22.19        & 32.89         & 9.61          \\ \cline{1-2} \cline{4-6} \cline{8-10} \cline{12-14} 
\multirow{2}{*}{\textbf{10\%}}                                       & \textbf{w/Point}   &                                  & 15.09        & 22.79         & 12.88         &                                  & 20.91        & 30.63         & 12.87         &                                  & 23.77        & 34.83         & 10.23         \\ \cline{2-2} \cline{4-6} \cline{8-10} \cline{12-14} 
                                                                     & \textbf{w/Block}   &                                  & 15.04        & 22.74         & 12.77         &                                  & 20.87        & 30.64         & 30.64         &                                  & 23.61        & 34.70          & 10.17         \\ \cline{1-2} \cline{4-6} \cline{8-10} \cline{12-14} 
\multirow{2}{*}{\textbf{30\%}}                                       & \textbf{w/Point}   &                                  & 15.58        & 23.45         & 13.16         &                                  & 21.79        & 31.73         & 13.27         &                                  & 24.99        & 36.32         & 10.81         \\ \cline{2-2} \cline{4-6} \cline{8-10} \cline{12-14} 
                                                                     & \textbf{w/Block}   &                                  & 15.60         & 23.49         & 13.13         &                                  & 21.63        & 31.61         & 13.19         &                                  & 24.98        & 36.31         & 10.84         \\ \cline{1-2} \cline{4-6} \cline{8-10} \cline{12-14} 
\multirow{2}{*}{\textbf{50\%}}                                       & \textbf{w/Point}   &                                  & 16.34        & 24.42         & 13.84         &                                  & 23.12        & 33.49         & 14.11         &                                  & 26.65        & 38.47         & 11.69          \\ \cline{2-2} \cline{4-6} \cline{8-10} \cline{12-14} 
                                                                     & \textbf{w/Block}   &                                  & 16.20         & 24.29         & 13.66         &                                  & 23.24        & 33.84         & 14.11         &                                  & 26.63        & 38.33         & 11.54         \\ \hline
\end{tabular}
}
\end{table*}

\begin{table*}[!h]
\centering
\renewcommand{\arraystretch}{1.2}
\caption{Pointwise forecasting error on irregular PeMSD7(M) and PeMSD8 datasets at horizon@12}
\label{tab:misssingtable_2}
\vspace{1mm}
\setlength{\tabcolsep}{5pt}
\resizebox{0.65\textwidth}{!}{%
\begin{tabular}{c|c|c|c|c|c|c|c|c|c}
\hline
\textbf{\begin{tabular}[c]{@{}c@{}}Missing\\      Rate\end{tabular}} & \textbf{Model}     & \multirow{8}{*}{\rotatebox[origin=c]{90}{\textbf{PeMSD7(M)}}} & \textbf{MAE} & \textbf{RMSE} & \textbf{MAPE} & \multirow{8}{*}{\rotatebox[origin=c]{90}{\textbf{PeMSD8}}} & \textbf{MAE} & \textbf{RMSE} & \textbf{MAPE} \\ \cline{1-2} \cline{4-6} \cline{8-10} 
\textbf{0\%}                                                         & \textbf{JHgRF-Net} &                                     & 2.83         & 5.01          & 6.18          &                                  & 14.34        & 22.06         & 8.29          \\ \cline{1-2} \cline{4-6} \cline{8-10} 
\multirow{2}{*}{\textbf{10\%}}                                       & \textbf{w/Point}   &                                     & 3.34         & 5.53          & 7.14          &                                  & 16.95        & 25.23         & 9.94          \\ \cline{2-2} \cline{4-6} \cline{8-10} 
                                                                     & \textbf{w/Block}   &                                     & 3.27         & 5.47          & 7.02          &                                  & 16.85        & 25.08         & 9.84          \\ \cline{1-2} \cline{4-6} \cline{8-10} 
\multirow{2}{*}{\textbf{30\%}}                                       & \textbf{w/Point}   &                                     & 3.56         & 5.78          & 7.54          &                                  & 17.48        & 25.89         & 10.14         \\ \cline{2-2} \cline{4-6} \cline{8-10} 
                                                                     & \textbf{w/Block}   &                                     & 3.48         & 5.72          & 7.42          &                                  & 17.51        & 25.95         & 10.17         \\ \cline{1-2} \cline{4-6} \cline{8-10} 
\multirow{2}{*}{\textbf{50\%}}                                       & \textbf{w/Point}   &                                     & 3.66         & 5.93          & 7.71          &                                  & 18.72        & 27.53         & 10.88         \\ \cline{2-2} \cline{4-6} \cline{8-10} 
                                                                     & \textbf{w/Block}   &                                     & 3.76         & 6.02          & 7.89          &                                  & 18.58        & 27.36         & 10.88         \\ \hline
\end{tabular}
}
\end{table*}

\begin{table*}[!h]
\renewcommand{\arraystretch}{1.325}
\centering
\caption{Pointwise forecasting error on irregular METR-LA, PeMS-BAY and SWaT datasets in terms of MAE}
\label{tab:missingtable_3}
\vspace{1mm}
\setlength{\tabcolsep}{5pt}
\resizebox{1.10\textwidth}{!}{%
\hspace{-7mm}\begin{tabular}{c|c|c|ccc|c|ccc|c|ccc}
\hline
\multirow{2}{*}{\textbf{\begin{tabular}[c]{@{}c@{}}Missing\\      Rate\end{tabular}}} & \multirow{2}{*}{\textbf{Model}} & \multirow{9}{*}{\rotatebox[origin=c]{90}{\textbf{METR-LA}}} & \multicolumn{3}{c|}{\textbf{MAE}}                                            & \multirow{9}{*}{\rotatebox[origin=c]{90}{\textbf{PeMS-BAY}}} & \multicolumn{3}{c|}{\textbf{MAE}}                                            & \multirow{9}{*}{\rotatebox[origin=c]{90}{\textbf{SWaT}}} & \multicolumn{3}{c}{\textbf{MAE}}                                             \\ \cline{4-6} \cline{8-10} \cline{12-14} 
                                                                                      &                                 &                                   & \multicolumn{1}{c|}{Horizon@3} & \multicolumn{1}{c|}{Horizon@6} & Horizon@12 &                                    & \multicolumn{1}{c|}{Horizon@3} & \multicolumn{1}{c|}{Horizon@6} & Horizon@12 &                                & \multicolumn{1}{c|}{Horizon@3} & \multicolumn{1}{c|}{Horizon@6} & Horizon@12 \\ \cline{1-2} \cline{4-6} \cline{8-10} \cline{12-14} 
\textbf{0\%}                                                                          & \textbf{JHgRF-Net}              &                                   & \multicolumn{1}{c|}{2.039}     & \multicolumn{1}{c|}{2.059}     & 4.937      &                                    & \multicolumn{1}{c|}{0.806}     & \multicolumn{1}{c|}{1.217}     & 1.758      &                                & \multicolumn{1}{c|}{0.074}     & \multicolumn{1}{c|}{0.136}     & 0.17       \\ \cline{1-2} \cline{4-6} \cline{8-10} \cline{12-14} 
\multirow{2}{*}{\textbf{10\%}}                                                        & \textbf{w/Point}                &                                   & \multicolumn{1}{c|}{2.083}     & \multicolumn{1}{c|}{2.093}     & 4.951      &                                    & \multicolumn{1}{c|}{0.813}     & \multicolumn{1}{c|}{1.225}     & 1.763      &                                & \multicolumn{1}{c|}{0.076}     & \multicolumn{1}{c|}{0.138}     & 0.174      \\ \cline{2-2} \cline{4-6} \cline{8-10} \cline{12-14} 
                                                                                      & \textbf{w/Block}                &                                   & \multicolumn{1}{c|}{2.127}     & \multicolumn{1}{c|}{2.138}     & 4.983      &                                    & \multicolumn{1}{c|}{0.824}     & \multicolumn{1}{c|}{1.239}     & 1.767      &                                & \multicolumn{1}{c|}{0.078}     & \multicolumn{1}{c|}{0.141}     & 0.178      \\ \cline{1-2} \cline{4-6} \cline{8-10} \cline{12-14} 
\multirow{2}{*}{\textbf{30\%}}                                                        & \textbf{w/Point}                &                                   & \multicolumn{1}{c|}{2.137}     & \multicolumn{1}{c|}{2.143}     & 5.113      &                                    & \multicolumn{1}{c|}{0.842}     & \multicolumn{1}{c|}{1.241}     & 1.772      &                                & \multicolumn{1}{c|}{0.086}     & \multicolumn{1}{c|}{0.153}     & 0.188      \\ \cline{2-2} \cline{4-6} \cline{8-10} \cline{12-14} 
                                                                                      & \textbf{w/Block}                &                                   & \multicolumn{1}{c|}{2.175}     & \multicolumn{1}{c|}{2.189}     & 5.267      &                                    & \multicolumn{1}{c|}{0.856}     & \multicolumn{1}{c|}{1.253}     & 1.793      &                                & \multicolumn{1}{c|}{0.091}     & \multicolumn{1}{c|}{0.164}     & 0.198      \\ \cline{1-2} \cline{4-6} \cline{8-10} \cline{12-14} 
\multirow{2}{*}{\textbf{50\%}}                                                        & \textbf{w/Point}                &                                   & \multicolumn{1}{c|}{2.166}     & \multicolumn{1}{c|}{2.178}     & 5.671      &                                    & \multicolumn{1}{c|}{0.861}     & \multicolumn{1}{c|}{1.379}     & 1.845      &                                & \multicolumn{1}{c|}{0.099}     & \multicolumn{1}{c|}{0.189}     & 0.201      \\ \cline{2-2} \cline{4-6} \cline{8-10} \cline{12-14} 
                                                                                      & \textbf{w/Block}                &                                   & \multicolumn{1}{c|}{2.181}     & \multicolumn{1}{c|}{2.193}     & 5.703      &                                    & \multicolumn{1}{c|}{0.878}     & \multicolumn{1}{c|}{1.384}     & 1.863      &                                & \multicolumn{1}{c|}{0.112}     & \multicolumn{1}{c|}{0.197}     & 0.211      \\ \hline
\end{tabular}
}
\end{table*}

\begin{table*}[!ht]
\renewcommand{\arraystretch}{1.325}
\centering
\caption{Pointwise forecasting error on irregular Electricity, Exchange Rate and Solar Energy datasets in terms of MAE}
\label{tab:missingtable_4}
\vspace{-1mm}
\setlength{\tabcolsep}{5pt}
\resizebox{1.10\textwidth}{!}{%
\hspace{-7mm}\begin{tabular}{c|c|c|ccc|c|ccc|c|ccc}
\hline
\multirow{2}{*}{\begin{tabular}[c]{@{}c@{}}Missing\\      Rate\end{tabular}} & \multirow{2}{*}{Model} & \multirow{9}{*}{\rotatebox[origin=c]{90}{\textbf{Solar Energy}}} & \multicolumn{3}{c|}{MAE}                                                     & \multirow{9}{*}{\rotatebox[origin=c]{90}{\textbf{Electricity}}} & \multicolumn{3}{c|}{MAE}                                                     & \multirow{9}{*}{\rotatebox[origin=c]{90}{\textbf{Exchange Rate}}} & \multicolumn{3}{c}{MAE}                                                      \\ \cline{4-6} \cline{8-10} \cline{12-14} 
                                                                             &                        &                               & \multicolumn{1}{c|}{Horizon@3} & \multicolumn{1}{c|}{Horizon@6} & Horizon@12 &                              & \multicolumn{1}{c|}{Horizon@3} & \multicolumn{1}{c|}{Horizon@6} & Horizon@12 &                                & \multicolumn{1}{c|}{Horizon@3} & \multicolumn{1}{c|}{Horizon@6} & Horizon@12 \\ \cline{1-2} \cline{4-6} \cline{8-10} \cline{12-14} 
0\%                                                                          & JHgRF-Net              &                               & \multicolumn{1}{c|}{0.575}     & \multicolumn{1}{c|}{0.868}     & 0.873      &                              & \multicolumn{1}{c|}{160.955}   & \multicolumn{1}{c|}{178.781}   & 225.724    &                                & \multicolumn{1}{c|}{0.0043}    & \multicolumn{1}{c|}{0.0044}    & 0.0064     \\ \cline{1-2} \cline{4-6} \cline{8-10} \cline{12-14} 
\multirow{2}{*}{10\%}                                                        & w/Point                &                               & \multicolumn{1}{c|}{0.579}     & \multicolumn{1}{c|}{0.871}     & 0.877      &                              & \multicolumn{1}{c|}{163.072}   & \multicolumn{1}{c|}{181.213}   & 228.358    &                                & \multicolumn{1}{c|}{0.00435}   & \multicolumn{1}{c|}{0.00456}   & 0.00653    \\ \cline{2-2} \cline{4-6} \cline{8-10} \cline{12-14} 
                                                                             & w/Block                &                               & \multicolumn{1}{c|}{0.588}     & \multicolumn{1}{c|}{0.879}     & 0.893      &                              & \multicolumn{1}{c|}{165.321}   & \multicolumn{1}{c|}{184.369}   & 230.967    &                                & \multicolumn{1}{c|}{0.00437}   & \multicolumn{1}{c|}{0.00459}   & 0.0066     \\ \cline{1-2} \cline{4-6} \cline{8-10} \cline{12-14} 
\multirow{2}{*}{30\%}                                                        & w/Point                &                               & \multicolumn{1}{c|}{0.613}     & \multicolumn{1}{c|}{0.911}     & 0.937      &                              & \multicolumn{1}{c|}{170.325}   & \multicolumn{1}{c|}{192.565}   & 244.637    &                                & \multicolumn{1}{c|}{0.00553}   & \multicolumn{1}{c|}{0.00578}   & 0.00641    \\ \cline{2-2} \cline{4-6} \cline{8-10} \cline{12-14} 
                                                                             & w/Block                &                               & \multicolumn{1}{c|}{0.637}     & \multicolumn{1}{c|}{0.943}     & 0.981      &                              & \multicolumn{1}{c|}{173.368}   & \multicolumn{1}{c|}{200.098}   & 257.974    &                                & \multicolumn{1}{c|}{0.00693}   & \multicolumn{1}{c|}{0.00702}   & 0.00725    \\ \cline{1-2} \cline{4-6} \cline{8-10} \cline{12-14} 
\multirow{2}{*}{50\%}                                                        & w/Point                &                               & \multicolumn{1}{c|}{0.689}     & \multicolumn{1}{c|}{0.988}     & 1.103      &                              & \multicolumn{1}{c|}{189.765}   & \multicolumn{1}{c|}{215.131}   & 274.387    &                                & \multicolumn{1}{c|}{0.00778}   & \multicolumn{1}{c|}{0.00869}   & 0.00913    \\ \cline{2-2} \cline{4-6} \cline{8-10} \cline{12-14} 
                                                                             & w/Block                &                               & \multicolumn{1}{c|}{0.713}     & \multicolumn{1}{c|}{0.997}     & 1.119      &                              & \multicolumn{1}{c|}{198.508}   & \multicolumn{1}{c|}{221.965}   & 293.365    &                                & \multicolumn{1}{c|}{0.00896}   & \multicolumn{1}{c|}{0.00913}   & 0.0132     \\ \hline
\end{tabular}
}
\end{table*}

\begin{table*}[!ht]
\centering
\renewcommand{\arraystretch}{1.175}
\caption{Pointwise forecasting error on irregular WADI and Traffic  datasetsin terms of MAE}
\label{tab:missingtable_5}
\vspace{-1mm}
\setlength{\tabcolsep}{5pt}
\resizebox{0.85\textwidth}{!}{%
\begin{tabular}{c|c|c|ccc|c|ccc}
\hline
\multirow{2}{*}{\textbf{\begin{tabular}[c]{@{}c@{}}Missing\\      Rate\end{tabular}}} & \multirow{2}{*}{\textbf{Model}} & \multirow{9}{*}{\rotatebox[origin=c]{90}{\textbf{WADI}}} & \multicolumn{3}{c|}{\textbf{MAE}}                                            & \multirow{9}{*}{\rotatebox[origin=c]{90}{\textbf{Traffic}}} & \multicolumn{3}{c}{\textbf{MAE}}                                             \\ \cline{4-6} \cline{8-10} 
                                                                                      &                                 &                                & \multicolumn{1}{c|}{Horizon@3} & \multicolumn{1}{c|}{Horizon@6} & Horizon@12 &                                   & \multicolumn{1}{c|}{Horizon@3} & \multicolumn{1}{c|}{Horizon@6} & Horizon@12 \\ \cline{1-2} \cline{4-6} \cline{8-10} 
\textbf{0\%}                                                                          & \textbf{JHgRF-Net}              &                                & \multicolumn{1}{c|}{4.149}     & \multicolumn{1}{c|}{4.592}     & 4.758      &                                   & \multicolumn{1}{c|}{0.006}     & \multicolumn{1}{c|}{0.009}     & 0.011      \\ \cline{1-2} \cline{4-6} \cline{8-10} 
\multirow{2}{*}{\textbf{10\%}}                                                        & \textbf{w/Point}                &                                & \multicolumn{1}{c|}{4.153}     & \multicolumn{1}{c|}{4.602}     & 4.761      &                                   & \multicolumn{1}{c|}{0.0063}    & \multicolumn{1}{c|}{0.0092}    & 0.0113     \\ \cline{2-2} \cline{4-6} \cline{8-10} 
                                                                                      & \textbf{w/Block}                &                                & \multicolumn{1}{c|}{4.155}     & \multicolumn{1}{c|}{4.608}     & 4.773      &                                   & \multicolumn{1}{c|}{0.0065}    & \multicolumn{1}{c|}{0.0095}    & 0.0117     \\ \cline{1-2} \cline{4-6} \cline{8-10} 
\multirow{2}{*}{\textbf{30\%}}                                                        & \textbf{w/Point}                &                                & \multicolumn{1}{c|}{4.236}     & \multicolumn{1}{c|}{4.703}     & 4.893      &                                   & \multicolumn{1}{c|}{0.00723}   & \multicolumn{1}{c|}{0.0103}    & 0.0123     \\ \cline{2-2} \cline{4-6} \cline{8-10} 
                                                                                      & \textbf{w/Block}                &                                & \multicolumn{1}{c|}{4.264}     & \multicolumn{1}{c|}{4.796}     & 4.913      &                                   & \multicolumn{1}{c|}{0.00768}   & \multicolumn{1}{c|}{0.0118}    & 0.0133     \\ \cline{1-2} \cline{4-6} \cline{8-10} 
\multirow{2}{*}{\textbf{50\%}}                                                        & \textbf{w/Point}                &                                & \multicolumn{1}{c|}{4.286}     & \multicolumn{1}{c|}{4.811}     & 5.031      &                                   & \multicolumn{1}{c|}{0.00898}   & \multicolumn{1}{c|}{0.0163}    & 0.0138     \\ \cline{2-2} \cline{4-6} \cline{8-10} 
                                                                                      & \textbf{w/Block}                &                                & \multicolumn{1}{c|}{4.293}     & \multicolumn{1}{c|}{4.977}     & 5.113      &                                   & \multicolumn{1}{c|}{0.00955}   & \multicolumn{1}{c|}{0.0171}    & 0.0145     \\ \hline
\end{tabular}
}
\end{table*}

\vspace{2mm}
\subsection{Sensitivity analysis}
\label{sec:hyperparameters}
We carried out a hyperparameter study to evaluate the impact of specific hyperparameters on the proposed framework performance. Our aim was to find the ideal set of hyperparameter values that could result in achieving the best possible performance on the benchmark datasets. We tuned four hyperparameters - embedding size($\textit{d}$), number of hyperedges($|\mathbf{E}|$), batch size($\textit{b}$), and learning rate($\textit{lr}$) - within specific ranges of values. The opted ranges were as follows: $\textit{d}$ $\in \{2, 6, 10, 18, 24\}$, $|\mathbf{E}|$ $\in \{2, 5, 8\}$, $\textit{b}$ $ \in \{2, 6, 10, 18, 24, 32, 64\}$, and $\textit{lr}$ $ \in \{\num{1e-1}, \num{1e-2}, \num{1e-3}, \num{1e-4}\}$. We have carefully selected ranges for the hyperparameters to prevent memory errors and limit model size. We optimized the framework hyperparameters through grid search and measured the model performance by measuring metrics like MAE and RMSE. These experimental results offered valuable insights into the effect of these hyperparameters on the framework ability to produce accurate forecasts in multivariate time series analysis, enhancing our understanding of its overall performance. The optimal hyperparameter configurations that yielded the best performance for each dataset are presented below,

\vspace{0mm}  
\begin{itemize}
    \item For PeMSD3, we set the batch size($\textit{b}$) to 18, the initial learning rate($\textit{lr}$) to \num{1e-3}, and the embedding size($\textit{d}$) to 18. Additionally, the number of hyperedges is 5.    
    \item For PeMSD4, we set the batch size($\textit{b}$) to 32, the initial learning rate($\textit{lr}$) to \num{1e-3}, and the embedding size($\textit{d}$) to 18. Additionally, the number of hyperedges is 5.    
    \item For PeMSD7, we set the batch size($\textit{b}$) to 6, the initial learning rate($\textit{lr}$) to \num{1e-3}, and the embedding size($\textit{d}$) to 18. Additionally, the number of hyperedges is 6.    
    \item For PeMSD8, we set the batch size($\textit{b}$) to 48, the initial learning rate($\textit{lr}$) to \num{1e-3}, and the embedding size($\textit{d}$) to 18. Additionally, the number of hyperedges is 8.     
    \item For PeMSD7(M), we set the batch size($\textit{b}$) to 48, the initial learning rate($\textit{lr}$) to \num{1e-3}, and the embedding size($\textit{d}$) to 18. The number of hyperedges is 6. 
    \item For METR-LA, we set the batch size($\textit{b}$) to 48, the initial learning rate($\textit{lr}$) to \num{1e-3}, and the embedding size($\textit{d}$) to 18. Additionally, the number of hyperedges is 5. 
    \item For PEMS-BAY, we set the batch size($\textit{b}$) to 12, the initial learning rate($\textit{lr}$) to \num{1e-3}, and the embedding size($\textit{d}$) to 18. The number of hyperedges is 5. 
    \item For SWAT, we set the batch size($\textit{b}$) to 256, the initial learning rate($\textit{lr}$) to \num{1e-3}, and the embedding size($\textit{d}$) to 18. Additionally, the number of hyperedges is 5. 
    \item For WADI, we set the batch size($\textit{b}$) to 64, the initial learning rate($\textit{lr}$) to \num{1e-3}, and the embedding size($\textit{d}$) to 12. Additionally, the number of hyperedges is 5. 
    \item For Electricity, we set the batch size($\textit{b}$) to 32, the initial learning rate($\textit{lr}$) to \num{1e-3}, and the embedding size($\textit{d}$) to 18. Additionally, the number of hyperedges is 2. 
    \item For Solar-energy, we set the batch size($\textit{b}$) to 32, the initial learning rate($\textit{lr}$) to \num{1e-3}, and the embedding size($\textit{d}$) to 18. The number of hyperedges is 6. 
    \item For Exchange-rate, we set the batch size($\textit{b}$) to 32, the initial learning rate($\textit{lr}$) to \num{1e-3}, and the embedding size($\textit{d}$) to 18.  The number of hyperedges is 6. 
    \item For Traffic, we set the batch size($\textit{b}$) to 8, the initial learning rate($\textit{lr}$) to \num{1e-3}, and the embedding size($\textit{d}$) to 18. Additionally, the number of hyperedges is 5. 
\end{itemize}

The proportion of hypernodes connected to hyperedges in a hypergraph indicates the network's ``edge density". A higher fraction of connected hypernodes suggests a denser network, while a lower fraction implies a sparser network. Modifying the number of hyperedges enables control over the hypergraph's density. The hyperparameter study yields the optimal number of hyperedges for an MTSF task by evaluating the impact of the number of predefined hyperedges on the learned hypergraph structures. This study sheds light on how the hypergraph's density changes as the number of hyperedges increases or decreases for a particular dataset in the MTSF task, with Table \ref{tab:hyperparameter} presenting the experimental results.

\begin{table*}[!h]
\centering
\caption{Experimental results of the hyperparameter study on the benchmark datasets.}
\label{tab:hyperparameter}
\vspace{0mm}
\resizebox{1.1\textwidth}{!}{%
\hspace{-5mm}\begin{tabular}{cll|c|c|c|c|c|c|c|cc|cccccc}
\hline
\multicolumn{3}{c|}{\textbf{Hyperparameter}} &
  \multicolumn{1}{l|}{} &
  \textbf{RMSE} &
  \textbf{MAE} &
  \multicolumn{1}{l|}{\textbf{}} &
  \textbf{RMSE} &
  \textbf{MAE} &
  \multicolumn{1}{l|}{\textbf{}} &
  \textbf{RMSE} &
  \textbf{MAE} &
  \multicolumn{1}{l|}{\textbf{}} &
  \multicolumn{1}{c|}{\textbf{RMSE}} &
  \multicolumn{1}{c|}{\textbf{MAE}} &
  \multicolumn{1}{l|}{\textbf{}} &
  \multicolumn{1}{c|}{\textbf{RMSE}} &
  \textbf{MAE} \\ \hline
\multicolumn{2}{c|}{\multirow{4}{*}{\textbf{\begin{tabular}[c]{@{}c@{}}Embedded \\      Dimension\end{tabular}}}} &
  \textbf{2} &
  \multirow{8}{*}{\textbf{\rotatebox[origin=c]{90}{\textbf{METR-LA}}}} &
  9.4477 &
  5.56 &
  \multirow{8}{*}{\textbf{\rotatebox[origin=c]{90}{\textbf{Solar Energy}}}} &
  3.6908 &
  2.4007 &
  \multirow{8}{*}{\textbf{\rotatebox[origin=c]{90}{\textbf{SWaT}}}} &
  1.8902 &
  0.435 &
  \multicolumn{1}{c|}{\multirow{8}{*}{\textbf{\rotatebox[origin=c]{90}{\textbf{WADI}}}}} &
  \multicolumn{1}{c|}{54.8259} &
  \multicolumn{1}{c|}{88.1317} &
  \multicolumn{1}{c|}{\multirow{8}{*}{\textbf{\rotatebox[origin=c]{90}{\textbf{Traffic}}}}} &
  \multicolumn{1}{c|}{0.0276} &
  0.0276 \\ \cline{3-3} \cline{5-6} \cline{8-9} \cline{11-12} \cline{14-15} \cline{17-18} 
\multicolumn{2}{c|}{} &
  \textbf{6} &
   &
  9.0538 &
  5.1174 &
   &
  2.5359 &
  1.5693 &
   &
  0.8783 &
  0.2213 &
  \multicolumn{1}{c|}{} &
  \multicolumn{1}{c|}{43.879} &
  \multicolumn{1}{c|}{5.0819} &
  \multicolumn{1}{c|}{} &
  \multicolumn{1}{c|}{0.0207} &
  0.0121 \\ \cline{3-3} \cline{5-6} \cline{8-9} \cline{11-12} \cline{14-15} \cline{17-18} 
\multicolumn{2}{c|}{} &
  \textbf{12} &
   &
  8.9438 &
  5.1621 &
   &
  2.3676 &
  1.4146 &
   &
  0.7175 &
  0.1839 &
  \multicolumn{1}{c|}{} &
  \multicolumn{1}{c|}{43.0202} &
  \multicolumn{1}{c|}{4.691} &
  \multicolumn{1}{c|}{} &
  \multicolumn{1}{c|}{0.0197} &
  0.0114 \\ \cline{3-3} \cline{5-6} \cline{8-9} \cline{11-12} \cline{14-15} \cline{17-18} 
\multicolumn{2}{c|}{} &
  \textbf{18} &
   &
  8.8559 &
  5.0357 &
   &
  2.3268 &
  1.3956 &
   &
  0.7087 &
  0.168 &
  \multicolumn{1}{c|}{} &
  \multicolumn{1}{c|}{42.9041} &
  \multicolumn{1}{c|}{4.766} &
  \multicolumn{1}{c|}{} &
  \multicolumn{1}{c|}{0.0194} &
  0.0112 \\ \cline{1-3} \cline{5-6} \cline{8-9} \cline{11-12} \cline{14-15} \cline{17-18} 
\multicolumn{2}{c|}{\multirow{4}{*}{\textbf{\begin{tabular}[c]{@{}c@{}}Number of \\      Hyperedges\end{tabular}}}} &
  \textbf{2} &
   &
  8.8461 &
  5.0311 &
   &
  2.3598 &
  1.4037 &
   &
  0.7733 &
  0.1897 &
  \multicolumn{1}{c|}{} &
  \multicolumn{1}{c|}{43.4547} &
  \multicolumn{1}{c|}{5.0108} &
  \multicolumn{1}{c|}{} &
  \multicolumn{1}{c|}{0.0193} &
  0.0111 \\ \cline{3-3} \cline{5-6} \cline{8-9} \cline{11-12} \cline{14-15} \cline{17-18} 
\multicolumn{2}{c|}{} &
  \textbf{5} &
   &
  8.8559 &
  5.0365 &
   &
  2.3261 &
  1.3946 &
   &
  0.7177 &
  0.1839 &
  \multicolumn{1}{c|}{} &
  \multicolumn{1}{c|}{43.112} &
  \multicolumn{1}{c|}{4.7899} &
  \multicolumn{1}{c|}{} &
  \multicolumn{1}{c|}{0.0194} &
  0.0111 \\ \cline{3-3} \cline{5-6} \cline{8-9} \cline{11-12} \cline{14-15} \cline{17-18} 
\multicolumn{2}{c|}{} &
  \textbf{6} &
   &
  8.8487 &
  5.0837 &
   &
  2.2706 &
  1.3412 &
   &
  0.7764 &
  0.1907 &
  \multicolumn{1}{c|}{} &
  \multicolumn{1}{c|}{43.9438} &
  \multicolumn{1}{c|}{5.1023} &
  \multicolumn{1}{c|}{} &
  \multicolumn{1}{c|}{0.0194} &
  0.0111 \\ \cline{3-3} \cline{5-6} \cline{8-9} \cline{11-12} \cline{14-15} \cline{17-18} 
\multicolumn{2}{c|}{} &
  \textbf{8} &
   &
  8.8794 &
  5.0472 &
   &
  2.3323 &
  1.3982 &
   &
  0.7594 &
  0.7594 &
  \multicolumn{1}{c|}{} &
  \multicolumn{1}{c|}{42.8957} &
  \multicolumn{1}{c|}{4.7274} &
  \multicolumn{1}{c|}{} &
  \multicolumn{1}{c|}{0.0195} &
  0.0112 \\ \hline
\multicolumn{2}{c|}{\multirow{4}{*}{\textbf{\begin{tabular}[c]{@{}c@{}}Embedded \\      Dimension\end{tabular}}}} &
  \textbf{2} &
  \multirow{8}{*}{\textbf{\rotatebox[origin=c]{90}{\textbf{PeMSD3}}}} &
  29.8281 &
  19.3176 &
  \multirow{8}{*}{\textbf{\rotatebox[origin=c]{90}{\textbf{PeMSD4}}}} &
  41.3629 &
  29.3875 &
  \multirow{8}{*}{\textbf{\rotatebox[origin=c]{90}{\textbf{PeMSD7}}}} &
  39.879 &
  26.6195 &
  \multicolumn{1}{c|}{\multirow{8}{*}{\textbf{\rotatebox[origin=c]{90}{\textbf{PeMSD8}}}}} &
  \multicolumn{1}{c|}{29.7079} &
  \multicolumn{1}{c|}{20.4021} &
  \multicolumn{1}{c|}{\multirow{8}{*}{\textbf{\rotatebox[origin=c]{90}{\textbf{PeMSD7(M)}}}}} &
  \multicolumn{1}{c|}{6.1256} &
  3.8583 \\ \cline{3-3} \cline{5-6} \cline{8-9} \cline{11-12} \cline{14-15} \cline{17-18} 
\multicolumn{2}{c|}{} &
  \textbf{6} &
   &
  29.938 &
  15.9012 &
   &
  32.521 &
  21.7021 &
   &
  32.8907 &
  22.1893 &
  \multicolumn{1}{c|}{} &
  \multicolumn{1}{c|}{25.7841} &
  \multicolumn{1}{c|}{16.9532} &
  \multicolumn{1}{c|}{} &
  \multicolumn{1}{c|}{5.492} &
  3.1196 \\ \cline{3-3} \cline{5-6} \cline{8-9} \cline{11-12} \cline{14-15} \cline{17-18} 
\multicolumn{2}{c|}{} &
  \textbf{12} &
   &
  22.1569 &
  14.4857 &
   &
  29.7458 &
  19.9336 &
   &
  31.9132 &
  21.4281 &
  \multicolumn{1}{c|}{} &
  \multicolumn{1}{c|}{23.658} &
  \multicolumn{1}{c|}{15.3998} &
  \multicolumn{1}{c|}{} &
  \multicolumn{1}{c|}{5.2024} &
  2.9499 \\ \cline{3-3} \cline{5-6} \cline{8-9} \cline{11-12} \cline{14-15} \cline{17-18} 
\multicolumn{2}{c|}{} &
  \textbf{18} &
   &
  21.5551 &
  14.2369 &
   &
  28.7556 &
  19.2327 &
   &
  30.9304 &
  20.6119 &
  \multicolumn{1}{c|}{} &
  \multicolumn{1}{c|}{22.6019} &
  \multicolumn{1}{c|}{14.6955} &
  \multicolumn{1}{c|}{} &
  \multicolumn{1}{c|}{5.0874} &
  2.9073 \\ \cline{1-3} \cline{5-6} \cline{8-9} \cline{11-12} \cline{14-15} \cline{17-18} 
\multicolumn{2}{c|}{\multirow{4}{*}{\textbf{\begin{tabular}[c]{@{}c@{}}Number of \\      Hyperedges\end{tabular}}}} &
  \textbf{2} &
   &
  21.5534 &
  14.3112 &
   &
  29.1418 &
  19.4152 &
   &
  31.4684 &
  21.0394 &
  \multicolumn{1}{c|}{} &
  \multicolumn{1}{c|}{23.042} &
  \multicolumn{1}{c|}{15.0624} &
  \multicolumn{1}{c|}{} &
  \multicolumn{1}{c|}{5.1011} &
  2.9132 \\ \cline{3-3} \cline{5-6} \cline{8-9} \cline{11-12} \cline{14-15} \cline{17-18} 
\multicolumn{2}{c|}{} &
  \textbf{5} &
   &
  21.5748 &
  14.243 &
   &
  28.768 &
  19.2375 &
   &
  31.0854 &
  20.7662 &
  \multicolumn{1}{c|}{} &
  \multicolumn{1}{c|}{22.6021} &
  \multicolumn{1}{c|}{14.6956} &
  \multicolumn{1}{c|}{} &
  \multicolumn{1}{c|}{5.0866} &
  2.9067 \\ \cline{3-3} \cline{5-6} \cline{8-9} \cline{11-12} \cline{14-15} \cline{17-18} 
\multicolumn{2}{c|}{} &
  \textbf{6} &
   &
  21.646 &
  14.2859 &
   &
  29.1065 &
  19.4 &
   &
  31.0845 &
  20.7209 &
  \multicolumn{1}{c|}{} &
  \multicolumn{1}{c|}{22.558} &
  \multicolumn{1}{c|}{14.6561} &
  \multicolumn{1}{c|}{} &
  \multicolumn{1}{c|}{5.0333} &
  2.8265 \\ \cline{3-3} \cline{5-6} \cline{8-9} \cline{11-12} \cline{14-15} \cline{17-18} 
\multicolumn{2}{c|}{} &
  \textbf{8} &
   &
  21.8008 &
  14.3245 &
   &
  28.6527 &
  19.1951 &
   &
  31.7462 &
  21.2531 &
  \multicolumn{1}{c|}{} &
  \multicolumn{1}{c|}{22.3803} &
  \multicolumn{1}{c|}{14.5223} &
  \multicolumn{1}{c|}{} &
  \multicolumn{1}{c|}{5.0403} &
  2.8618 \\ \hline
\multicolumn{2}{c|}{\multirow{4}{*}{\textbf{\begin{tabular}[c]{@{}c@{}}Embedded \\      Dimension\end{tabular}}}} &
  \textbf{2} &
  \multirow{8}{*}{\textbf{\rotatebox[origin=c]{90}{\textbf{Exchange Rate}}}} &
  0.037 &
  0.0267 &
  \multirow{8}{*}{\textbf{\rotatebox[origin=c]{90}{\textbf{Electricity}}}} &
  2806.454 &
  384.8923 &
  \multirow{8}{*}{\textbf{\rotatebox[origin=c]{90}{\textbf{PeMS-BAY}}}} &
  3.5032 &
  1.8714 &
  \multicolumn{6}{c}{\multirow{8}{*}{}} \\ \cline{3-3} \cline{5-6} \cline{8-9} \cline{11-12}
\multicolumn{2}{c|}{} &
  \textbf{6} &
   &
  0.0136 &
  0.0091 &
   &
  1504.247 &
  252.516 &
   &
  3.2833 &
  1.8136 &
  \multicolumn{6}{c}{} \\ \cline{3-3} \cline{5-6} \cline{8-9} \cline{11-12}
\multicolumn{2}{c|}{} &
  \textbf{12} &
   &
  0.0128 &
  0.0085 &
   &
  1323.743 &
  232.582 &
   &
  3.2317 &
  1.7753 &
  \multicolumn{6}{c}{} \\ \cline{3-3} \cline{5-6} \cline{8-9} \cline{11-12}
\multicolumn{2}{c|}{} &
  \textbf{18} &
   &
  0.0128 &
  0.0084 &
   &
  1304.031 &
  228.0085 &
   &
  3.2071 &
  1.7585 &
  \multicolumn{6}{c}{} \\ \cline{1-3} \cline{5-6} \cline{8-9} \cline{11-12}
\multicolumn{2}{c|}{\multirow{4}{*}{\textbf{\begin{tabular}[c]{@{}c@{}}Number of \\      Hyperedges\end{tabular}}}} &
  \textbf{2} &
   &
  0.0125 &
  0.0082 &
   &
  1222.109 &
  222.5563 &
   &
  3.2255 &
  1.7822 &
  \multicolumn{6}{c}{} \\ \cline{3-3} \cline{5-6} \cline{8-9} \cline{11-12}
\multicolumn{2}{c|}{} &
  \textbf{5} &
   &
  0.0128 &
  0.0084 &
   &
  1303.631 &
  227.898 &
   &
  3.2077 &
  1.7588 &
  \multicolumn{6}{c}{} \\ \cline{3-3} \cline{5-6} \cline{8-9} \cline{11-12}
\multicolumn{2}{c|}{} &
  \textbf{6} &
   &
  0.0124 &
  0.0081 &
   &
  1275.573 &
  224.6772 &
   &
  3.2198 &
  1.7646 &
  \multicolumn{6}{c}{} \\ \cline{3-3} \cline{5-6} \cline{8-9} \cline{11-12}
\multicolumn{2}{c|}{} &
  \textbf{8} &
   &
  0.0127 &
  0.0083 &
   &
  1262.914 &
  225.6226 &
   &
  3.2058 &
  1.7533 &
  \multicolumn{6}{c}{} \\ \cline{1-12}
\end{tabular}
}
\end{table*}

\subsection{Time series forecasting visualization}
\begin{figure*}[!htbp]
\centering
\subfloat[Node 12 in PeMSD3]{
  \includegraphics[width=45mm]{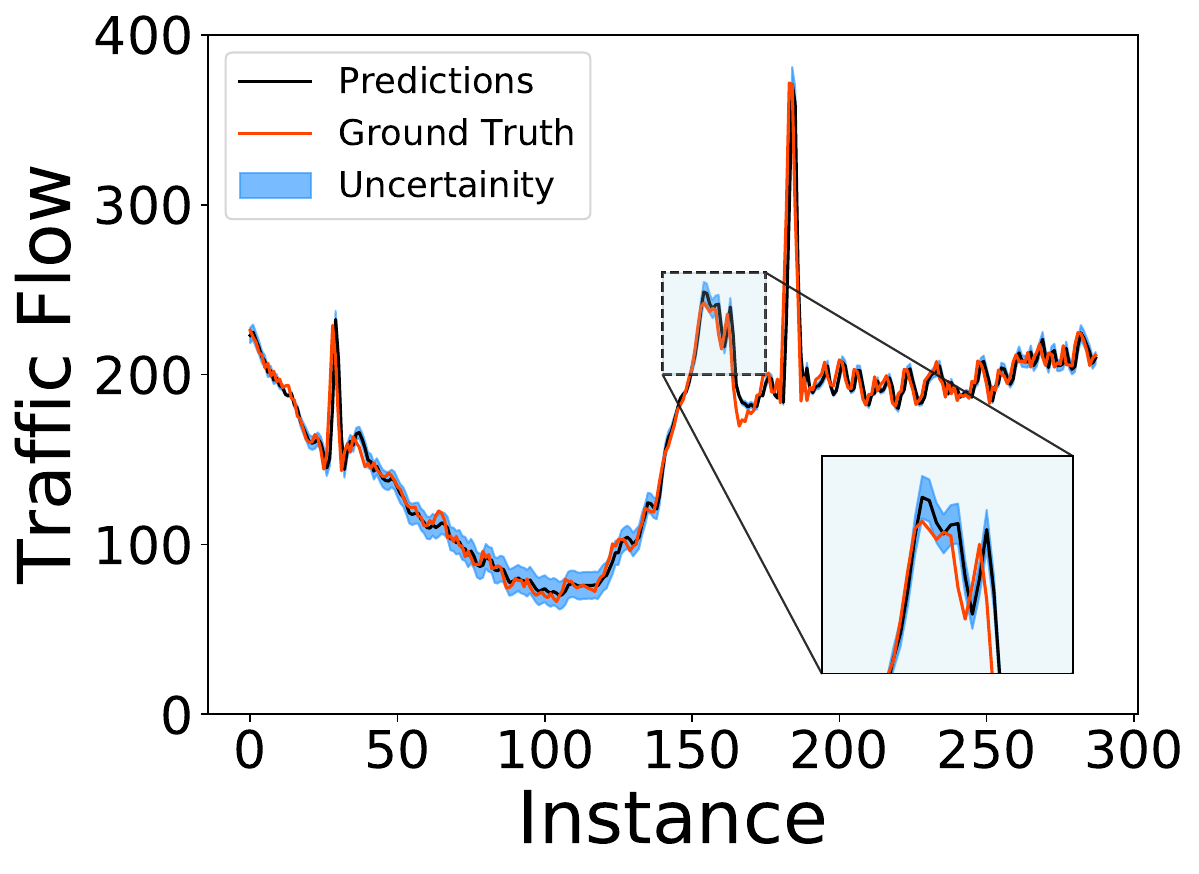}
}
\subfloat[Node 99 in PeMSD3]{
  \includegraphics[width=45mm]{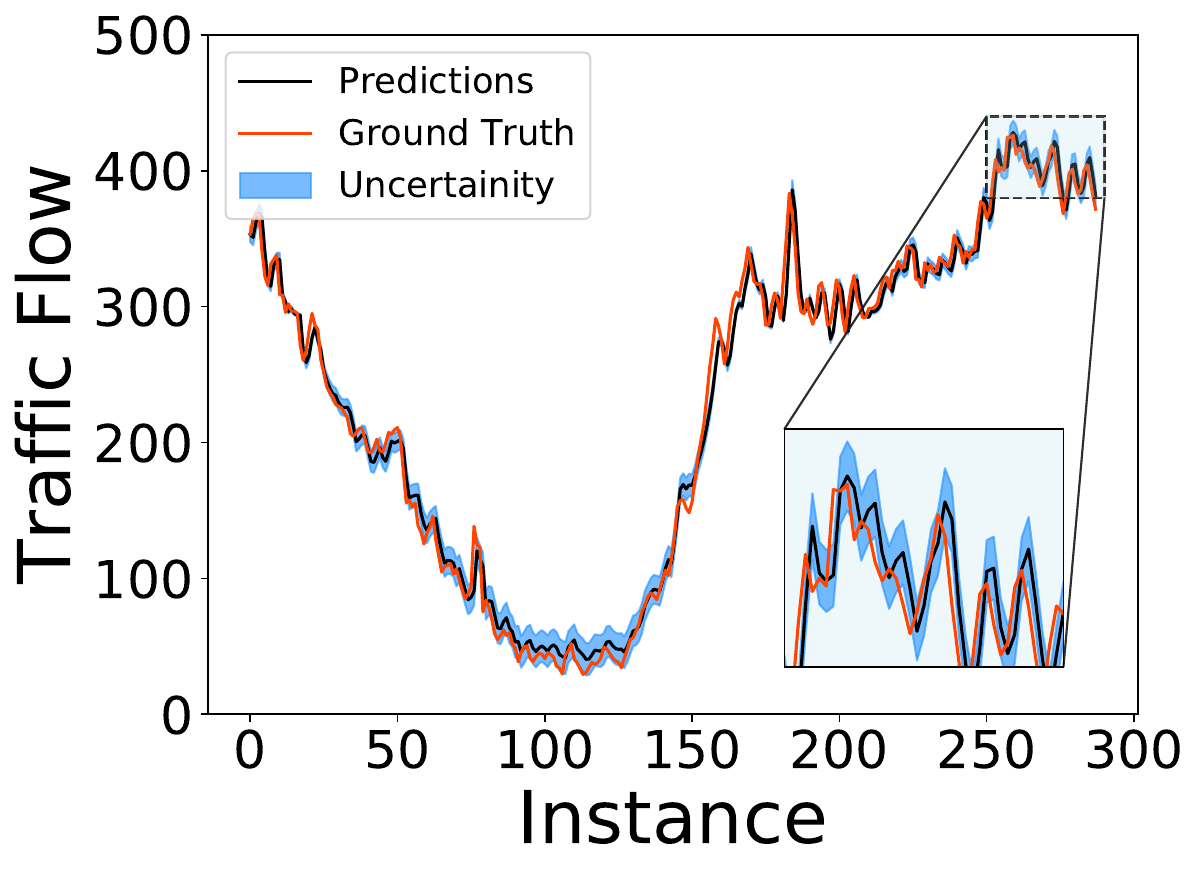}
}
\subfloat[Node 108 in PeMSD3]{
  \includegraphics[width=45mm]{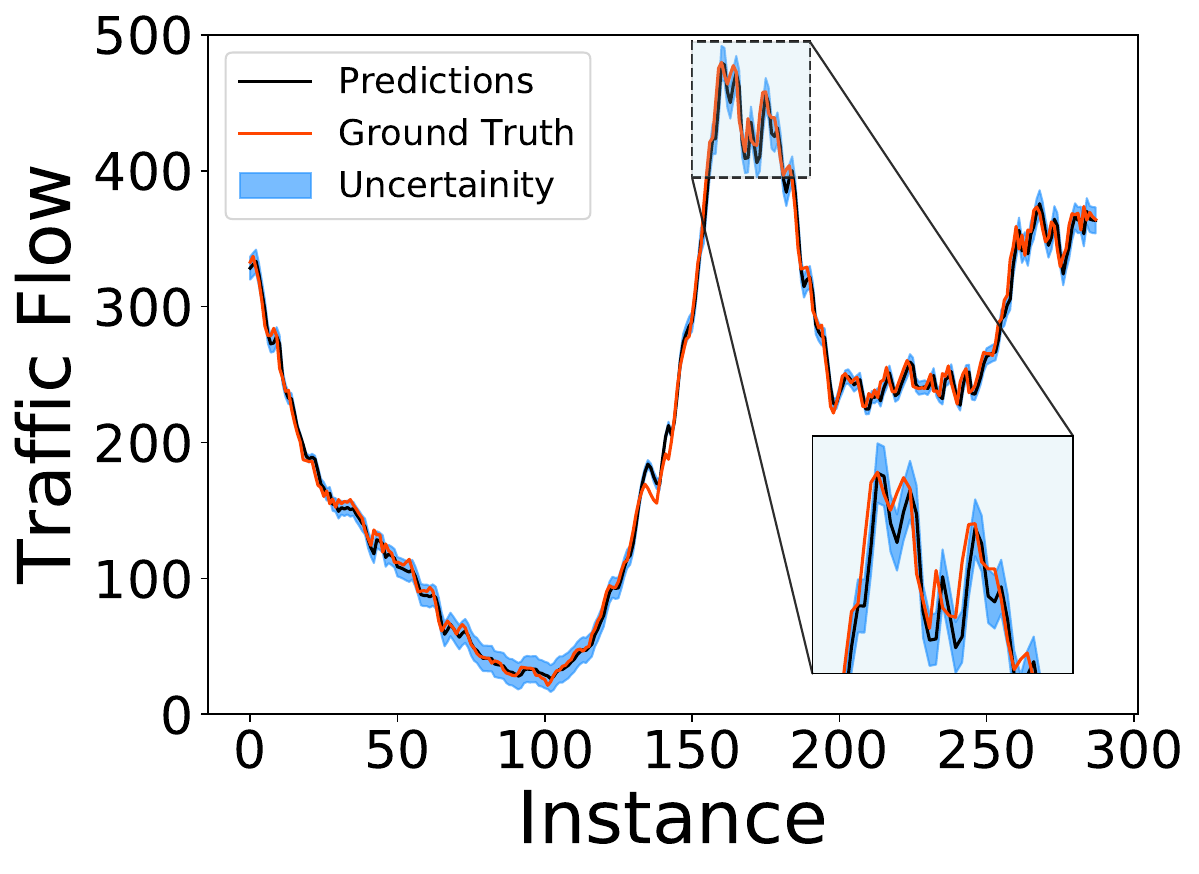}
}
\subfloat[Node 141 in PeMSD3]{
  \includegraphics[width=45mm]{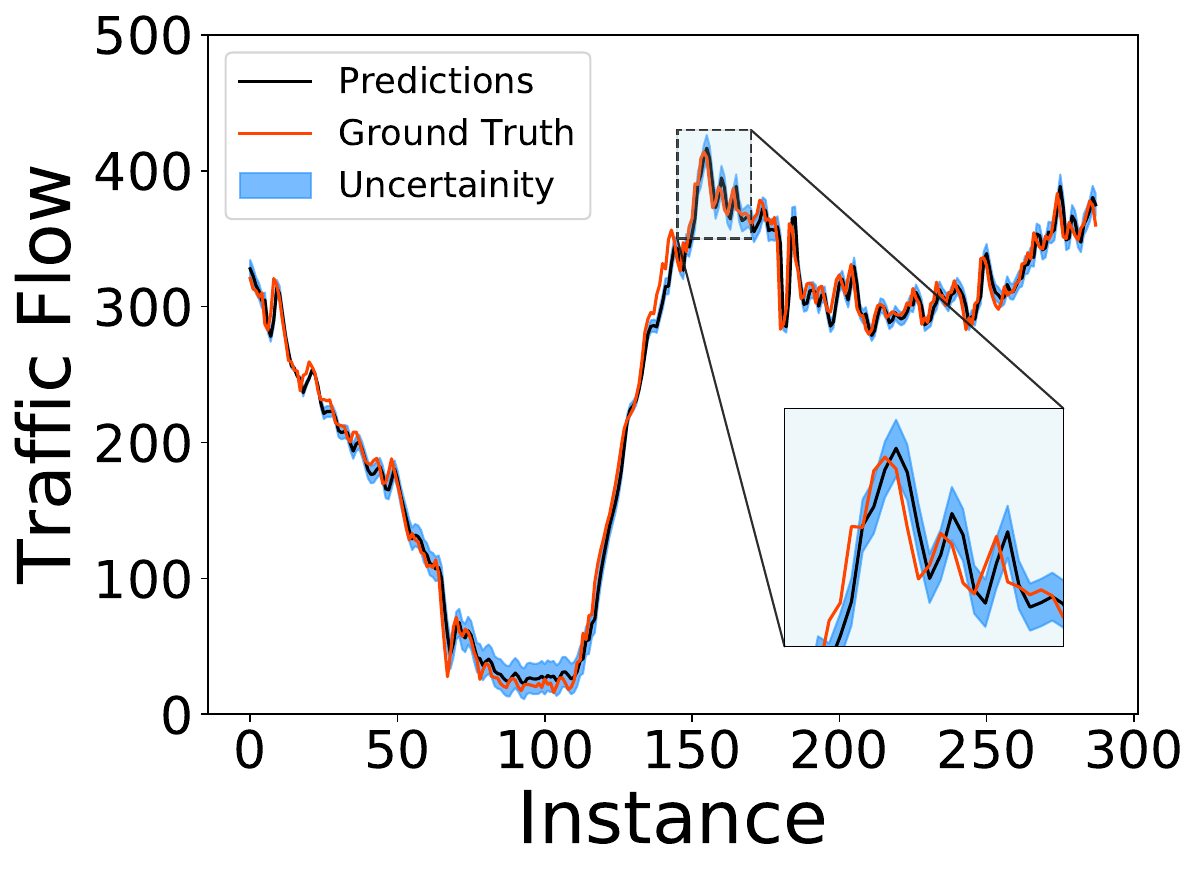}
} 
\hspace{0mm}
\subfloat[Node 149 in PeMSD4]{
  \includegraphics[width=45mm]{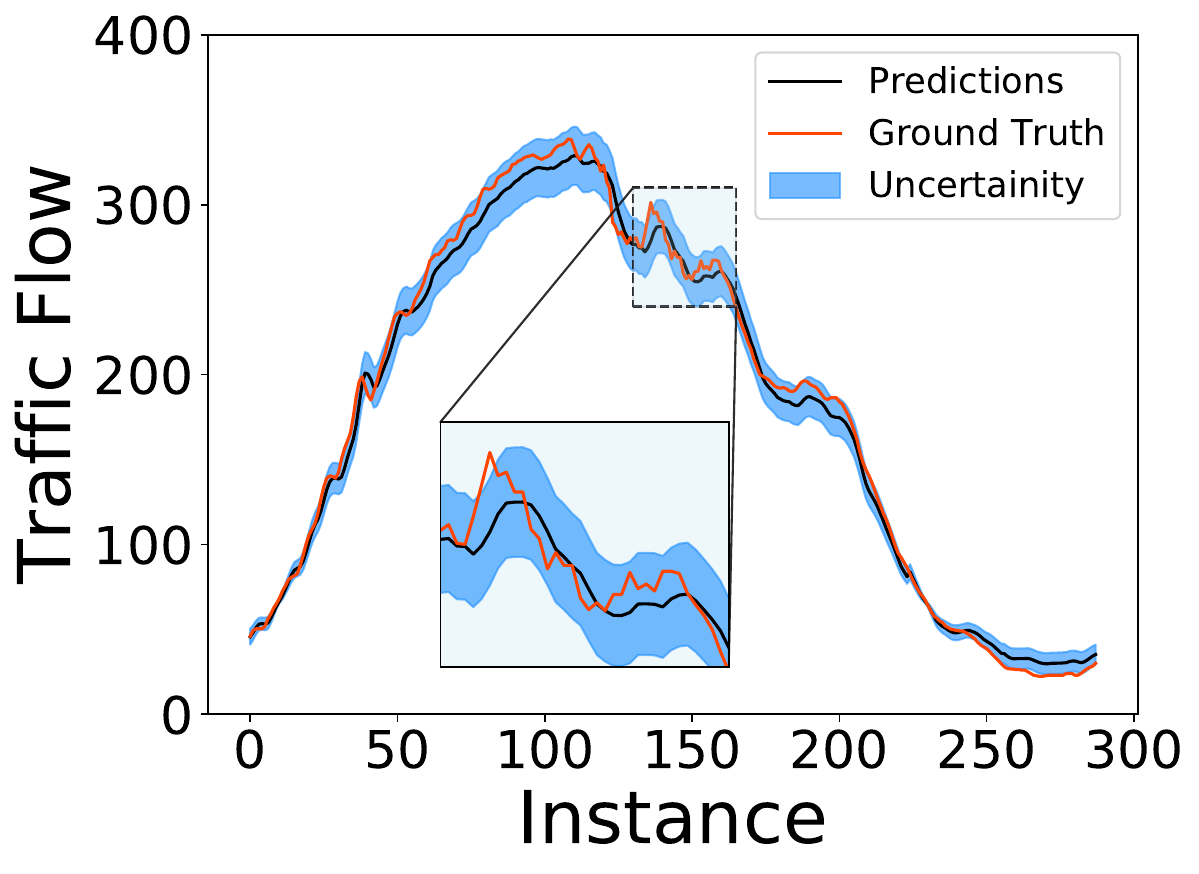}
}
\subfloat[Node 170 in PeMSD4]{
  \includegraphics[width=45mm]{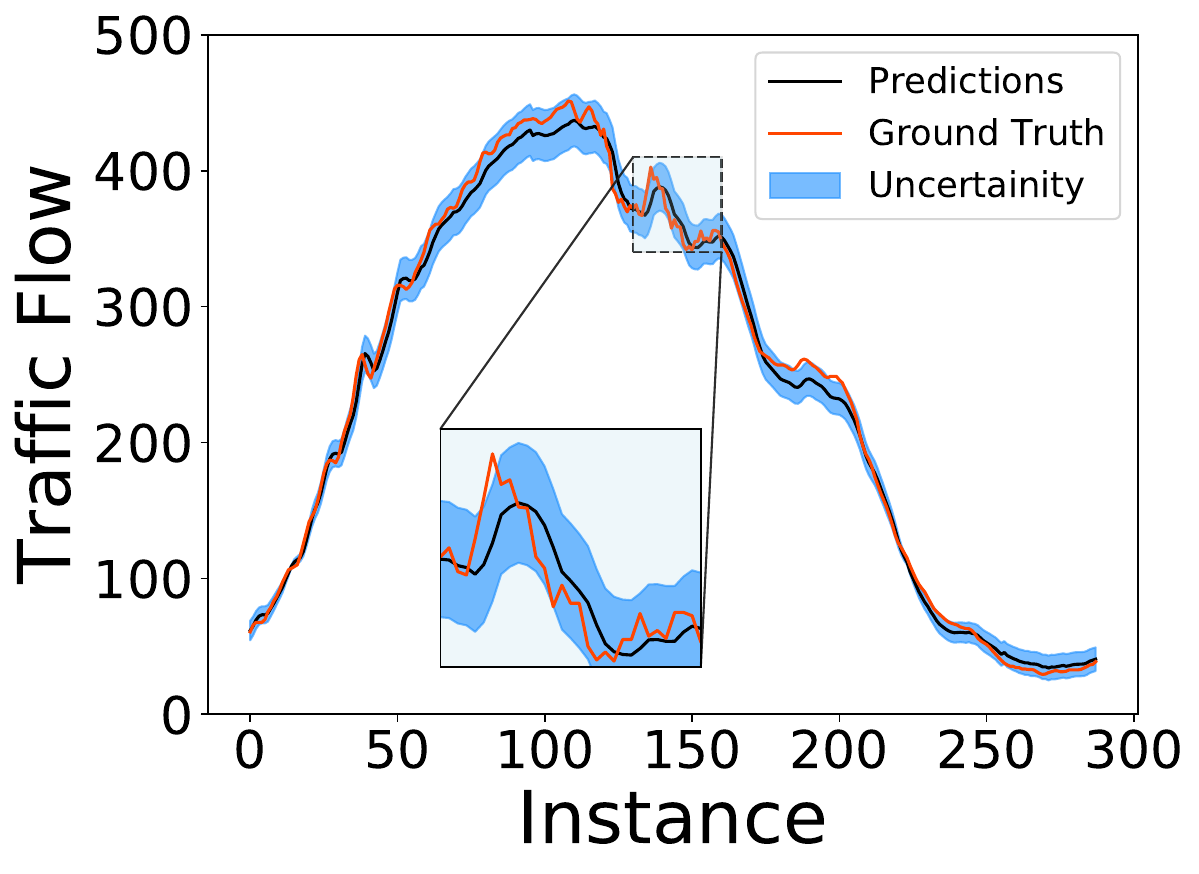}
}
\subfloat[Node 211 in PeMSD4]{
  \includegraphics[width=45mm]{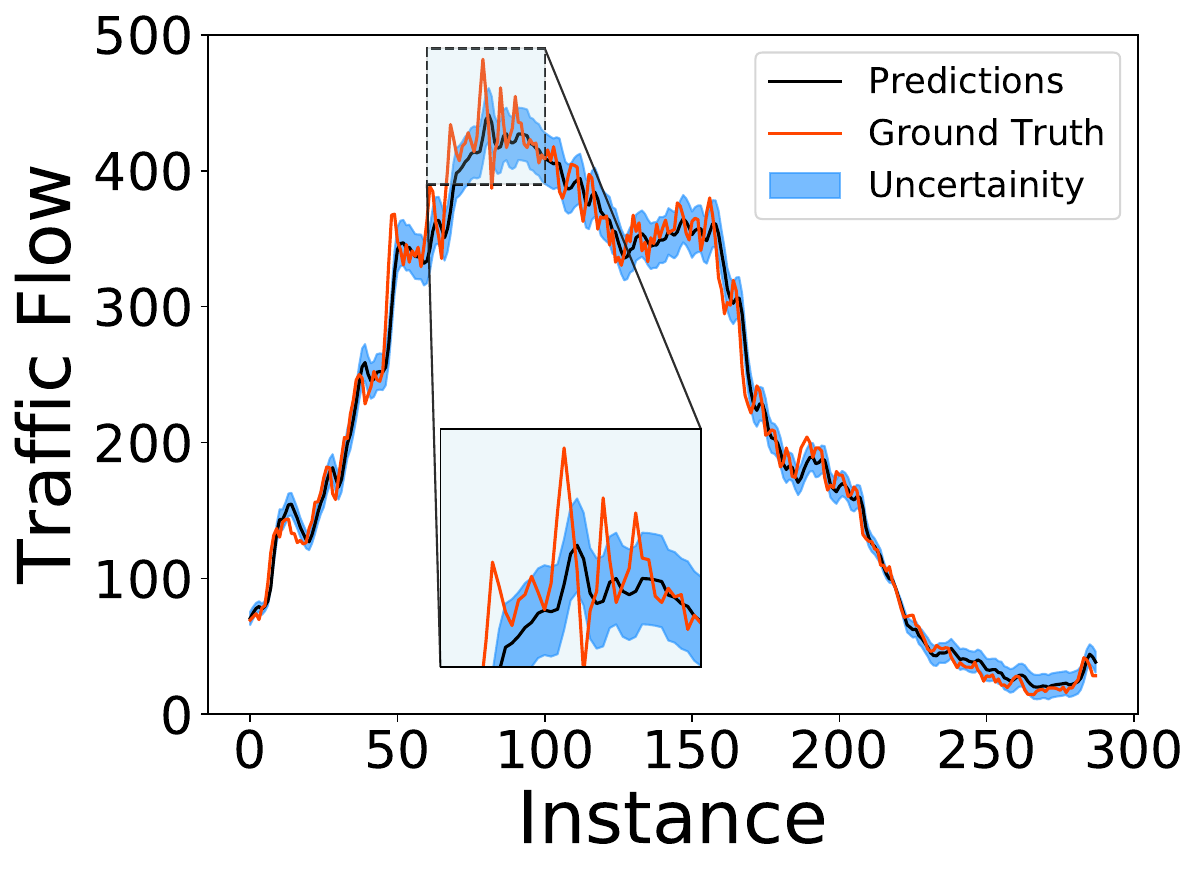}
}
\subfloat[Node 287 in PeMSD4]{
  \includegraphics[width=45mm]{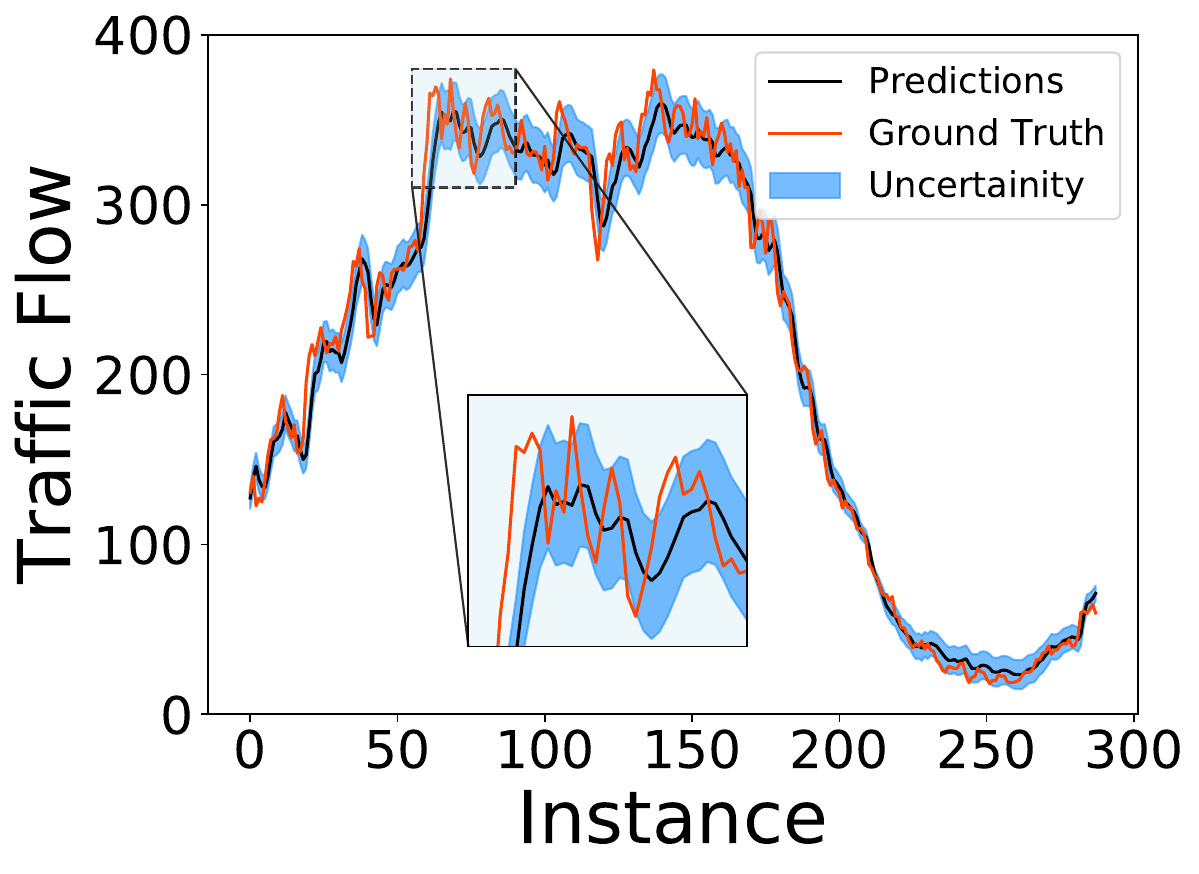}
} 
\hspace{0mm}
\subfloat[Node 85 in PeMSD8]{
  \includegraphics[width=45mm]{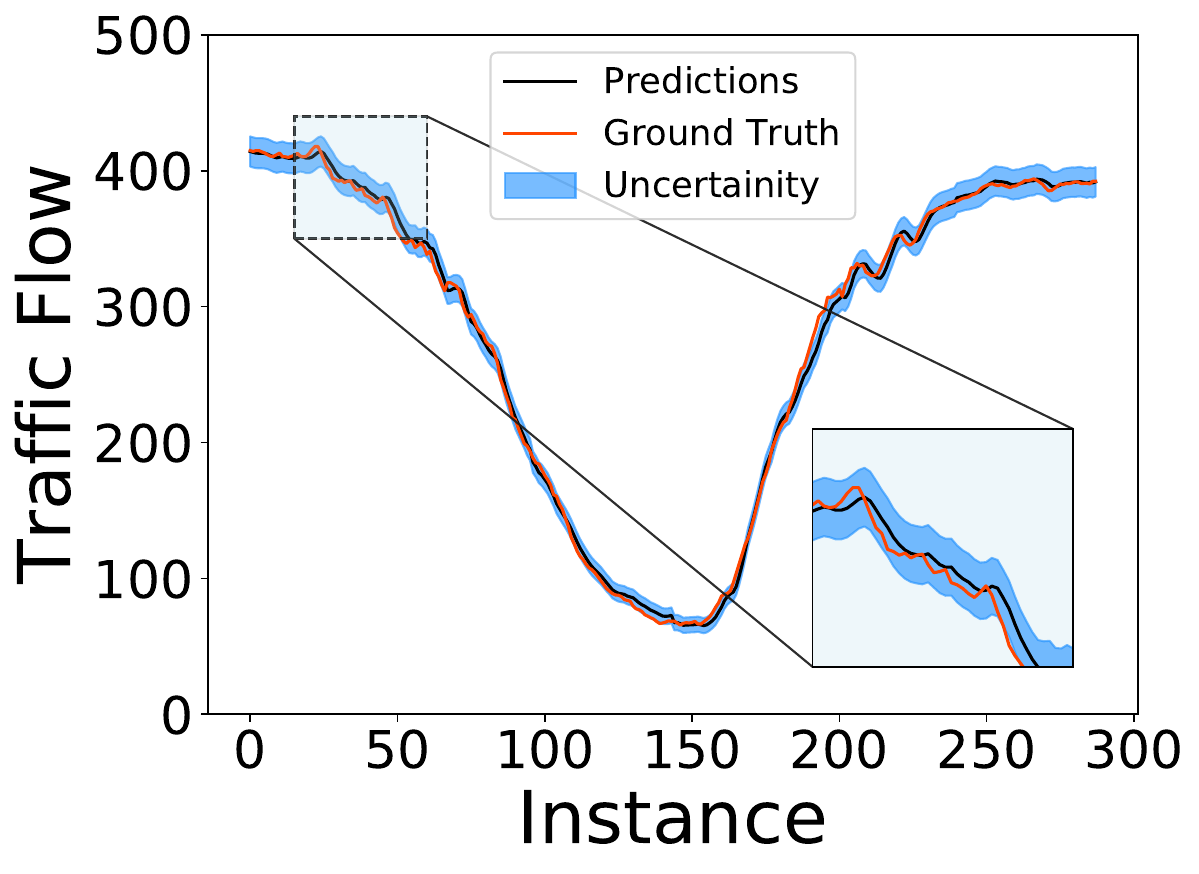}
}
\subfloat[Node 104 in PeMSD8]{
  \includegraphics[width=45mm]{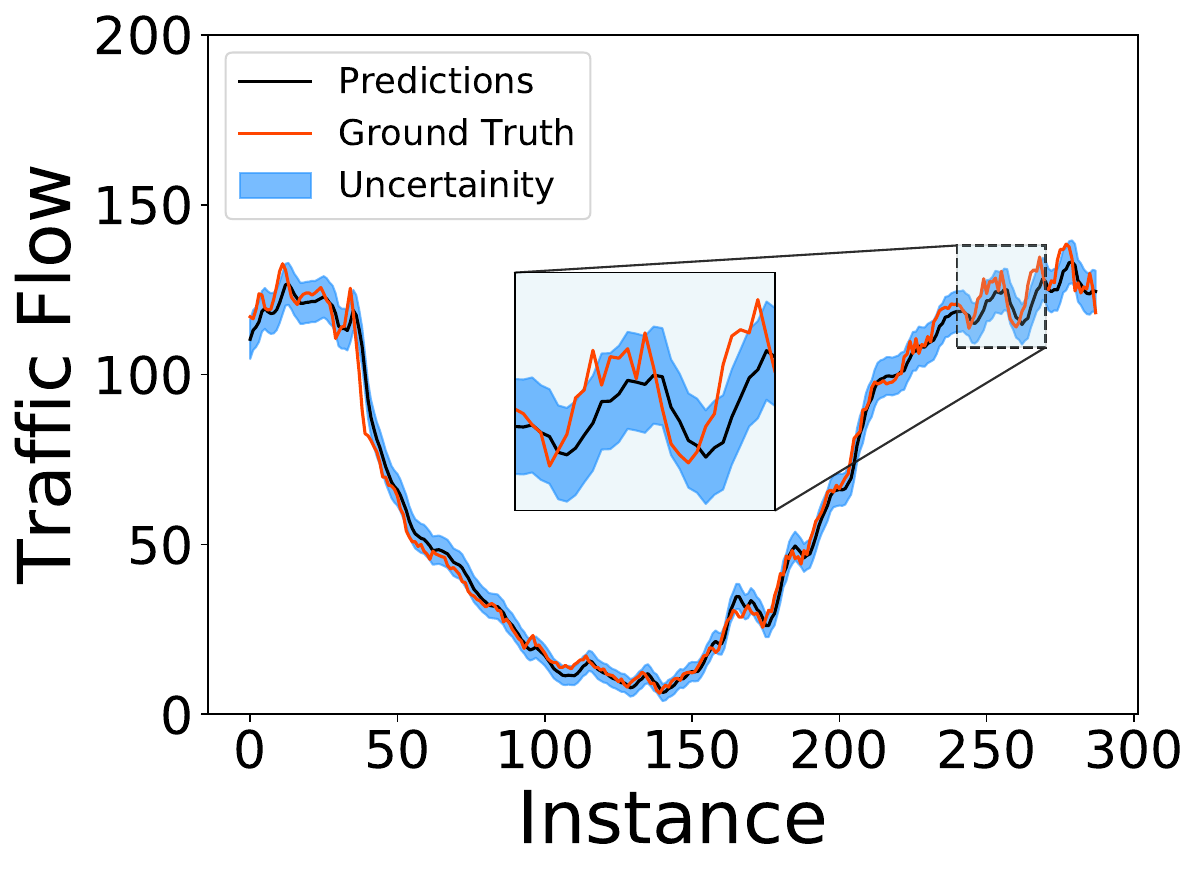}
}
\subfloat[Node 155 in PeMSD8]{
  \includegraphics[width=45mm]{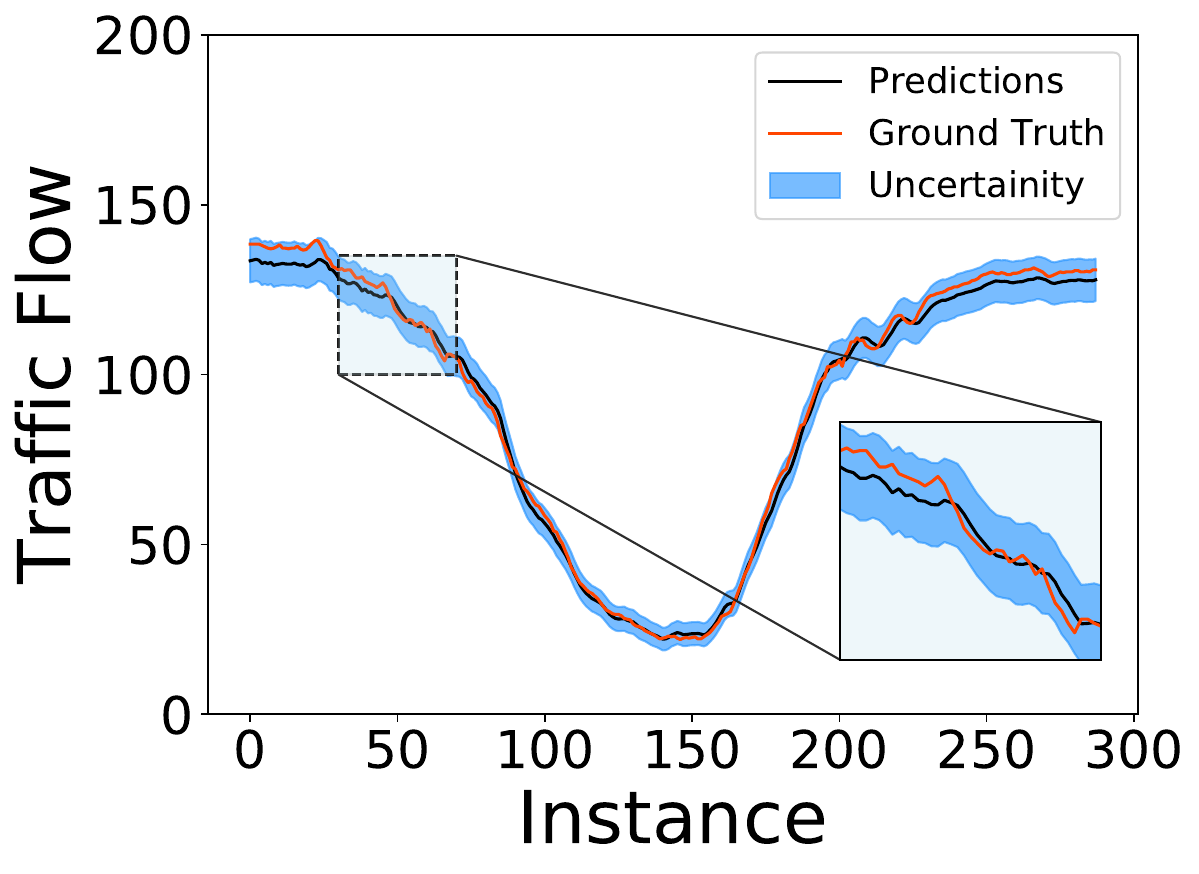}
}
\subfloat[Node 162 in PeMSD8]{
  \includegraphics[width=45mm]{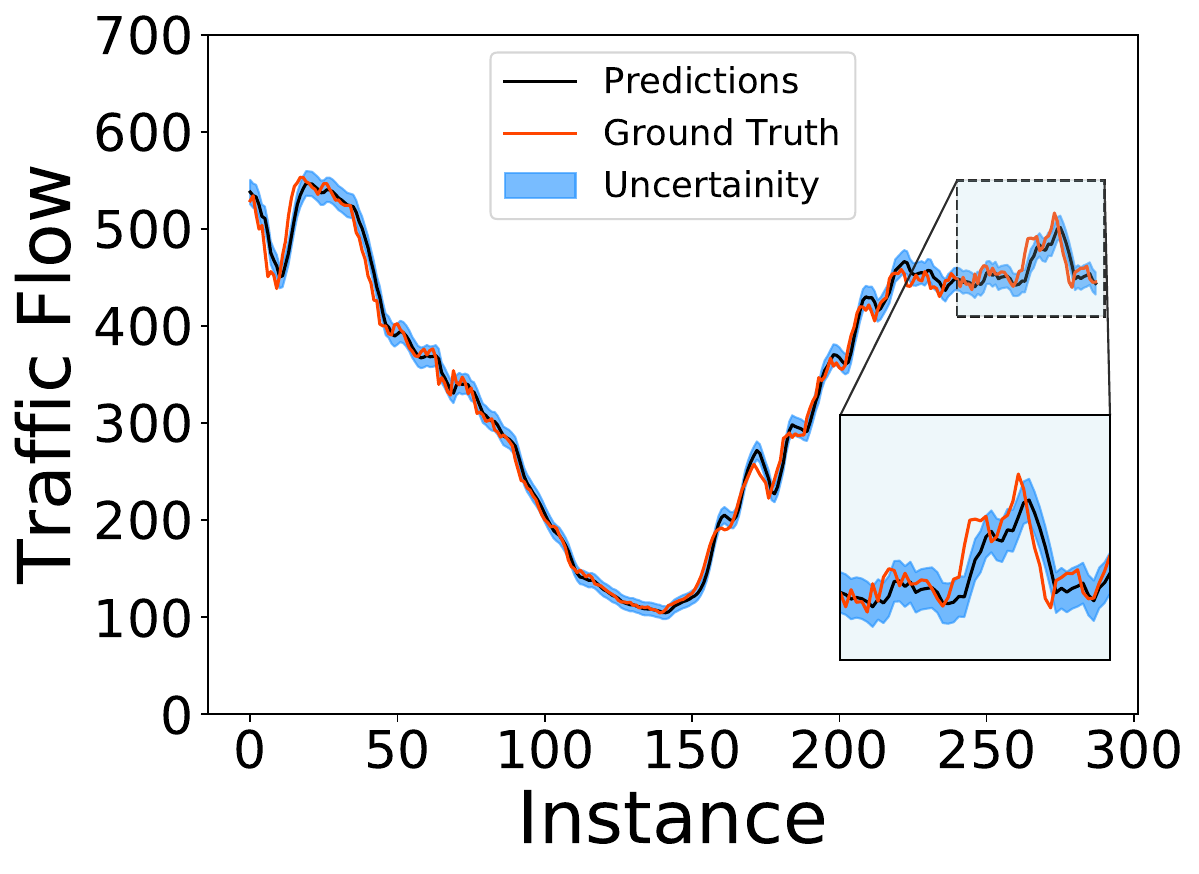}
 }
 \hspace{0mm}
 \subfloat[Node 16 in Electricity]{
  \includegraphics[width=45mm]{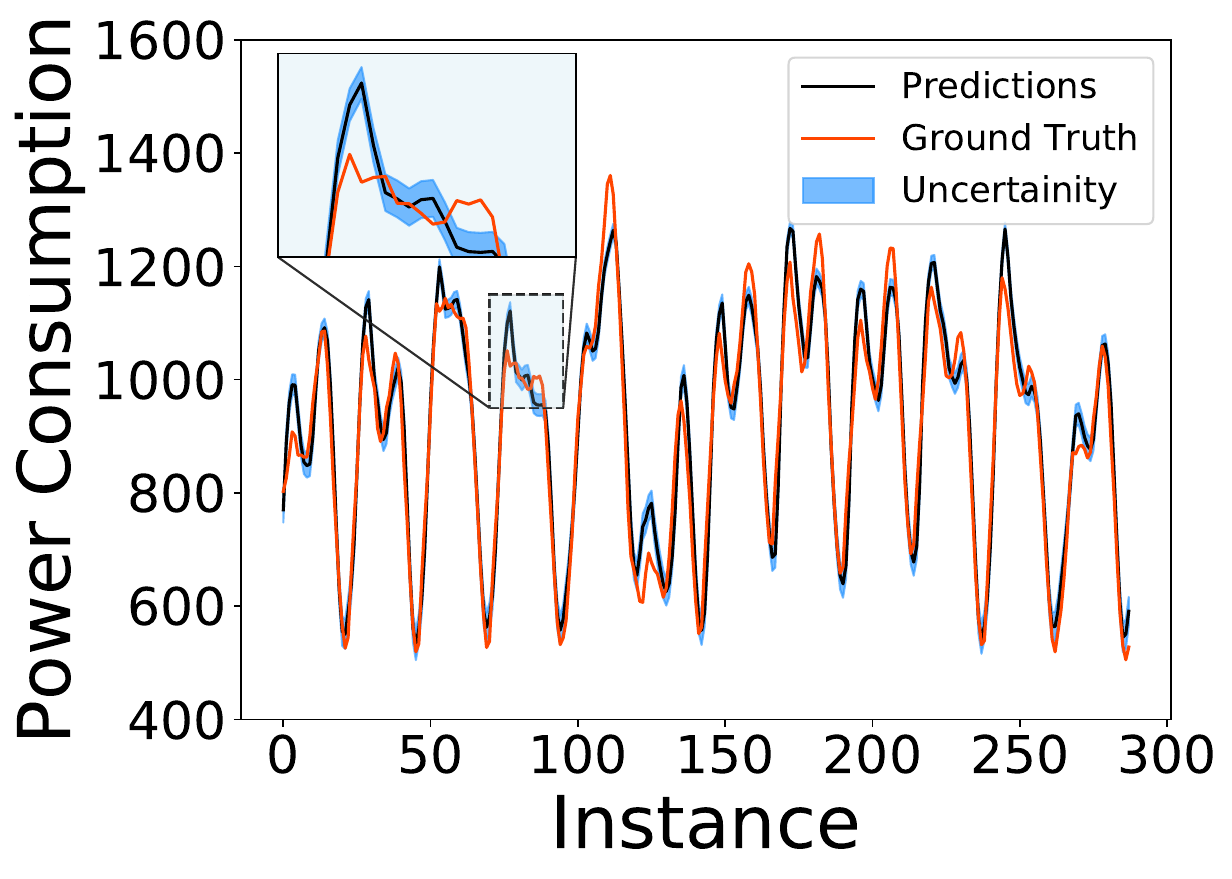}
}
\subfloat[Node 99 in Electricity]{
  \includegraphics[width=45mm]{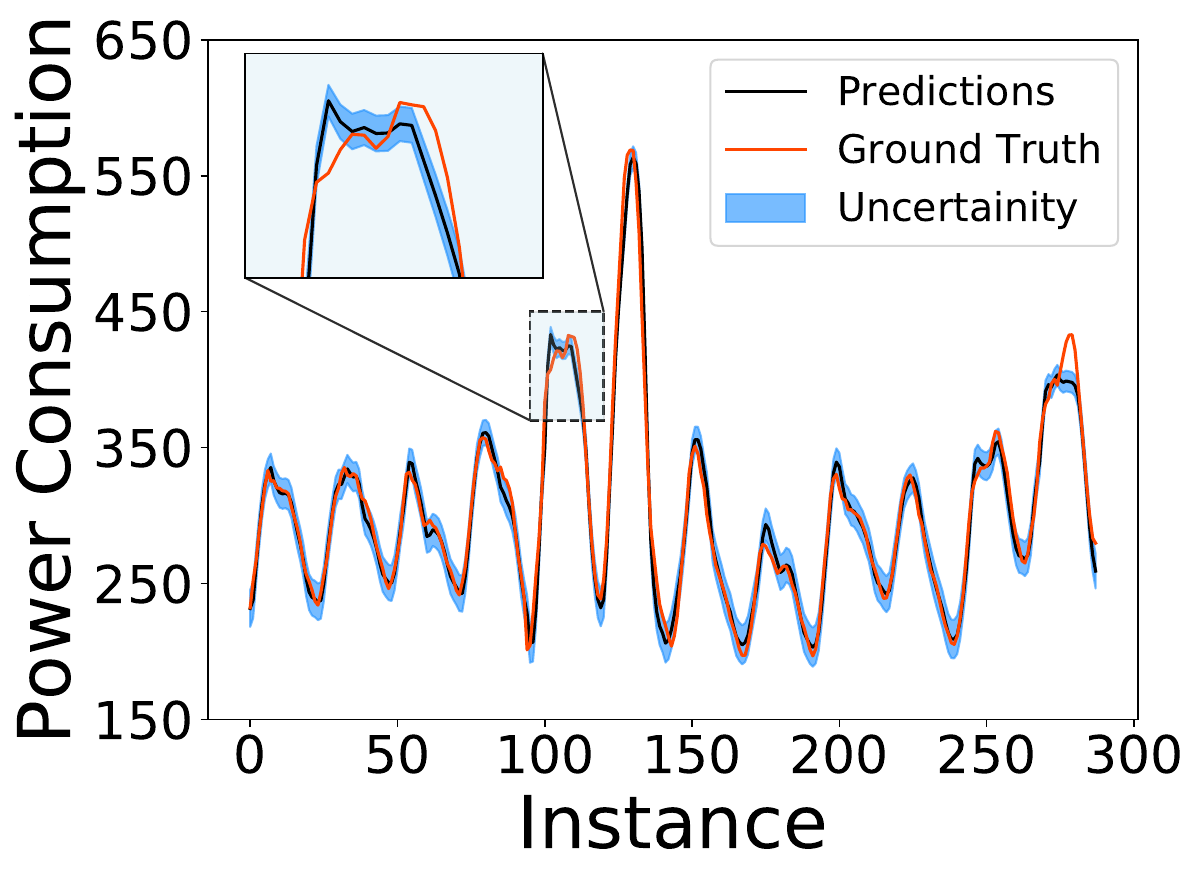}
}
\subfloat[Node 196 in Electricity]{
  \includegraphics[width=45mm]{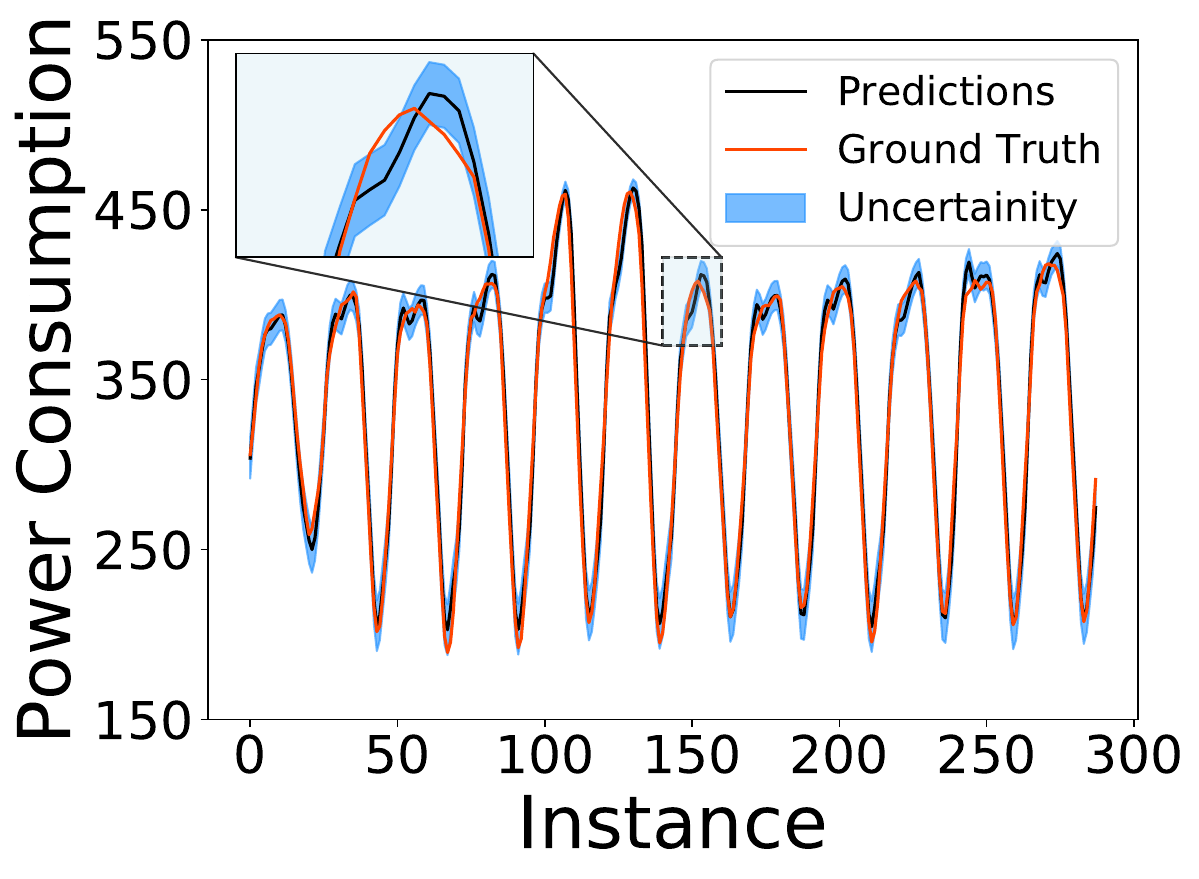}
}
\subfloat[Node 290 in Electricity]{
  \includegraphics[width=45mm]{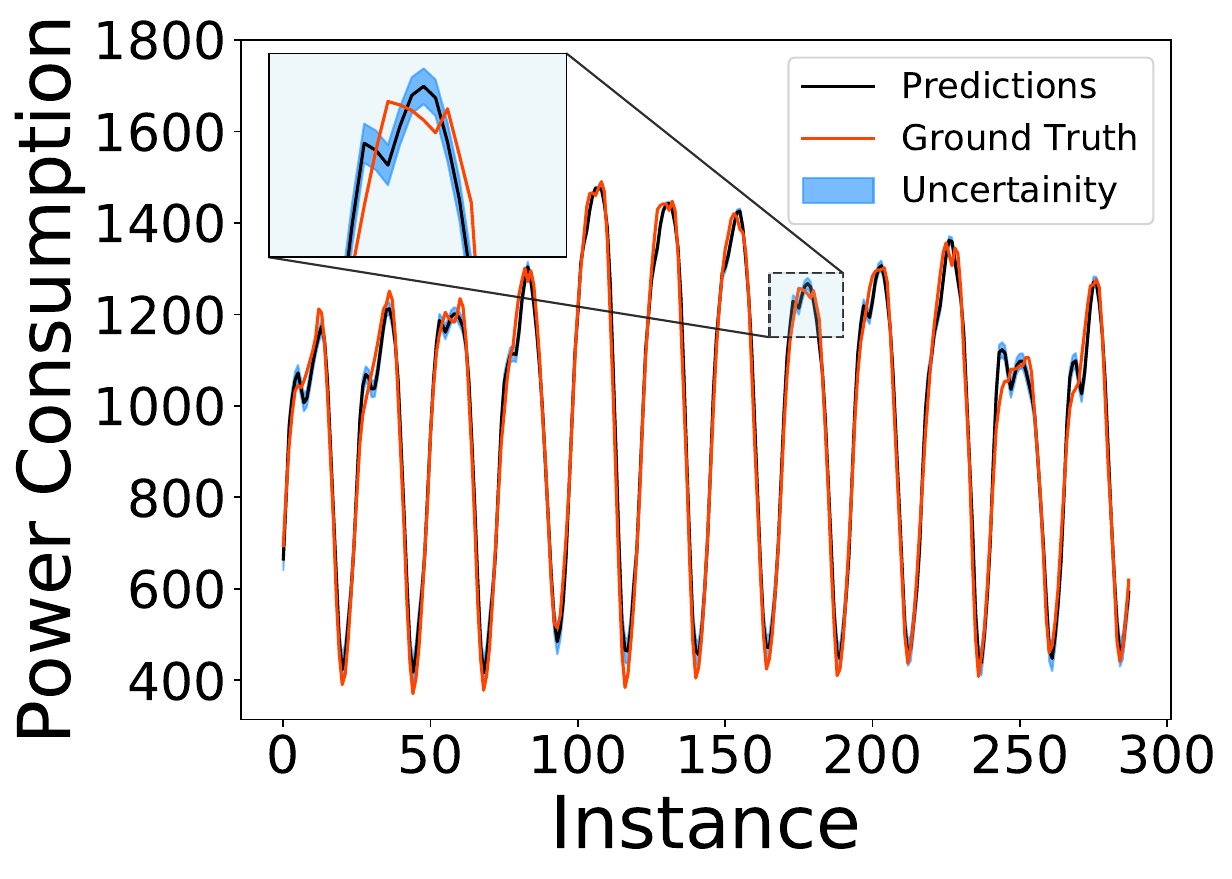}
}
\hspace{0mm}
\subfloat[Node 14 in Solar]{
  \includegraphics[width=45mm]{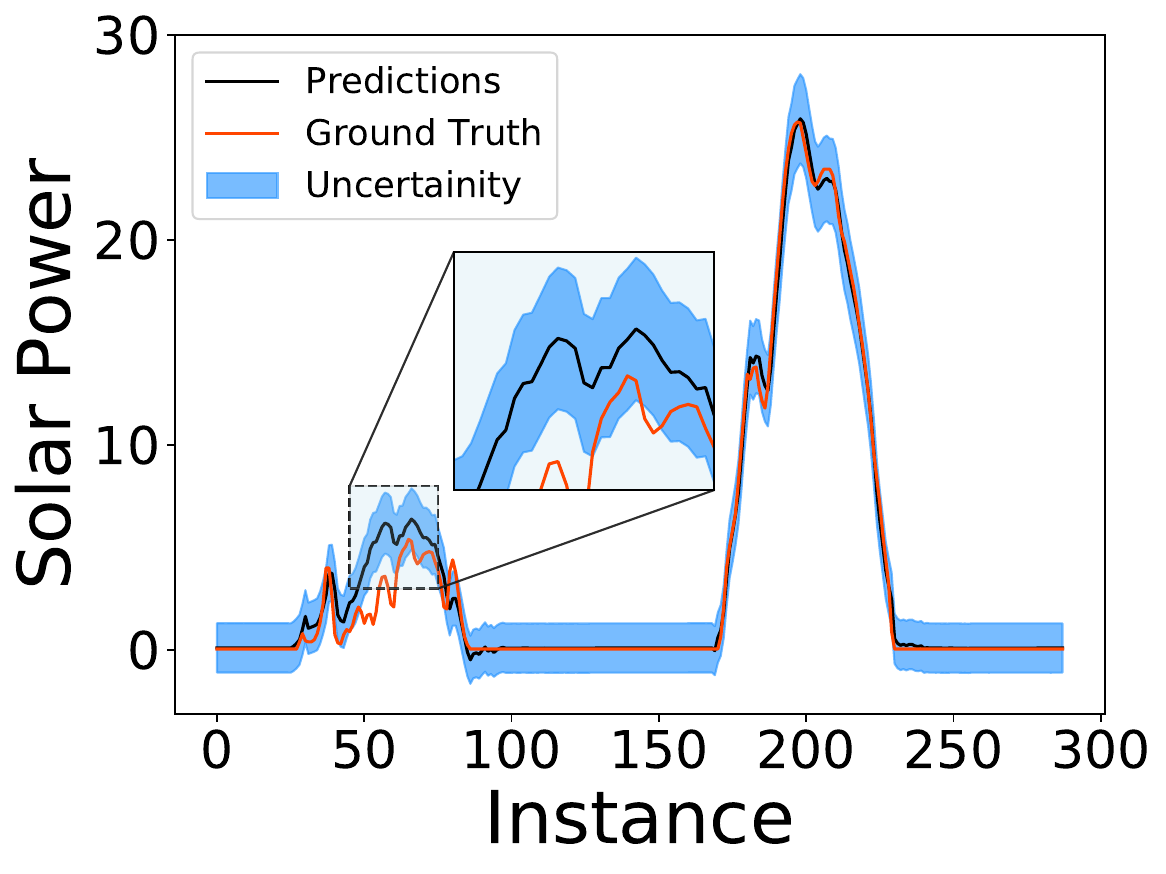}
}
\subfloat[Node 46 in Solar]{
  \includegraphics[width=45mm]{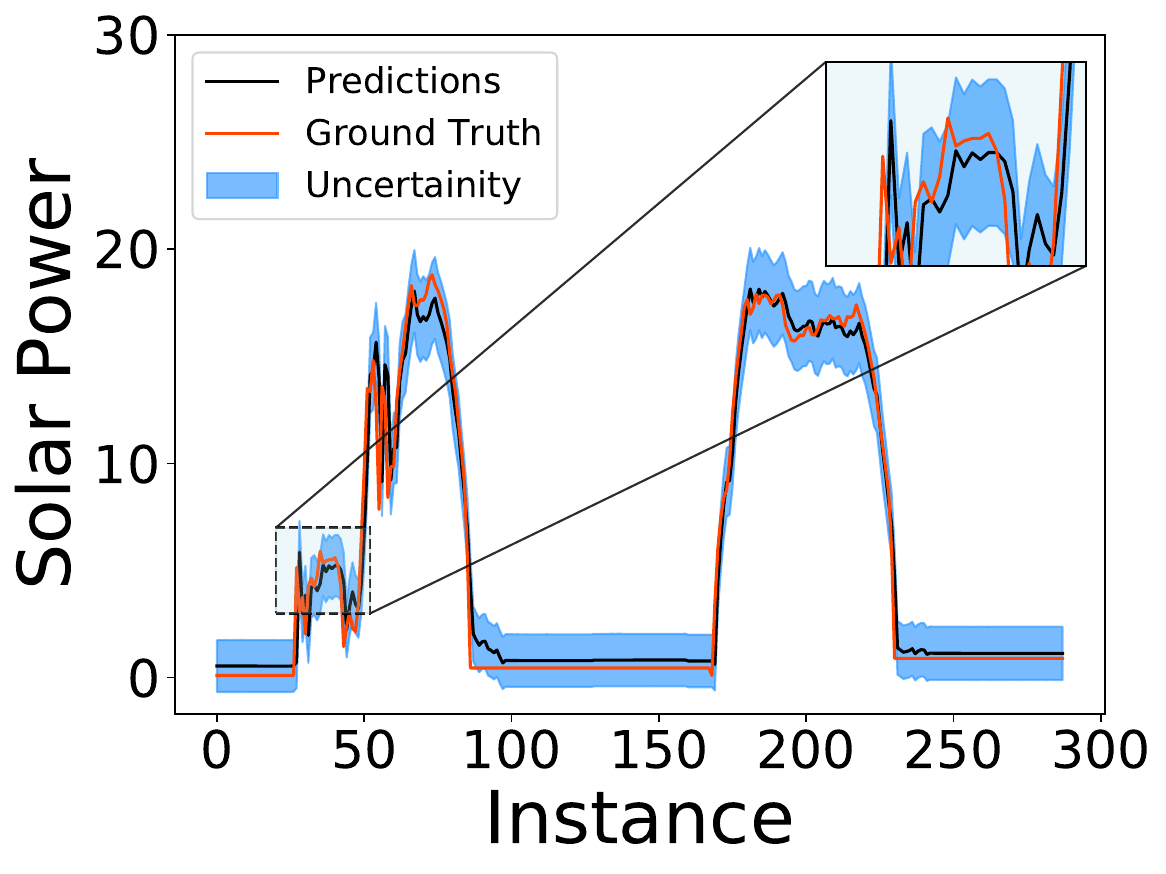}
}
\subfloat[Node 83 in Solar]{
  \includegraphics[width=45mm]{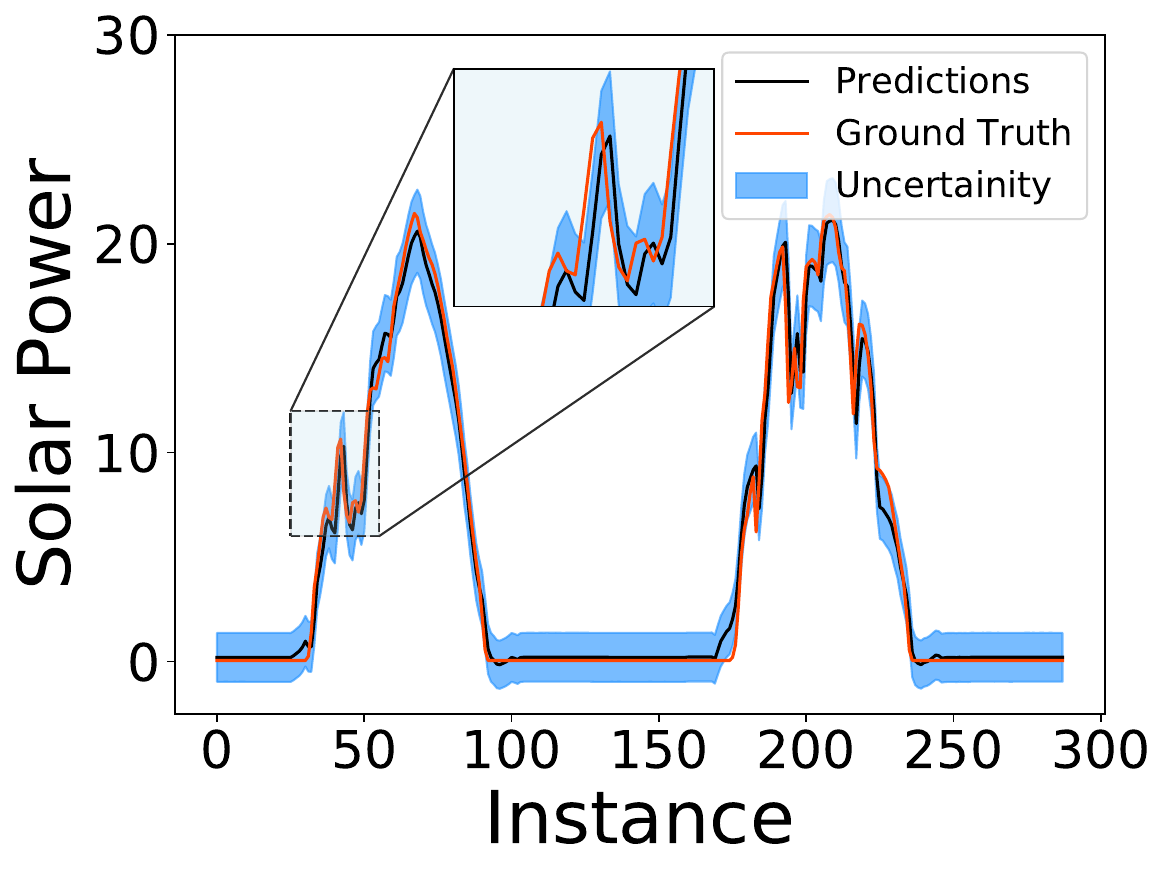}
}
\subfloat[Node 130 in Solar]{
  \includegraphics[width=45mm]{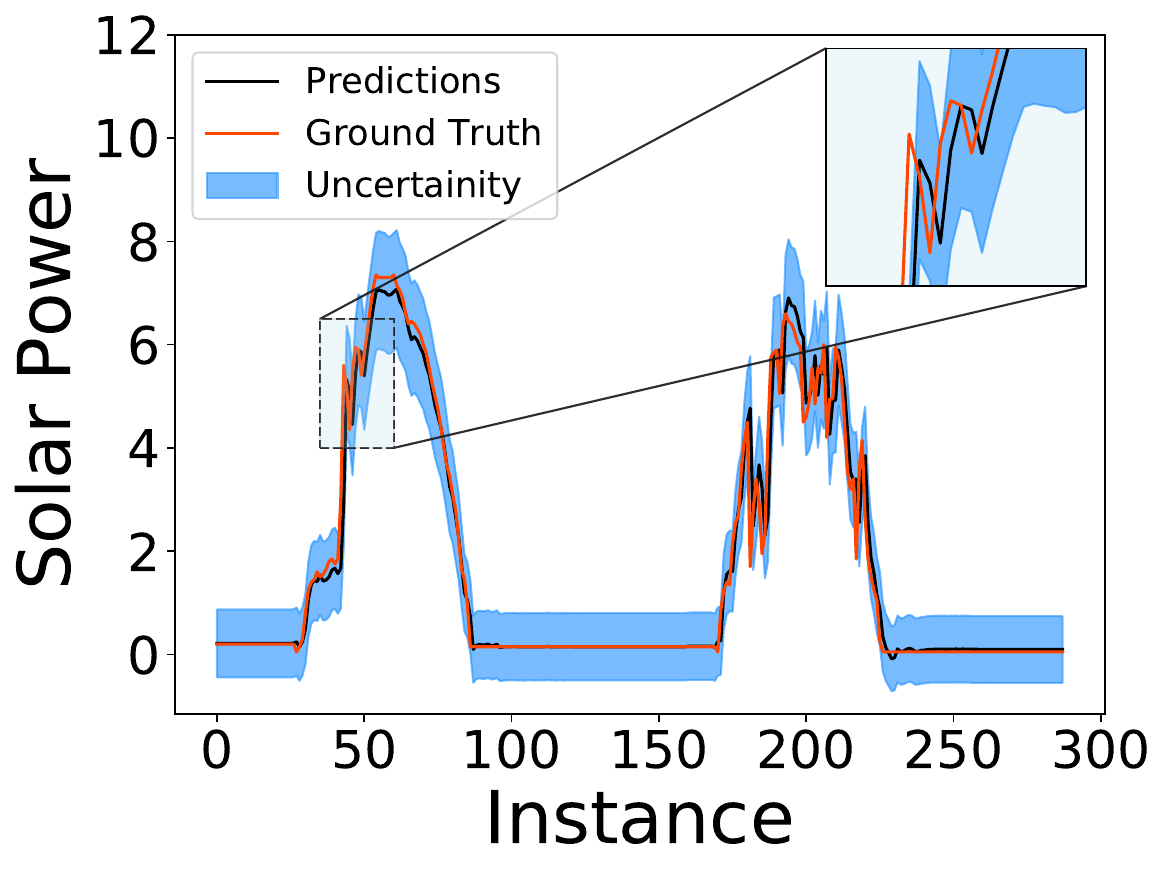}
 }
\caption{The visualizations provide an overview of the model forecasts, ground-truth data, and uncertainty estimates obtained through multi-horizon forecasting on benchmark datasets.}
\label{fig:Overall_Plots}
\end{figure*}

Figure \ref{fig:Overall_Plots} depicts the ground truth, pointwise forecasts, and time-varying uncertainty estimates obtained from the proposed \textbf{w/Unc-JHgRF-Net} framework.
The visualizations provide valuable insights into the framework performance and facilitates comprehensive analysis and result interpretation. Existing methods for MTSF can model nonlinear spatio-temporal dependencies within interconnected sensor networks but often fail to provide accurate measures of uncertainty. In contrast, the proposed \textbf{w/Unc-JHgRF-Net} framework(\textbf{JHgRF-Net} framework with local uncertainty estimation) effectively utilizes relational inductive bias via spatio-temporal propagation architecture to quantitatively estimate uncertainty of multi-horizon forecasts. The framework accurately estimates uncertainty, outperforming existing methods that solely provide pointwise forecasts for MTSF. The multifaceted visualizations shows the framework effectiveness in time series representation learning for the MTSF task, making it a valuable contribution to the field of multivariate time series analysis for uncertainty estimation.

\subsection{Datasets} 
Our novel \textbf{JHgRF-Net} framework effectiveness was evaluated by comparing it with existing benchmark models on several real-world datasets, including PeMSD3, PeMSD4, PeMSD7, PeMSD7(M), PeMSD8, Electricity, Solar-Energy, Exchange-Rate, Traffic, METR-LA,  PEMS-BAY, SWaT, and WADI. Tables \ref{tab:summarydatasets1} - \ref{tab:summarydatasets3} provides additional information regarding these benchmark datasets. The datasets used in this study comprise Solar-Energy, which records solar power production from 137 PV plants in Alabama state at 10-minute intervals in 2016; Electricity, which includes hourly records of electricity consumption(kWh) for 321 clients from 2012 to 2014; Exchange-Rate, which collects daily exchange rates of eight foreign countries from 1990 to 2016; and Traffic dataset provides hourly data on road occupancy rates(ranging from 0 to 1) that were recorded on the various lanes of San Francisco Bay area freeways from 2015 to 2016, spanning 48 months in total. Moreover, METR-LA contains hourly data of traffic speed from loop detectors on the highways in Los Angeles, while PEMS-BAY includes data on traffic volume and speed from sensors on San Francisco Bay area freeways. Additionally, PeMS is an open-access dataset that consists of five traffic network datasets(PeMSD3, PeMSD4, PeMSD7, PeMSD7(M), and PeMSD8), obtained from the Caltrans Performance Measurement System across five California districts, with data points available at 5-minute intervals, providing 288 data points per day. The SWaT dataset consists of 11 days of continuous operation from an industrial water treatment plant, comprising 7 days of normal operation and 4 days with 41 attack scenarios. Two versions of the dataset are available, with Version 1 excluding the first 30 minutes and containing only normal operation data. The WADI dataset is a collection of 16 days of continuous operation from a testbed, encompassing 14 days of normal operation and 2 days with 15 attack scenarios. An updated version is available that removes affected readings due to plant instability during certain periods.

\vspace{-2mm}
\begin{table}[htbp]
\center
\caption{Summary of the traffic-related benchmark datasets.}
\vspace{-3mm}
\label{tab:summarydatasets1}
\setlength{\tabcolsep}{0.25em} 
\renewcommand\arraystretch{1.15} 
\centering
 \resizebox{0.515\textwidth}{!}{
\hspace{-10mm}\begin{tabular}{c|c|c|c|c|c}
\hline
\textbf{Dataset} & \textbf{Variables} & \textbf{Timepoints} & \textbf{Time-Range} & \multicolumn{1}{l|}{\textbf{Split-Ratio}} & \multicolumn{1}{l}{\textbf{Granularity}} \\ \hline
PeMSD3 & 358 & 26,208 & 09/2018 - 11/2018 & \multirow{5}{*}{6 / 2 / 2} & \multirow{7}{*}{\rotatebox[origin=c]{270}{5 mins}} \\
PeMSD4 & 307 & 16,992 & 01/2018 - 02/2018 &  &  \\
PeMSD7 & 883 & 28,224 & 05/2017 - 08/2017 &  &  \\
PeMSD8 & 170 & 17,856 & 07/2016 - 08/2016 &  &  \\
PeMSD7(M) & 228 & 12,672 & 05/2012 - 06/2012 &  &  \\ \cline{1-5}
METR-LA & 207 & 34,272 & 03/2012 - 06/2012 & \multirow{2}{*}{7 / 1 / 2} &  \\
PEMS-BAY & 325 & 52,116 & 01/2017 - 05/2017 &  &  \\ \hline
\end{tabular}
}
\vspace{-2mm}
\end{table}

\vspace{-3mm}
\begin{table}[!h]
\center
\caption{Summary of the traffic, solar, electricity, and exchange-rate datasets}
\vspace{-2mm}
\label{tab:summarydatasets2}
\setlength{\tabcolsep}{0.25em} 
\renewcommand\arraystretch{1.15} 
\centering
\hspace*{-2mm}
 \resizebox{0.5\textwidth}{!}{
\begin{tabular}{c|c|c|c|c}
\hline
\textbf{Dataset} &
  \multicolumn{1}{l|}{\textbf{Variables}} &
  \multicolumn{1}{l|}{\textbf{Timepoints}} &
  \multicolumn{1}{l|}{\textbf{Granularity}} &
  \multicolumn{1}{l}{\textbf{Split-Ratio}} \\ \hline
Traffic       & 862 & 17544 & 1 hour & \multirow{4}{*}{6/2/2} \\
Solar         & 137 & 52560 & 10 min &                        \\
Electricity   & 321 & 26304 & 1 hour &                        \\
Exchange-rate & 8   & 7588  & 1 day  &                        \\ \hline
\end{tabular}
}
\vspace{-2mm}
\end{table}

\vspace{-3mm}
\begin{table}[!htbp]
\center
\caption{Summary of the SWaT and WADI datasets}
\vspace{-3mm}
\label{tab:summarydatasets3}
\setlength{\tabcolsep}{0.25em} 
\renewcommand\arraystretch{1.15} 
\centering
\hspace*{-2mm}
 \resizebox{0.5\textwidth}{!}{
\begin{tabular}{c|c|c|c|c}
\hline
\textbf{Dataset} & \textbf{Variables} & \textbf{Training Points} & \textbf{Testing Points} & \textbf{Anomalies} \\ \hline
SWaT             & 51                 & 47515                    & 44986                   & 11.97              \\
WADI             & 127                & 118795                   & 17275                   & 5.99               \\ \hline
\end{tabular}
}
\vspace{-2mm}
\end{table}

\vspace{-3mm}
\subsection{Experimental Study design}
\vspace{0mm}
In order to examine the effectiveness of the proposed models(\textbf{JHgRF-Net}, \textbf{w/Unc-JHgRF-Net}) compared to the baselines, the various benchmark datasets were split into training, validation, and test sets. The PEMS-BAY and METR-LA datasets were split with a 7/1/2 ratio, while all other datasets were split with a 6/2/2 ratio. SWaT and WADI datasets have predefined splits, where the training set is anomaly-free and the test set contains anomalies. We utilize the training sets for our experiments. To prepare the SWaT and WADI datasets for training and evaluation, we normalize each variable data by rescaling it to fit within the range of [0, 1], as in \cite{deng2021graph}. For all other benchmark datasets, each variable data was preprocessed by scaling to have zero mean and unit variance. During the training and evaluation of forecasting models, various accuracy metrics, including MAE, RMSE, and MAPE, were calculated based on the original scale of the time series data.  The \textbf{JHgRF-Net} architecture was trained for 30 epochs on the training set to minimize forecast error. The validation set was employed to identify the optimal model that improves overall performance and early stopping was utilized to prevent overfitting. The framework performance was evaluated on the test set to examine its ability to perform well on unseen data. To improve convergence, the model training was optimized by using a learning rate scheduler to effectively learn from the training set. If there was no improvement in the evaluation metrics on the validation set over five epochs, the learning rate was reduced by half. The Adam optimizer was employed to fine-tune the trainable parameters of the models. An initial learning rate of \num{1e-3} was set to minimize the MAE loss for the \textbf{JHgRF-Net} model and the negative Gaussian log-likelihood for the \textbf{w/Unc-JHgRF-Net} model, ensuring a better fit between the ground truth and the model predictions. The use of powerful GPUs such as NVIDIA Tesla T4, Nvidia Tesla V100, and GeForce RTX 2080 GPUs expedited the training process and allowed for the utilization of larger models and datasets based on the PyTorch framework. Multiple independent experimental runs were conducted, and the ensemble average was reported to ensure reliable model evaluation. In the Section \ref{sec:hyperparameters}, we reported the optimum hyperparameters values of the learning algorithm for each dataset, such as embedding size, number of hyperedges, batch size, and learning rate. 
 
\subsection{Baselines}
Established algorithms are commonly used as benchmarks for evaluating the performance of proposed neural forecasting models such as \textbf{JHgRF-Net} and \textbf{w/Unc-JHgRF-Net} on the MTSF task. The selection of benchmark algorithms depends on their extensive usage in the literature and their demonstrated performance on benchmark datasets.

\vspace{0mm}
\begin{itemize}
\item HA~\cite{hamilton2020time} is a time series prediction technique that involves using the average of a predefined historical window of observations to predict the next value in the time series.
\item ARIMA is a statistical analysis model commonly used for handling non-stationary time series data, but it has limitations in handling long-term trends or changing seasonal patterns over time. 
\item VAR(\cite{hamilton2020time}) is a linear multivariate time series model that extends the univariate autoregressive(AR) model. It is designed to capture the inter-dependencies among multiple time series variables for analyzing and forecasting complex systems.
\item TCN(~\cite{BaiTCN2018})  is specifically designed to handle sequential data in multistep-ahead time series prediction tasks by using causal convolutions and dilation layers to incorporate past information and learn long-range correlations. The use of these techniques allows the model to capture and learn relationships between multiple variables in time series data, making it effective in handling such complex data.
\item FC-LSTM(~\cite{sutskever2014sequence}) is an encoder-decoder architecture that employs Long Short-Term Memory(LSTM) units with peephole connections to perform multistep-ahead time series prediction. By capturing both short-term and long-term relationships among multiple time series variables in MTS data, this architecture effectively models the intricate patterns and relationships, resulting in a highly complex and nonlinear representation of the data and improved forecasting accuracy.
\item GRU-ED(~\cite{cho2014grued}) is an encoder-decoder architecture that utilizes Gated Recurrent Unit (GRU) units to handle sequential data in multi-horizon time series prediction tasks. By capturing relevant information from previous time steps, this framework effectively models the sequential dependencies in the data.
\item DSANet(~\cite{Huang2019DSANet}) is a time series forecasting method that utilizes convolutional neural networks (CNNs) to capture long-range intra-temporal dependencies among multiple time series, without relying on recurrent networks. In addition, it further incorporates self-attention blocks to adaptively capture interdependencies and generate multi-horizon forecasts for MTS data, resulting in an extremely effective and precise forecasting method.
\item DCRNN(~\cite{li2018dcrnn_traffic}) is a highly effective technique that combines graph convolution with recurrent neural networks, utilizing bidirectional random walks on graphs. This unique approach enables the model to predict multistep-ahead forecasts in MTS data through an encoder-decoder architecture, which effectively captures complex spatial-temporal dependencies in the data, resulting in highly accurate predictions.
\item STGCN(~\cite{bing2018stgcn})is a cutting-edge technique that seamlessly integrates graph convolution and gated temporal convolution networks. By accurately capturing the spatial-temporal correlations among multiple time series variables, this approach enables multi-horizon time series prediction with high precision and accuracy.
\item GraphWaveNet(~\cite{wu2019graphwavenet}) uses a wave-based propagation mechanism and graph representations that are computed from dilated causal convolution neural networks to model MTS data. By jointly learning an adaptive dependency matrix and capturing spatial-temporal dependencies, this method effectively captures the dependencies between multiple time series variables, leading to improved multistep-ahead time series prediction accuracy.
\item ASTGCN(~\cite{guo2019astgcn}) utilizes an attention-based spatio-temporal graph convolutional neural network to capture inter- and intra-dependencies for predicting multihorizon forecasts in time series data. By using attention mechanisms, this technique effectively models the spatial-temporal relationships between multiple time series variables, resulting in highly accurate predictions.
\item STG2Seq(~\cite{bai2019STG2Seq}) is an advanced technique for predicting multistep-ahead forecasts in MTS data that incorporates gated graph convolutional networks(GGCNs) with a sequence-to-sequence(seq2seq) architecture featuring attention mechanisms. Through this unique combination, the model captures dynamic temporal and cross-channel information to effectively model the complex relationships among multiple time series variables, resulting in highly accurate and precise predictions.
\item STSGCN(~\cite{song2020stsgcn}) is a time series prediction technique that utilizes multiple layers of spatial-temporal graph convolutional networks to predict multistep-ahead forecasts in MTS data. This approach captures localized intra- and inter-dependencies in the graph-structured MTS data, effectively modeling the complex relationships between multiple time series variables and resulting in high precision and accuracy for multi-horizon time series prediction.
\item LSGCN(~\cite{huang2020lsgcn}) is a method used for multi-horizon time series forecasting, utilizing a graph attention mechanism integrated into a spatial gated block to predict multistep-ahead forecasts in MTS data. By utilizing attention mechanisms, it can effectively model dynamic spatial-temporal dependencies between multiple time series variables, leading to improved forecasting accuracy.
\item AGCRN(~\cite{NEURIPS2020_ce1aad92}) predicts multistep-ahead forecasts in MTS data by utilizing a data-adaptive graph structure learning method. This approach captures node-specific intra- and inter-correlations, effectively modeling complex spatial-temporal dependencies among the multiple time series variables, resulting in improved forecasting accuracy.
\item STFGNN(~\cite{li2021stfgnn}) is a time series prediction technique that fuses representations obtained from temporal graph and gated convolutional neural networks to predict multistep-ahead forecasts in MTS data. By operating these networks in parallel and learning spatial-temporal correlations, STFGNN effectively models the complex relationships between multiple time series variables, leading to improved forecasting accuracy.
\item Z-GCNETs(~\cite{chen2021ZGCNET}) predicts multi-horizon forecasts in MTS data by integrating a time-aware zigzag topological layer into time-conditioned graph convolutional networks. It captures hidden spatial-temporal dependencies and salient time-conditioned topological information to effectively model complex relationships between multiple time series variables while considering their topological properties. This approach performs well in time series prediction tasks that require the modeling of complex dependencies and relationships.
\item STGODE(~\cite{fang2021STODE}) predicts multistep-ahead forecasts in MTS data using a tensor-based ordinary differential equation (ODE) to capture inter- and intra-dependency dynamics among multiple time series variables. By effectively representing the MTS data, the model can capture the complex relationships among variables and their temporal dynamics, resulting in highly accurate and reliable predictions.
\item GDN(~\cite{deng2021graph}) is a graph-based anomaly detection model that leverages graph embeddings to learn the inherent complex graph structure underlying the MTS data and further employs a graph forecasting network to compute deviation scores for anomaly detection based on a threshold on forecast error.  
\item NRI(~\cite{kipf2018neural}) is an unsupervised technique that learns to deduce interactions and dynamics from observational data within a variational auto-encoder framework. This model effectively predicts the inherent dynamics of complex systems, making it a valuable tool for a wide range of applications.
\item MTGNN(~\cite{wu2020connecting}) presents a graph neural network framework for modeling multivariate time series data. This innovative approach automatically extracts uni-directed relations among multiple time series variables through a graph learning module, while also incorporating mix-hop propagation and dilated inception layers to capture spatial and temporal dependencies within the time series. The result is a highly accurate model capable of multi-horizon forecasts, making it a suitable tool for various time-series applications.
\item The vanilla LSTM(~\cite{hochreiter1997long}) predicts multistep-ahead forecasts in MTS data using gating mechanism. The LSTM-U, consisting of N univariate LSTMs, treats all time series variables as independent and performs univariate multi-horizon forecasting.
\end{itemize}

\vspace{-2mm}
\subsection{Forecasting uncertainty}
The loss function for training the \textbf{JHgRF-Net} framework is the mean absolute error(MAE), which is calculated by comparing the pointwise forecasts of the model predictions (\resizebox{.095\textwidth}{!}{$\widehat{\mathbf{X}}_{(t  : t + \upsilon-1)}$}) with the corresponding ground-truth data(\resizebox{.095\textwidth}{!}{$\mathbf{X}_{(t  : t + \upsilon-1)}$}), computed as follows:

\vspace{-2mm}
\resizebox{0.945\linewidth}{!}{
\begin{minipage}{\linewidth}
\begin{align}
\mathcal{L}_{\text{MAE}}\left(\theta\right) \hspace{0.5mm} = \hspace{0.5mm}\frac{1}{\upsilon}\left|\mathbf{X}_{(t  : t + \upsilon-1)}-\widehat{\mathbf{X}}_{(t  : t + \upsilon-1)}\right| \label{eq:UCE} \nonumber
\end{align}
\end{minipage}
}

\vspace{2mm}
During the training process, the model parameters $\theta$ are fine-tuned to minimize the mean absolute error(MAE) loss function, represented as $\mathcal{L}_{\text{MAE}}\left(\theta\right)$. The \textbf{w/Unc-JHgRF-Net} is a variant of the \textbf{JHgRF-Net} that estimates the uncertainty in model predictions to enhance the reliability of decision-making. The framework utilizes a heteroscedastic Gaussian distribution to predict time-varying uncertainty in model predictions, characterized by mean and variance denoted by \resizebox{.125\textwidth}{!}{$\mu_\phi\big(\overline{\mathbf{X}}_{(t  : t + \upsilon-1)}\big)$} and \resizebox{.125\textwidth}{!}{$\sigma_\phi^2\big(\overline{\mathbf{X}}_{(t  : t + \upsilon-1)}\big)$}, respectively, while the input time series is denoted as \resizebox{.09\textwidth}{!}{$\overline{\mathbf{X}}_{(t  : t + \upsilon-1)}$}.  It is mathematically described as follows, 

\vspace{-2mm}
\resizebox{0.935\linewidth}{!}{
\begin{minipage}{\linewidth}
\begin{align}
\widehat{\mathbf{X}}_{(t  : t + \upsilon-1)} \hspace{0.5mm} \sim \hspace{1mm}\mathcal{N}\big(\mu_\phi\big(\overline{\mathbf{X}}_{(t  : t + \upsilon-1)}\big), \sigma_\phi^2\big(\overline{\mathbf{X}}_{(t  : t + \upsilon-1)}\big)\big)  \nonumber
\end{align}
\end{minipage}
}

\vspace{1mm}
The predicted mean and standard deviation can be obtained from the following equation, as follows:

\vspace{-2mm}
\resizebox{0.935\linewidth}{!}{
\begin{minipage}{\linewidth}
\begin{align}
\mu_\phi\big(\overline{\mathbf{X}}_{(t  : t + \upsilon-1)}\big), \sigma_\phi^2\big(\hat{\mathbf{X}}_{(t  : t + \upsilon-1)}\big) &= f_\theta\big( \widehat{\mathbf{X}}_{(t  : t + \upsilon-1)}\big) \nonumber
\end{align}
\end{minipage}
}

\vspace{2mm}
A neural network $f_{\theta}$ takes the output of the spatio-temporal inference component, represented by \resizebox{.09\textwidth}{!}{$\widehat{\mathbf{X}}_{(t  : t + \upsilon-1)}$}, and predicts the mean and standard deviation of future observations. The future observations are represented by \resizebox{.09\textwidth}{!}{$\widehat{\mathbf{X}}_{(t  : t + \upsilon-1)}$} and it is the Maximum likelihood estimation(MLE) of the predicted Gaussian distribution. It is mathematically described as follows,

\vspace{-4mm}
\resizebox{1.025\linewidth}{!}{
\begin{minipage}{\linewidth}
\begin{align}
\widehat{\mathbf{X}}_{(t  : t + \upsilon-1)} \hspace{0.5mm}= \hspace{0.5mm} \mu_\phi\big(\overline{\mathbf{X}}_{(t  : t + \upsilon-1)}\big) \nonumber
\end{align}
\end{minipage}
}

\vspace{2mm}
To predict a Gaussian(normal) distribution for sampling future observations, we typically use the maximum likelihood estimates(MLE) of the distribution's parameters, namely the mean and standard deviation. The MLE values are obtained by maximizing the likelihood of the observed data being generated by the predicted distribution. In simpler terms, \resizebox{.125\textwidth}{!}{$\mu_\phi\big(\overline{\mathbf{X}}_{(t  : t + \upsilon-1)}\big)$} estimates the future values, \resizebox{.08\textwidth}{!}{$\widehat{\mathbf{X}}_{(t  : t + \upsilon-1)}$}, using the input time series, \resizebox{.08\textwidth}{!}{$\overline{\mathbf{X}}_{(t  : t + \upsilon-1)}$}. Meanwhile, \resizebox{.125\textwidth}{!}{$\sigma_\phi^2\big(\overline{\mathbf{X}}_{(t  : t + \upsilon-1)}\big)$} predicts the uncertainty in the model's predictions over the next $\upsilon$ time steps, starting from the current time point, $t$. The uncertainty modeling framework optimizes the negative Gaussian log likelihood(\cite{nix1994estimating}) of the observations, based on estimates of the mean and variance. This approach provides a more comprehensive understanding and measurement of prediction uncertainty. The negative Gaussian log likelihood measures the likelihood of the observations, given the estimated mean and variance of the Gaussian distribution. A lower negative Gaussian log likelihood indicates a better fit of the Gaussian distribution to the observed values. The Gaussian distribution to predict the future observations is described by,

\vspace{-1mm}
\resizebox{0.5\linewidth}{!}{
\hspace{-30mm}\begin{minipage}{\linewidth}
\begin{equation*}
\mathcal{N}(\widehat{\mathbf{X}}_{(t  : t + \upsilon-1)};\mu_\phi\big(\overline{\mathbf{X}}_{(t - \tau : \hspace{1mm}t-1)}\big),\sigma_\phi\big(\overline{\mathbf{X}}_{(t - \tau : \hspace{1mm}t-1)}\big)) \hspace{0.5mm} = \hspace{0.5mm} {\frac {1}{\sigma_\phi\big(\overline{\mathbf{X}}_{(t - \tau : \hspace{1mm}t-1)}\big) {\sqrt {2\pi }}}}\hspace{1.5mm} e^{-{\dfrac {1}{2}}\left({\dfrac {\mathbf{X}_{(t  : t + \upsilon-1)}-\mu_\phi\big(\overline{\mathbf{X}}_{(t - \tau : \hspace{1mm}t-1)}\big) }{\sigma_\phi\big(\overline{\mathbf{X}}_{(t - \tau : \hspace{1mm}t-1)}\big)}}\right)^{2}}
\end{equation*} 
\end{minipage} 
}

\vspace{2mm}
We apply logarithm transformation on both sides of the equation, 

\vspace{-2mm}
\resizebox{1.25\linewidth}{!}{
\hspace{-3mm}\begin{minipage}{\linewidth}
\begin{align*}
\log\ \mathcal{N} &= \hspace{0.5mm} \log\left[{\frac {1}{\sigma_\phi\big(\overline{\mathbf{X}}_{(t - \tau : \hspace{1mm}t-1)}\big) {\sqrt {2\pi }}}}\right] + \log \left[e^{-{\dfrac {1}{2}}\left({\dfrac {\mathbf{X}_{(t  : t + \upsilon-1)}-\mu_\phi\big(\overline{\mathbf{X}}_{(t - \tau : \hspace{1mm}t-1)}\big) }{\sigma_\phi\big(\overline{\mathbf{X}}_{(t - \tau : \hspace{1mm}t-1)}\big)}}\right)^{2}}\right] \\
 \hspace{0.5mm} &= \hspace{0.5mm}\log\ {\frac {1}{\sigma_\phi\big(\overline{\mathbf{X}}_{(t - \tau : \hspace{1mm}t-1)}\big)}} + \log\ {\frac {1}{{\sqrt {2\pi }}}} -{\frac {1}{2}}\left(\dfrac {\mathbf{X}_{(t  : t + \upsilon-1)}-\mu_\phi\big(\overline{\mathbf{X}}_{(t - \tau : \hspace{1mm}t-1)}\big) }{\sigma_\phi\big(\overline{\mathbf{X}}_{(t - \tau : \hspace{1mm}t-1)}\big)}\right)^{2} \\
 \hspace{0.5mm} &= \hspace{0.5mm} -\log\ \sigma_\phi\big(\overline{\mathbf{X}}_{(t - \tau : \hspace{1mm}t-1)}\big) + C -{\frac {1}{2}}\left(\dfrac {\mathbf{X}_{(t  : t + \upsilon-1)}-\mu_\phi\big(\overline{\mathbf{X}}_{(t - \tau : \hspace{1mm}t-1)}\big) }{\sigma_\phi\big(\overline{\mathbf{X}}_{(t - \tau : \hspace{1mm}t-1)}\big)}\right)^{2}
\end{align*}
\end{minipage}
}

\vspace{3mm}
We drop the constant(C) and the Gaussian negative log likelihood loss(i.e., negative log gaussian probability density function(pdf)) for the dataset is described by, 

\vspace{0mm}
\resizebox{1.15\linewidth}{!}{
\hspace{-15mm}\begin{minipage}{\linewidth}
\begin{align}
\hspace{5mm}\mathcal{L}_{\text{GaussianNLLLoss}} =  \sum_{t=1}^{\text{T}} \left[\frac{\log \sigma_\phi\big(\overline{\mathbf{X}}_{(t - \tau : \hspace{1mm}t-1)}\big)^2}{2}+\frac{\left(\mathbf{X}_{(t  : t + \upsilon-1)}-\mu_\phi\left(\overline{\mathbf{X}}_{(t - \tau : \hspace{1mm}t-1)}\right)\right)^2}{2 \sigma_\phi\left(\overline{\mathbf{X}}_{(t - \tau : \hspace{1mm}t-1)}\right)^2}\right] \label{eq:GUCE} \nonumber
\end{align}
\end{minipage}
}

\vspace{2mm}
The negative Gaussian log-likelihood is used to evaluate the model's fit for estimated mean and variance to the observations in the training set, where $\text{T}$ denotes the time steps. To recapitulate, the objective of the \textbf{JHgRF-Net} framework is to minimize the Mean Absolute Error(MAE), while the \textbf{w/Unc-JHgRF-Net} framework aims to minimize the $\textbf{L}_\textbf{GaussianNLLLoss}$ to provide quantitative uncertainty estimation. The \textbf{w/Unc-JHgRF-Net} framework uses a Gaussian likelihood function to model the mean and variance, optimize the model parameters for data fitting, and provide uncertainty estimates.

\end{document}

%% file: math_commands.tex
\usepackage{amsmath,amsfonts,bm}

\def\eqref#1{equation~\ref{#1}}

\def\1{\bm{1}}

\DeclareMathAlphabet{\mathsfit}{\encodingdefault}{\sfdefault}{m}{sl}
\SetMathAlphabet{\mathsfit}{bold}{\encodingdefault}{\sfdefault}{bx}{n}

